\documentclass[runningheads]{llncs}

\usepackage{eccv}

\usepackage{eccvabbrv}
\usepackage{multirow}
\usepackage{graphicx}
\usepackage{booktabs}
\usepackage{subcaption}
\usepackage{arydshln}
\usepackage{bbding}
\usepackage{algorithm,algorithmic}
\usepackage[accsupp]{axessibility}  

\usepackage[normalem]{ulem}

\usepackage[pagebackref,breaklinks,colorlinks,citecolor=eccvblue]{hyperref}

\usepackage{orcidlink}
\usepackage[symbol]{footmisc}

\begin{document}

\title{LIPT: Latency-aware Image Processing Transformer} 

\titlerunning{LIPT}
\authorrunning{Junbo Qiao, Wei Li et al.}
\author{Junbo Qiao{$^{1,\dag}$}\and
Wei Li{$^{2,\dag}$} \and
Haizhen Xie{$^{2}$} \and Hanting Chen{$^{2}$} \and Yunshuai Zhou{$^{1}$} \and Zhijun Tu{$^{2}$} \and  Jie Hu{$^{2}$} \and Shaohui Lin{$^{1,*}$}}


\institute{School of Computer Science and Technology, East China Normal University  \and
Huawei Noah‘s Ark Lab\\
}
\maketitle
\footnotetext[2]{These authors contributed equally to this work.} 
\footnotetext[1]{Corresponding author (e-mail: shlin@cs.ecnu.edu.cn)}

\begin{abstract}
 Transformer is leading a trend in the field of image processing.
Despite the great success that existing lightweight image processing transformers have achieved, they are tailored to FLOPs or parameters reduction, rather than practical inference acceleration. In this paper, we present a latency-aware image processing transformer, termed LIPT. We devise the low-latency proportion LIPT block that substitutes memory-intensive operators with the combination of self-attention and convolutions to achieve practical speedup. Specifically, we propose a novel non-volatile sparse masking self-attention (NVSM-SA) that utilizes a pre-computing sparse mask to capture contextual information from a larger window with no extra computation overload. Besides, a high-frequency reparameterization module (HRM) is proposed to make LIPT block reparameterization friendly, which improves the model's detail reconstruction capability. Extensive experiments on multiple image processing tasks (e.g., image super-resolution (SR), JPEG artifact reduction, and image denoising) demonstrate the superiority of LIPT on both latency and PSNR. LIPT achieves real-time GPU inference with state-of-the-art performance on multiple image SR benchmarks.
\end{abstract}

\section{Introduction}
\label{sec:intro}

Image processing is one of the key components in the low-level vision system, whose results largely affect the high-level image and scene understanding. In recent years, deep convolutional neural networks (CNNs) ~\cite{dong2014learning,zhang2018image,dai2019second,zhang2018ffdnet,wang2020model} have been applied to solve the image degradation for image processing, including but not limited to image super-resolution (SR), denoising, deblurring, deraining, JPEG artifact reduction. Although CNN-based methods have achieved impressive performance over traditional methods, they lack the ability to model long-range pixel dependencies and cannot well model the relations among pixels to adapt to the input content with the usage of static convolutional weights.
\begin{figure}[t]
  \centering
  \begin{minipage}{0.46\linewidth}
		\centerline{\includegraphics[width=1.0\linewidth]{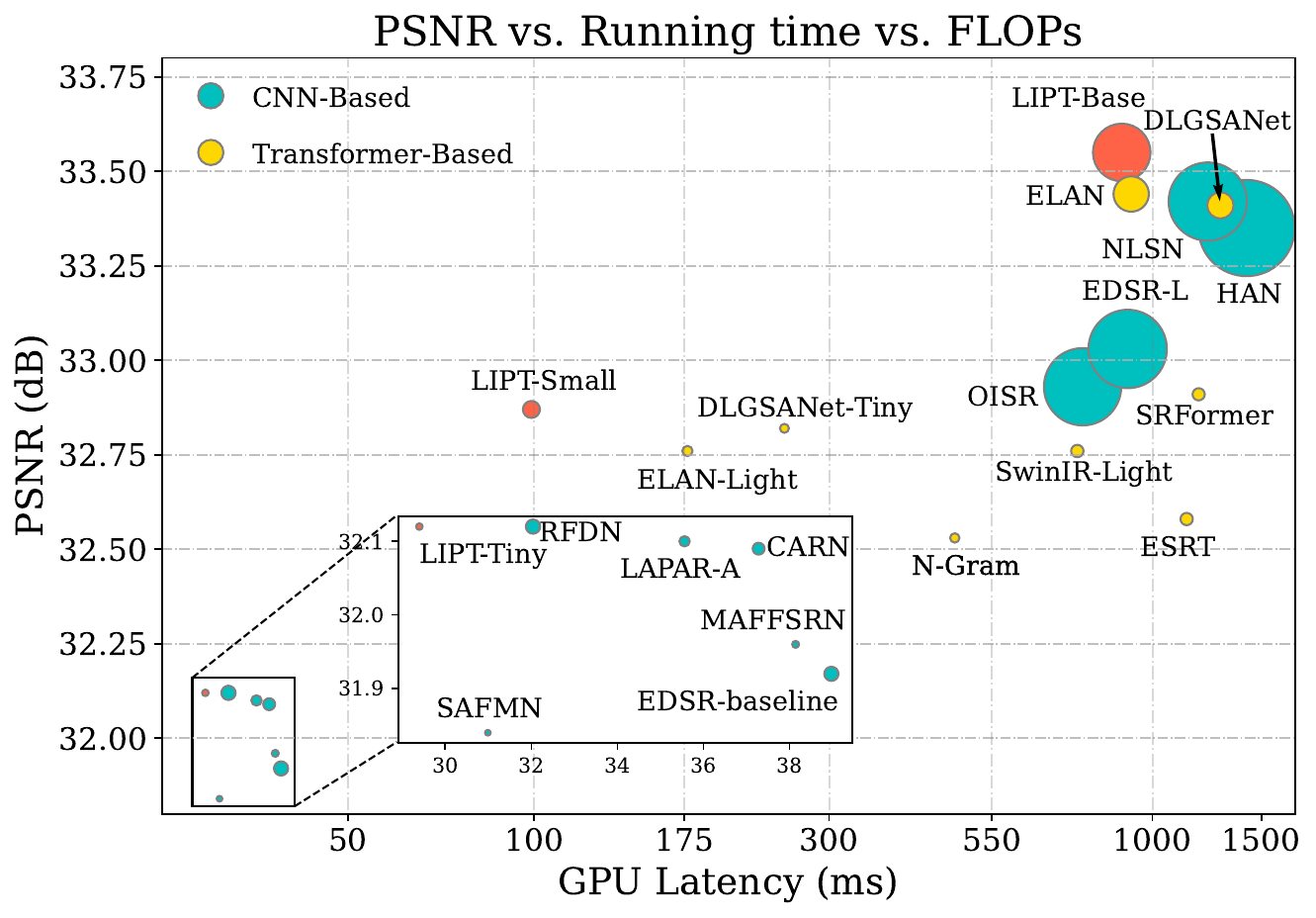}}
		\centerline{\footnotesize (a)}
	\end{minipage}
	\begin{minipage}{0.53\linewidth}
        \vspace{1em}
		\centerline{\includegraphics[width=1.0\linewidth]{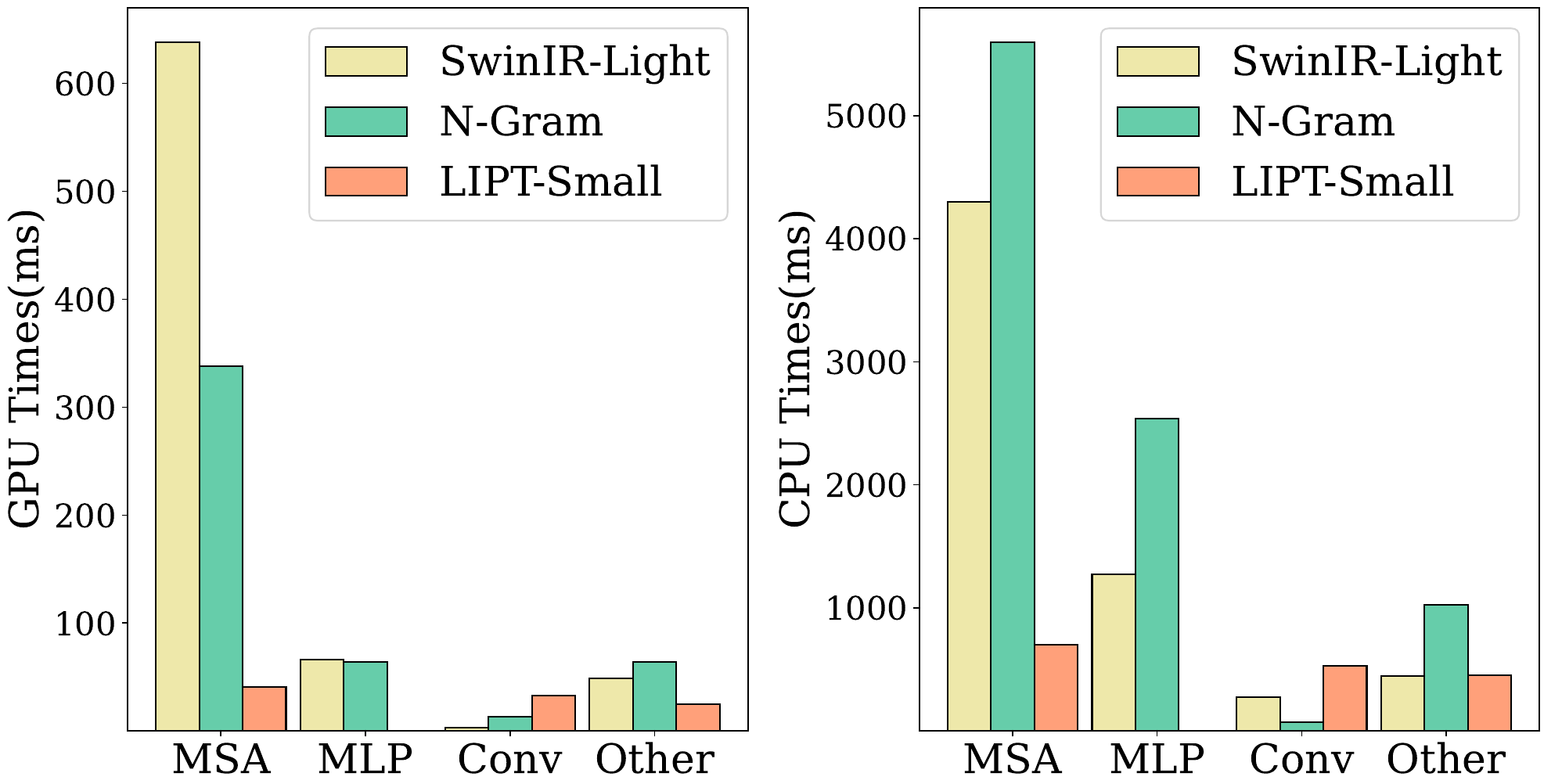}}
        \vspace{0.5em}
		\centerline{\footnotesize (b)}
	\end{minipage}
   \caption{(a) Performance on Urban100 for $\times 2$ SR. The larger circles present larger computation costs on FLOPs, which however is not directly proportional to the practical running latency. (b) The inference time of MSA, MLP and Conv in the low-level Transformers. The total inference time on GPU (CPU) of SwinIR-Light, N-Gram and LIPT-Small are 756ms (6.3S), 479ms (9.2S) and 99ms (2.8S), respectively.}
   \vspace{-2em}
   \label{fig:result}
\end{figure}

Recently, Transformers~\cite{dosovitskiy2020image,liu2021swin,ramachandran2019stand,touvron2021training}  have introduced the self-attention mechanism to model long-range dependencies, which has been applied to low-level vision tasks.
For example, IPT~\cite{chen2021pre} constructed a pre-trained model using the conventional Transformer structure for image processing and is learned by contrastive learning on ImageNet data. 
SwinIR~\cite{liang2021swinir} introduced the local spatial window scheme from Swin Transformer~\cite{liu2021swin} for image restoration. Later, Chen \emph{et al.}~\cite{chen2023dual} found the strong modeling ability of both local spatial window and channel-wise self-attention, and further proposed a dual aggregation Transformer for SR. 
Although these Transformers outperform CNNs in several low-level vision tasks, the conventional self-attention is computation-intensive to hamper fast inference, especially for high-resolution image reconstruction.

\begin{figure*}[t]
  \centering
   \includegraphics[width=0.9\linewidth]{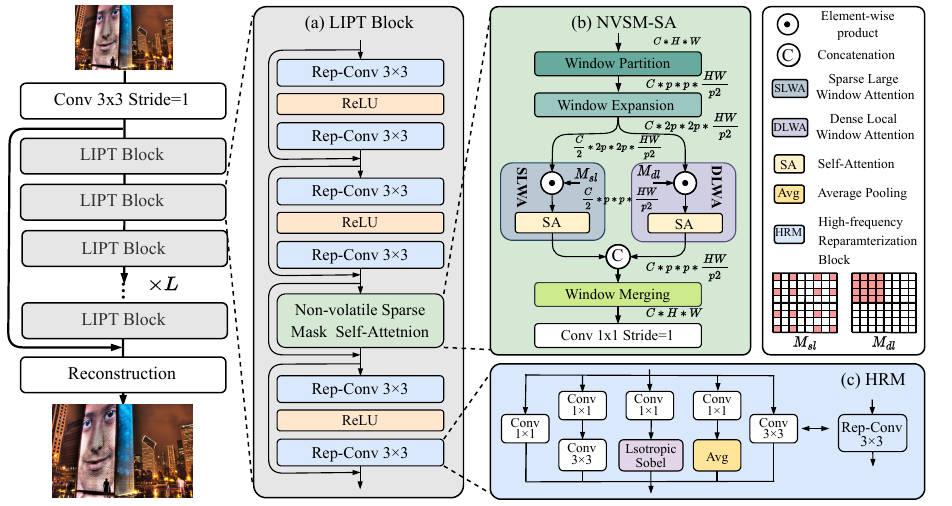}
   \vspace{-.5em}
   \caption{Illustration of the proposed LIPT (taking the SR task and expansion window size $s=2$ for example). (a) LIPT block with low-latency proportion MSA-Conv. (b) NVSM-SA is proposed to capture contextual information from the larger window with no extra computation cost, where SLWA and DLWA denote sparse large and dense local window attentions, respectively. (c) HRM is developed to improve the model's detail reconstruction capability.}
   \vspace{-2em}
   \label{fig:str}
\end{figure*}

To reduce the computation overhead, various efficient Transformers have been proposed to redesign the self-attention for image processing. For example, 
Restormer~\cite{zamir2022restormer} and DLGSANet~\cite{li2023dlgsanet} proposed transposed attentions to calculate the self-attention on the channel dimension rather than the spatial dimension, because of the cheap computation based on channel-wise self-attention. GRL~\cite{li2023efficient} proposed an anchored stripe self-attention mechanism to use the low-rankness of the original attention map, which reduces the computational complexity of self-attention. N-Gram~\cite{choi2023n} introduced an asymmetric U-Net architecture, which reduces the feature size through downsampling, as well as the number of decoder stages. ESRT~\cite{zhisheng2021efficient} achieves a notable reduction in memory usage by computing attention maps in a group manner.
Unfortunately, these methods aim to reduce theoretical FLOPs, whose metrics are not directly proportional to practical speedup. For example, as shown in Fig.~\ref{fig:result} (a),  Despite OISR~\cite{he2019ode} requiring more 40$\times$ FLOPs than SRFormer~\cite{zhou2023srformer}, its inference speed is faster with 23\% time reduction. Moreover, these Transformers retain a large gap to the real-time image processing ($<$33ms per image). Thus, \emph{it is of great interest to develop a latency-aware transformer architecture for practically fast inference while obtaining high-quality image reconstruction.}

In this paper, we propose a novel \emph{Latency-aware Image Processing Transformer} architecture, termed LIPT, which obtains high-quality image reconstruction with significantly practical speedup. Fig.~\ref{fig:str} depicts the components of the proposed LIPT, whose key component lies in the new LIPT blocks based on two-level designs. First,
we observe that Multi-head Self-Attention (MSA) and Multi-Layer Perception (MLP) occupy two main practical computation overhead, as shown in Fig.~\ref{fig:result} (b). Thus, at the block level, we merge two transformer blocks into one block by replacing one memory-intensive MSA with one convolution block, as well as the same processing for MLP. The simple block design significantly improves the running efficiency, while reducing the representation ability for image reconstruction.
To address this problem, we propose two key modules within the LIPT block, \emph{Non-Volatile Sparse Masking Self Attention} (NVSM-SA) and \emph{High-frequency Reparameterization Module} (HRM). NVSM-SA aims to expand the receptive field of window attention by the combination of sparse large-window attention and dense local-window attention while maintaining the same computational and memory complexity as the local window. HRM extracts high-frequency information by fusing multi-branch convolutions and high-frequency feature extraction operator (\emph{e.g.}, Sobel operator) for better reconstructing the edge and texture. After training, HRM can be parameterized to a vanilla convolution (denoted by Rep-Conv) in the LIPT block without incurring additional inference costs. 

We summarize our main contributions as follows: 
\begin{itemize}
    \item We design a novel latency-aware LIPT architecture for fast inference while obtaining high-quality image reconstruction. The LIPT block significantly reduces the memory access intensive operations, consisting of low-latency MSA-Conv modules. 
    \item  We define a non-volatile sparse masking (NVSM) operation for self-attention, which guides us to expand the receptive field by the sparse sampling on the large window. Moreover, NVSM-SA improves the reconstruction performance while keeping the same computation as the small local window attention. 
    HRM is proposed to improve the reconstruction ability on edge and texture without incurring additional 
 computation costs for inference.
    \item Our LIPT outperforms lightweight Transformers on multiple image processing tasks in terms of latency and PSNR. In particular, our LIPT achieves real-time image SR, whose PSNR is comparable to or even higher than those of lightweight CNN models.
\end{itemize}

\section{Related Work}
\label{sec:related}

\subsection{CNN-based image processing models}
The CNN-based approach has achieved impressive performance for image restoration. SRCNN~\cite{dong2014learning}, ARCNN~\cite{dong2015compression} and DnCNN~\cite{zhang2017beyond} first introduce CNN models for image super-resolution, JEPG compression artifact reduction, and image denoising, respectively. After these three methods, various CNN-based methods improve the model presentation ability by increasing network depth. VDSR ~\cite{kim2016accurate} improved the network's performance by incorporating VGG-19 and residual learning. By eliminating unnecessary BatchNormal layers, EDSR~\cite{lim2017enhanced} proposed an enhanced residual block and achieved a substantial enhancement in PSNR. Subsequently, spatial and channel attention mechanisms~\cite{zhang2018image,dai2019second,mei2021image,niu2020single,li2022blueprint} were utilized to further improve performance. As most CNN-based methods struggle to capture global dependencies, Transformers with self-attention modules were proposed for image restoration, which has been a growing research focus. 

\subsection{Efficient Image Processing  Transformers}
High-level vision tasks have witnessed remarkable results with the introduction of vision Transformer~\cite{dosovitskiy2020image,liu2021swin,carion2020end,zheng2021rethinking}, which employed the self-attention mechanism to effectively capture dependencies across information. Afterwards, various works~\cite{liang2021swinir,zamir2022restormer,wang2022uformer,chen2023recursive,chen2024hierarchical} have proposed numerous attention mechanisms for image restoration. SwinIR~\cite{liang2021swinir} built upon the Swin Transformer~\cite{liu2021swin} and employed the window-based attention mechanism to partition the image, where self-attention is computed separately within each window. Uformer~\cite{wang2022uformer} adopted an 8×8 local window approach and incorporates the UNet architecture to effectively capture extensive global dependencies.
Recently, various efficient Transformers have been proposed to reduce the quadratic computational complexity in the self-attention module. For example,
Restormer~\cite{zamir2022restormer} computed cross-covariance across channel dimensions instead of spatial dimensions to reduce self-attention complexity in the image resolution. CAT~\cite{chen2022cross} and ART~\cite{zhang2022accurate} designed rectangle and sparse self-attention respectively to improve the model performance. GRL~\cite{li2023efficient} proposed an anchored stripe self-attention mechanism, which can capture image structural features beyond the regional scope. Despite the remarkable results they achieve with the FLOPs reduction, they are still difficult to obtain practical acceleration.

For the real inference speedup, ELAN~\cite{zhang2022efficient} eliminated the redundant operation of SwinIR and proposed group multi-head self-attention, which shares an attention map to accelerate the calculation. DLGSANet\cite{li2023dlgsanet} employed multi-head dynamic local self-attention to further reduce computation and adopted channel-wise global self-attention with RELU instead of Softmax. 
Different from these methods, our LIPT focuses on reducing memory-intensive operations to design a new LIPT block from the perspective of low-latency computation, within which we propose non-volatile sparse masks for masking self-attention without incurring additional inference computation cost.

\subsection{Model Reparameterization}
Numerous studies have demonstrated the effectiveness of reparameterization techniques~\cite{ding2021repvgg,arora2018optimization,ding2019acnet,ding2021diverse,zagoruyko2017diracnets} on image classification. 
DiracNet~\cite{zagoruyko2017diracnets} trained a plain CNN without shortcuts and achieved comparable results to ResNet. 
ACNet~\cite{ding2019acnet} employed asymmetric structural re-parameterized convolution to enhance the normal convolution. 
RepVGG~\cite{ding2021repvgg} employed multi-branch blocks consisting of $1 \times 1$ convolution and $3 \times 3$ convolution to further improve the performance of VGG model. 
Further,
diverse branch block~\cite{ding2021diverse} leveraged different scales and complexities of features to enhance the representative capacity of the normal convolution. However, previous reparameterization techniques primarily focus on the CNN-based models, and direct application for Transformer-based structures may be problematic due to the insufficient ability of high-quality reconstruction caused by different architectures. To leverage the revival of convolution into the LIPT block, we designed the High-frequency Reparameterization (HRM) specifically to enhance the model's capability for the extraction of high-frequency information in low-level vision tasks.


\section{Proposed Method}
\label{sec3}

\subsection{Overall Pipeline of LIPT}
As shown in Fig.~\ref{fig:str}, LIPT architecture comprises three parts: shallow feature extraction, deep feature extraction based on the LIPT block, and reconstruction. Let $I_{d}\in\mathbb{R}^{C_{in} \times H \times W }$ be a degraded image, where $C_{in}$, $H$, and $W$ in are the channel, height and width of the input, respectively. 
We employ a single convolution layer to extract the shallow feature $F_{s} \in \mathbb{R}^{C \times H \times W }$ in the  shallow feature extraction $H_{SF}$:
\begin{equation}
F_{s}=H_{SF}(I_{d}),
\end{equation}
where $C$ is the embedding channel dimension. Subsequently, $F_{s}$ goes through cascaded LIPT blocks denoted by $H_{DF}$ to extract the deep feature $F_{d} \in \mathbb{R}^{C \times H \times W }$:
\begin{equation}
F_{d}=H_{DF}(F_{s}).
\end{equation}
By using $F_{s}$ and $F_{d}$ as inputs, the high-quality image $I_r$ is generated in the reconstruction process denoted by $H_{R}$ as:
\begin{equation}
I_{r}=H_{R}(F_{s}+F_{d}),
\end{equation}
where $H_{R}$ involves a single 3$\times$3 convolution followed by a pixel shuffle operation for SR. For other tasks 
without upsampling, $H_{R}$ utilizes a single convolution. 

For image SR, we optimize the parameters $\theta$ of LIPT models by minimizing the pixel-wise L1 loss:
\begin{equation}
\mathcal{L}(\theta)=\frac{1}{N}\sum\limits_{i=1}^{N}\left\|I_{r}^i-I_{gt}^i\right\|_1,
\end{equation}
where $N$ and $I_{gt}^{i}$ denote the number of training pairs and the ground truth of the $i$-th image, respectively.

For image denoising, JPEG compression artifact reduction, we use Charbonnier loss~\cite{charbonnier1994two} as the loss function:
\begin{equation}
\mathcal{L}(\theta)=\frac{1}{N}\sum\limits_{i=1}^{N}\sqrt{\left\|I_{r}^{i}-I_{gt}^{i}\right\|^2+\epsilon^2},
\end{equation}
where $\epsilon$ is a constant that is empirically set to $10^{-3}$ followed by SwinIR~\cite{liang2021swinir}.

\subsection{LIPT Block Design}
\label{sec3_1}
To better design the low-latency Transformer block, we first investigate the practical computation time of the previous low-level Transformer architectures (\emph{e.g.}, SwinIR~\cite{liang2021swinir}, N-Gram~\cite{choi2023n}), \emph{w.r.t.} different modules.
As shown in Fig.~\ref{fig:result} (b), we observe two modules with the largest computation overhead are Multi-head Self-Attention (MSA) and Multi-Layer Perception (MLP) both in SwinIR and N-Gram across different GPU and CPU  computing platforms. Thus, our block design lies in two following reasons: (1) MSA has memory-intensive operations like tensor reshaping, and matrix product with quadratic computational complexity in self-attention, which significantly increases the computation time. One feasible way is to directly remove it or replace it with a cheap operation.
(2) Inspired by~\cite{sun2023safmn}, convolution with $3\times3$ kernels benefits from recovering intricate details~\cite{li2023feature} and friendly hardware running, such that MLP can be effectively replaced with the convolution block with two Conv layers. To this end, we merge two traditional blocks into one new block (in Fig.~\ref{fig:str}(a)) by replacing one MSA with one convolution block, as well as the same processing for MLP. 

After the above block design, we can significantly reduce the running time at the same depth. Note that the shortcut path is also applied to each convolution block and MSA.  
Thus, the proposed LIPT block can be formulated as:
\begin{footnotesize}
\begin{equation}
I^{(l)}=G_{CB}\left(G_{MSA}\left(G_{CB}\left(G_{CB}\left(I^{(l-1)}\right)\right)\right)\right), l=1,\cdots,L,
\label{stru}
\end{equation}
\end{footnotesize}
where $G_{MSA}$ and $G_{CB}$ denote one MSA and a convolution block, respectively. 
The first block takes the shallow feature $F_s$ as an input (\emph{i.e.}, $I^{(0)}=F_s$).  

With the reduction in the number of MSA, it decreases the model's capability to capture distant spatial relationships. To address this issue, we propose Non-volatile Sparse Masking Self Attention (NVSM-SA) to achieve a wider receptive field and effectively model long-range spatial relationships with no extra computation overhead. Meanwhile, the revival of convolutions in $G_{CB}$ makes LIPT block reparameterization friendly, which motivates us to design the High-frequency Reparameterization Module (HRM) for better reconstructing the edge and texture without incurring additional inference costs. 
Thus, Eq.~\ref{stru} can be reformulated as:
\begin{small}
\begin{equation}
I^{(l)}=G_{HRM}\left(G_{N-SA}\left(G_{HRM}\left(G_{HRM}\left(I^{(l-1)}\right)\right)\right)\right),
\label{stru_2}
\end{equation}
\end{small}
where $G_{N-SA}$ and $G_{HRM}$ denote NVSM-SA and HRM, which will be introduced in Sec.~\ref{sec:MSA} and Sec.~\ref{sec:HRM}, respectively. We remove the superscript $(l)$ for convenient discussion due to the same LIPT block at different depths.

\begin{algorithm}[t]
\small
\caption{Non-volatility Sampling Rule.}
\renewcommand{\algorithmicrequire}{\textbf{Input:}} 
\renewcommand{\algorithmicensure}{\textbf{Output:}}
\begin{algorithmic}[1]
\REQUIRE 
Input feature $I\in \mathbb{R}^{C \times p \times p \times H/{p} \times W/{p}}$, fixed mask $M$, local window size $p$, extend scale $s \in N^{+}$ and $0<p\leq sp\leq\min (H, W)$.  
\ENSURE 
Decide the mask $M$ and its non-volatility drop rate $\beta$. \\
\STATE
Initialize $I_e = G_{we}(I)$, $I_{e}\in\mathbb{R}^{C \times sp \times sp\times H/{p} \times W/{p}}$.
\STATE
\textbf{Catch} $\forall F_{i,j}\in I$, $F_{i,j}\in\mathbb{R}^{C \times p \times p}$,$F_{i,j}^e\in I_e$, $F_{i,j}^e\in\mathbb{R}^{C \times sp \times sp}$, $i \in(1, H/{p})$, $j \in(1, W/{p})$.
\STATE
\textbf{Do} $F_{i,j}^{*} =  \bigcup_{k=i-s+1}^{i}\bigcup_{m=j-s+1}^{j} M \odot F_{k,m}^e$, and
Compute $\beta = 1-\frac{\left\|F_{i,j}^{*} \cap F_{i,j}\right\|}{\left\|F_{i,j}\right\|}$. \\
\STATE
 \textbf{If}   $\beta == 0$, $M$ Satisfies Non-volatile Sampling rule. 
\end{algorithmic}
\label{alg1}
\end{algorithm}

\subsection{Non-volatile Sparse Masking Self-Attention}
\label{sec:MSA}

As shown in Fig~\ref{fig:str} (b), NVSM-SA enlarges the receptive field by extending the original window, subsequently extracting local and global features within the expanded window using a non-volatile local-density mask and a non-volatile global-sparsity mask. After that, the self-attention is computed in both local and global features generated by Dense Local Window Attention (DLWA) and Sparse Large Window Attention (SLWA), respectively. 

Given an input feature $I \in \mathbb{R}^{C \times H \times W }$, the entire NVSM-SA module (\emph{i.e.}, $G_{N-SA}$ in Eq.~\ref{stru_2}) can be formulated as:
\begin{small}
\begin{equation}
\label{eq_nvsm}
\begin{split}
    & I_p = G_{wp}(I), I_e = G_{we}(I_p), [I_1, I_2] = G_{s}(I_e),\\
    & I_g = G_{slwa}(I_1, M_{sl}), I_l = G_{dlwa}(I_2, M_{dl}), \\
    & I_f = I+\text{Conv1}\Big(G_{me}\big (Concat(I_g, I_l)\big)\Big),
\end{split}
\end{equation}
\end{small}
where $I_p\in \mathbb{R}^{C \times p \times p \times {HW}/{p^2}}$ is the feature with $p\times p$ tokens via window partition $G_{wp}$ on $I$ in a non-overlapping manner, $I_e\in \mathbb{R}^{C \times sp \times sp \times {HW}/{p^2}}$ is the larger window feature via window expansion $G_{we}$ with a expansion size $s$ and padding (details shown in Appendix), $G_{s}$ is a split operation in the channel dimension to split $I_e$ into two-group features, $I_1$ and $I_2$ both with dimension of $\frac{C}{2} \times sp \times sp \times {HW}/{p^2}$. After that, the global feature $I_g$ and local feature $I_l$ are respectively generated by SLWA on $I_1$ and DLWA on $I_2$ through predefined non-volatile global-sparsity mask $M_{sl}$ and local-density mask $M_{dl}$, which satisfies the Non-volatility Sampling Rule presented in Alg.~\ref{alg1}. The detailed computation of $G_{slwa}$ and $G_{dlwa}$ will be introduced in the next paragraph. Finally, the output feature $I_f\in \mathbb{R}^{C \times H \times W }$ is generated by the concatenation of $I_g$ and $I_l$ in the channel dimension, window merging $G_{me}$, $1\times1$ convolution ($\text{conv1}$), and the addition to the input $I$. 

\begin{figure}[t]
	\centering
	\centering{\includegraphics[width=0.9\linewidth]{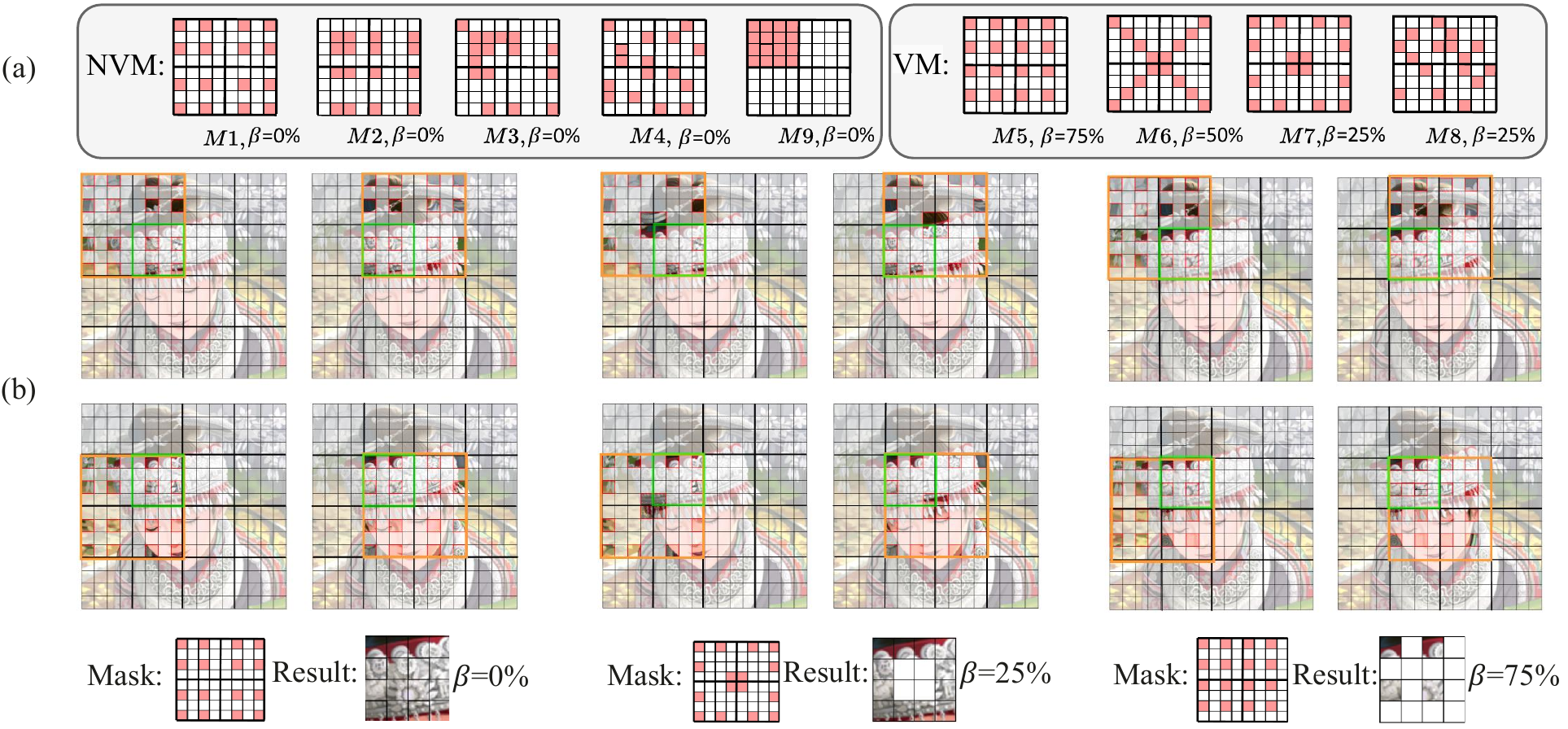}}
        \vspace{-0.5em}
	\caption{(a) Illustration of different non-volatile/volatile sampling masks (NVM/VM). $\beta$ denotes the non-volatility drop rate. (b) Sampling process of three masks, where the original window size $p$ and expansion window size $s$ are set to 4 and 2, respectively.}
	\vspace{-1.5em}
	\label{fig:non-volatile}
\end{figure}

To achieve faster inference speed, we discard the learnable masks~\cite{lin2017espace} and adopt the fixed masks. Note that both $M_{sl}$ and $M_{dl}$ (\emph{a.k.a.} binary masks) only select $p\times p$ feature points from $sp\times sp$ feature points according to their sampling location, while the usage of $M_{sl}$ expands the window ERF without incurring additional computation overhead, compared to the local window attention with size ($p\times p$) directly on $I$. $G_{dlwa}$ with the mask $M_{dl}$ is equivalent to the local window attention, which cannot be removed due to the rich information around the dense area (see Sec.~\ref{sec:aba}). In particular, we can reformulate $G_{slwa}$ and $G_{dlwa}$ in Eq.~\ref{eq_nvsm} by using traditional window self attention $G_{SA}$ to generate $I_g$ and $I_l$: 
\begin{small}
\begin{equation}
\label{eq_mask}
\begin{split}
    & I_g=[G_{SA}(I_1^{1}\odot M_{sl}), \cdots, G_{SA}(I_1^{HW/{p^2}}\odot M_{sl})], \\
    & I_l=[G_{SA}(I_2^{1}\odot M_{dl}), \cdots, G_{SA}(I_2^{HW/{p^2}}\odot M_{dl})], \\
\end{split}
\end{equation}
\end{small}
where [$\cdot$] and $\odot$ represnet concatenation operator and element-wise product, respectively. $I^i, i=1,\cdots, HW/{p^2}$ denotes the $i$-th spatial window in $I$.
%


\textbf{Non-volatile sampling masks.} 
Alg.~\ref{alg1} presents our sampling rule, whose main idea is to use the mask $M$ to sample $s^2$ windows from the size of $sp \times sp$ and spatially stitch the sampling results. If the result features equal to the original window feature, $M$ is a non-volatile mask. The sampling process is also depicted in Fig.~\ref{fig:non-volatile} (b). We also compute its non-volatility drop rate $\beta$, the larger values generate relatively lower reconstruction results, which is validated in experiments (Fig.~\ref{fig:non-volatile_result}). The sampling mask is non-volatile if $\beta$ is equal to 0 and volatile otherwise. 
%
Thus, according to Alg.~\ref{alg1}, we design multiple non-volatile masks and also give some volatile ones, as shown in Fig.~\ref{fig:non-volatile} (a). In this paper, we apply the $M1$ to $M_{sl}$, since it is the most efficient to $\odot$ operation in Eq.~\ref{eq_mask}. In addition, we apply the $M9$ non-volatile mask to $M_{dl}$.

To better understand the non-volatility drop rate $\beta$ in the non-volatile sampling mask (NVSM), we take the original window size of $p=4$ and the expanded size of $s=2$ as an example to show the sampling with three masks. 
In Fig.~\ref{fig:non-volatile} (b), the green box and orange box represent the original window and the extended windows (+ original), which are composed of $4\times4$ window pixels and $8\times8$ window pixels respectively. According to the non-volatile sampling rule in Alg.~\ref{alg1}, we first search the expanded windows consisting of the original small window and then employ spatial stitching to obtain the sampling features \emph{w.r.t.} the designed sampling masks. Finally, we achieve three sampling masks $M1$, $M7$ and $M5$ to obtain different non-volatility drop rates $\beta$ of 0\%, 25\% and 75\%, respectively. It is worth noting that the sampling method with the mask $M5$ is used in ART~\cite{zhang2022accurate}, which permanently drops 75\% of the feature information.

\subsection{High-frequency Reparameterization Module}
\label{sec:HRM}
Reparameterization~\cite{ding2021repvgg,ding2019acnet,ding2021diverse} has shown a strong ability to improve the feature presentation of CNN models. Different from the reparameterization module design of pure convolutional structure,  we revive the reparameterization in our LIPT architecture, and 
Fig~\ref{fig:str} (c) shows the internal branch structure of HRM. We employ various branch structures to enhance the network's representation capabilities. In particular, we design a high-frequency information extraction branch (\emph{e.g.}, isotropic Sobel) to extract more rich high-frequency information to improve the reconstruction performance. The isotropic Sobel operation is introduced in the Appendix. 

During training, the output of HRM $I_{hrm}$ is generated by taking the NVSM-SA output $I_f$ in Eq.~\ref{eq_nvsm} as input and fusing the outputs of diverse branches. Thus, this computation process of $G_{HRM}$ in Eq.~\ref{stru_2} can be formulated as:
\begin{equation}
\label{eq_hrm_o}
I_{hrm}=G_{HRM}(I_f)=I_f+G_{b}\circ\text{ReLU}\circ G_{b}(I_f),
\end{equation}
where $G_b$ is a multiple-branch operation formulated as:
\begin{footnotesize}
\begin{equation}
\label{eq_hrm}
\begin{split}
 G_b(I_f) & = \text{Conv1}(I_f)+\text{Conv3}\circ\text{Conv1}(I_f) \\
 & +G_{is}\circ\text{Conv1}(I_f)+G_{avg}\circ\text{Conv1}(I_f)+\text{Conv3}(I_f),
\end{split}
\end{equation}
\end{footnotesize}
where $G_{is}$ and $G_{avg}$ are isotropic Sobel and average pooling operations, respectively.
For inference, following ~\cite{ding2021diverse}, the output $G_b(I_f)$ is computed in a simplified $3\times3$ convolution, which significantly reduces computation cost.


\section{Experiments}


\subsection{Experimental Setups}
\textbf{Datasets and evaluation.} The proposed LIPT is evaluated on three lovel-level vision tasks, super-resolution (SR), JPEG compression artifact reduction (CAR) and denoising. For SR, we follow ELAN~\cite{zhang2022efficient} and DLGSANet~\cite{li2023dlgsanet}, by evaluating performance on three SR scales ($\times 2$, $\times 3$, $\times 4$). DIV2K~\cite{timofte2017ntire} dataset is used for training, and evaluate our method on the widely-used test datasets, including Set5~\cite{bevilacqua2012low}, Set14\cite{zeyde2012single}, BSD100~\cite{martin2001database}, Urban100\cite{huang2015single}, and Manga109~\cite{matsui2017sketch}. 
For image denoising and CAR, we utilize DIV2K~\cite{timofte2017ntire}, Flickr2~\cite{lim2017enhanced}, BSD500~\cite{arbelaez2010contour}, and WED~\cite{ma2016waterloo} datasets for training. CAR results are tested on classic5~\cite{foi2007pointwise} and LIVE1~\cite{sheikh2006statistical} at different JPEG quality factors of 10, 30, and 40.  We use BSD68~\cite{martin2001database}, Kodak24~\cite{franzen1999kodak}, McMaster~\cite{zhang2011color} and Urban100~\cite{huang2015single} as test datasets for image denoising.
For performance evaluation, we calculate PSNR and SSIM~\cite{wang2004image} on the Y channel, average inference time (5 runs) on one NVIDIA V100, and a 52-thread
Intel Xeon Gold 6278C CPU with the output resolution of $1280 \times 720$, as well as parameters and FLOPs.

\begin{table}[t]
	\centering
	\scriptsize       
	\caption{Quantative comparison of lightweight SR models on five benchmarks. In all tables and figures, we measure the inference latency by upscaling input images to the resolution of $1280 \times 720$ at all scales. The {\color[HTML]{FF0000}best} and {\color[HTML]{00B0F0}second-best} results for Transformers are marked in red and blue colors. At the same scale, the models above and below the dotted line are CNN-based and Transformer-based, respectively.}
        \vspace{-1em}
        \resizebox{.99\columnwidth}{!}{
        \label{tab:light}
	\begin{tabular}{|c|c|c|c|c|c|c|c|c|c|c|}
		\hline
		\multirow{2}*{\textbf{Scale}}        & \multirow{2}*{\textbf{Model}}       & \multicolumn{1}{c|}{\textbf{Latency}}    &\multicolumn{1}{c|}{\textbf{Latency}}    & \multicolumn{1}{c|}{\textbf{Params}} & \multicolumn{1}{c|}{\textbf{FLOPs}} & \multirow{2}*{\textbf{Set5}}                       & \multirow{2}*{\textbf{Set14}}                      & \multirow{2}*{\textbf{BSD100}}                       & \multirow{2}*{\textbf{Urban100}}                       & \multirow{2}*{\textbf{Manga109}} \\
  & & GPU (ms) & CPU (s) & (M)& (G) & & & & & \\ \hline
		& CARN~\cite{ahn2018fast}                 & 37             & 1.45          & 1.59                & 223              & \multicolumn{1}{c|}{37.76/0.9590}                        & \multicolumn{1}{c|}{33.52/0.9166}                        & \multicolumn{1}{c|}{32.09/0.8978}                        & \multicolumn{1}{c|}{31.92/0.9256}                        & 38.36/0.9765                        \\
		& EDSR-baseline~\cite{lim2017enhanced}        & 39         &    1.03           & 1.37                & 316              & \multicolumn{1}{c|}{37.99/0.9604}                        & \multicolumn{1}{c|}{33.57/0.9175}                        & \multicolumn{1}{c|}{32.16/0.8994}                        & \multicolumn{1}{c|}{31.98/0.9272}                        & 38.54/0.9769                        \\
		& LAPAR-A~\cite{li2020lapar}              & 36             &  0.82         & 0.55                & 171                & \multicolumn{1}{c|}{38.01/0.9605}                        & \multicolumn{1}{c|}{33.62/0.9183}                        & \multicolumn{1}{c|}{32.19/0.8999}                        & \multicolumn{1}{c|}{32.10/0.9283}                        & 38.67/0.9772                        \\
		& MAFFSRN~\cite{muqeet2020multi}              & 38         &   1.22            & 0.41                & 77               & \multicolumn{1}{c|}{37.97/0.9603}                        & \multicolumn{1}{c|}{33.49/0.9170}                        & \multicolumn{1}{c|}{32.14/0.8994}                        & \multicolumn{1}{c|}{31.96/0.9268}                        & - / -                               \\
		& RFDN~\cite{liu2020residual}                 & 32         &      1.22         & 0.57                & 124                  & \multicolumn{1}{c|}{38.05/0.9606}                        & \multicolumn{1}{c|}{33.68/0.9184}                        & \multicolumn{1}{c|}{32.16/0.8994}                        & \multicolumn{1}{c|}{32.12/0.9278}                        & 38.88/0.9773                        \\ 
        & SAFMN~\cite{sun2023safmn}                & 31                &  0.97      & 0.23                & 52                 & 38.00/0.9605                        & 33.54/0.9177                        & 32.16/0.8995                        & 31.84/0.9256                        & 38.71/0.9771                        \\ \cdashline{2-11}[1pt/1pt]
		& ESRT~\cite{zhisheng2021efficient}                 & 1,136        &   -              & 0.67                & -                  & \multicolumn{1}{c|}{38.03/0.9600}                        & \multicolumn{1}{c|}{33.75/0.9184}                        & \multicolumn{1}{c|}{32.25/0.9001}                        & \multicolumn{1}{c|}{32.58/0.9318}                        & 39.12/0.9774                        \\
		& N-Gram~\cite{choi2023n}               & 479             &    9.23         & 1.01                & 140                & \multicolumn{1}{c|}{38.05/0.9610}                        & \multicolumn{1}{c|}{33.79/0.9199}                        & \multicolumn{1}{c|}{32.27/0.9008}                        & \multicolumn{1}{c|}{32.53/0.9324}                        & 38.97/0.9777                        \\
		& SwinIR-Light~\cite{liang2021swinir}         & 756            &   6.30           & 1.08                & 243                & 38.14/0.9611                        & 33.86/0.9206                        & {\color[HTML]{00B0F0} 32.31}/0.9012 & 32.76/0.9340                        & 39.12/0.9783                        \\
                      & ELAN-Light~\cite{zhang2022efficient}           & 177         &   12.93              & 0.62                & 168                & {\color[HTML]{00B0F0} 38.17}/0.9611 & {\color[HTML]{00B0F0} 33.94}/0.9207 & 32.30/0.9012                        & 32.76/0.9340                        & 39.11/0.9782                        \\
                      & SPIN~\cite{zhang2023lightweight}                 & 3,454        &  36.31               & 0.49                & 116                & {\color[HTML]{FF0000} 38.20/0.9615} & 33.90/{\color[HTML]{00B0F0}0.9015}                        &  {\color[HTML]{00B0F0}32.31/0.9015} & 32.79/0.9340                        & {\color[HTML]{00B0F0} 39.18}/{\color[HTML]{FF0000}0.9784} \\
                      & DLGSANet-Tiny~\cite{li2023dlgsanet}        & 254          &    -            & 0.57                & 128              & 38.16/{\color[HTML]{00B0F0}0.9611}                        & {\color[HTML]{00B0F0} 33.92}/0.9202 & 32.26/0.9007                        & {\color[HTML]{00B0F0} 32.82/0.9343} & 39.14/0.9777                        \\
                      & \textbf{LIPT-Tiny}  & {\color[HTML]{FF0000} 29} &{\color[HTML]{FF0000} 0.70 }& 0.36                & 69              & 38.03/0.9608                        & 33.61/0.9184                        & 32.20/0.9000                        & 32.12/0.9286                        & 38.63/0.9772                        \\
        \multirow{-14}{*}{×2} & \textbf{LIPT-Small} & {\color[HTML]{00B0F0} 99} &{\color[HTML]{00B0F0} 2.81}  & 2.33                & 455                & {\color[HTML]{FF0000} 38.20/0.9615} & {\color[HTML]{FF0000} 33.96/0.9217} & {\color[HTML]{FF0000} 32.36/0.9021} & {\color[HTML]{FF0000} 32.87/0.9355} & {\color[HTML]{FF0000} 39.21/0.9784} \\
        \hline
		& CARN~\cite{ahn2018fast}                 & 17           &   0.58           & 1.59                & 119              & \multicolumn{1}{c|}{34.29/0.9255}                        & \multicolumn{1}{c|}{30.29/0.8407}                        & \multicolumn{1}{c|}{29.06/0.8034}                        & \multicolumn{1}{c|}{28.06/0.8493}                        & 33.50/0.9440                        \\
		& EDSR-baseline~\cite{lim2017enhanced}        & 17           &   0.38           & 1.56                & 160              & \multicolumn{1}{c|}{34.37/0.9270}                        & \multicolumn{1}{c|}{30.28/0.8417}                        & \multicolumn{1}{c|}{29.09/0.8052}                        & \multicolumn{1}{c|}{28.15/0.8527}                        & 33.45/0.9439                        \\
		& LAPAR-A~\cite{li2020lapar}              & 16          &  0.44            & 0.54                & 114                & \multicolumn{1}{c|}{34.36/0.9267}                        & \multicolumn{1}{c|}{30.34/0.8421}                        & \multicolumn{1}{c|}{29.11/0.8054}                        & \multicolumn{1}{c|}{28.15/0.8523}                        & 33.51/0.9441                        \\
		& MAFFSRN~\cite{muqeet2020multi}              & 19           &  0.51           & 0.42                & 34               & \multicolumn{1}{c|}{34.32/0.9269}                        & \multicolumn{1}{c|}{30.35/0.8429}                        & \multicolumn{1}{c|}{29.09/0.8052}                        & \multicolumn{1}{c|}{28.13/0.8521}                        & - / -                               \\
		& RFDN~\cite{liu2020residual}                 & 15           &  0.43           & 0.54                & 56                  & \multicolumn{1}{c|}{34.41/0.9273}                        & \multicolumn{1}{c|}{30.34/0.8420}                        & \multicolumn{1}{c|}{29.09/0.8050}                        & \multicolumn{1}{c|}{28.21/0.8525}                        & 33.67/0.9449                        \\ 
        & SAFMN~\cite{sun2023safmn}                & 15           &    0.39         & 0.23                & 23                 & 34.34/0.9267                        & 30.33/0.8418                        & 29.08/0.8048                        & 27.95/0.8474                        & 33.52/0.9437                        \\ \cdashline{2-11}[1pt/1pt]
		& ESRT~\cite{zhisheng2021efficient}                 & 465           &    12.27        & 0.77                & -                  & \multicolumn{1}{c|}{34.42/0.9268}                        & \multicolumn{1}{c|}{30.43/0.8433}                        & \multicolumn{1}{c|}{29.15/0.8063}                        & \multicolumn{1}{c|}{28.46/0.8574}                        & 33.95/0.9455                        \\
		& N-Gram~\cite{choi2023n}               & 163          &    4.32          & 1.01                & 67               & \multicolumn{1}{c|}{34.52/0.9282}                        & \multicolumn{1}{c|}{30.53/0.8456}                        & \multicolumn{1}{c|}{29.19/0.8078}                        & \multicolumn{1}{c|}{28.52/0.8603}                        & 33.89/0.9470                        \\
		& SwinIR-Light~\cite{liang2021swinir}         & 295           &    2.87        & 0.89                & 87               & \multicolumn{1}{c|}{34.62/0.9289}                        & \multicolumn{1}{c|}{30.54/0.8463}                        & \multicolumn{1}{c|}{29.2/0.8082}                         & \multicolumn{1}{c|}{28.66/0.8624}                        & 33.98/0.9478                        \\
		& ELAN-Light~\cite{zhang2022efficient}           & 76           &  4.81            & 0.59                & 76               & 34.61/0.9288                        & 30.55/0.8463                        & 29.21/0.8081                        & 28.69/0.8624                        & 34.00/0.9478                        \\
                      & SPIN~\cite{zhang2023lightweight}                 & 1,307       &   9.56               & 0.57               & 61                 & {\color[HTML]{00B0F0} 34.65/0.9293} & {\color[HTML]{00B0F0} 30.57/0.8464} & {\color[HTML]{00B0F0} 29.23/0.8089} & {\color[HTML]{00B0F0} 28.71}/0.8627 & {\color[HTML]{FF0000} 34.24/0.9489} \\
                      & DLGSANet-Tiny~\cite{li2023dlgsanet}        & 123        &  -              & 0.57                & 57               & 34.63/0.9288                        & {\color[HTML]{00B0F0} 30.57}/0.8459 & 29.21/0.8083                        & 28.69/{\color[HTML]{00B0F0}0.8630}                        & 34.10/0.9480                        \\
                      & \textbf{LIPT-Tiny}  & {\color[HTML]{FF0000} 12}  & {\color[HTML]{FF0000}0.29}  & 0.35                & 31               & 34.38/0.9279                        & 30.32/0.8430                        & 29.12/0.8072                        & 28.24/0.8542                        & 33.55/0.9449                        \\
\multirow{-14}{*}{×3} & \textbf{LIPT-Small} & {\color[HTML]{00B0F0} 53}   & {\color[HTML]{00B0F0} 1.02}   & 2.80                 & 242              & {\color[HTML]{FF0000} 34.66/0.9300} & {\color[HTML]{FF0000} 30.59/0.8475} & {\color[HTML]{FF0000} 29.26/0.8112} & {\color[HTML]{FF0000} 28.75/0.8650} & {\color[HTML]{00B0F0} 34.18/0.9486} \\ \hline
		& CARN~\cite{ahn2018fast}                 & 11              &   0.33        & 1.59                & 91               & \multicolumn{1}{c|}{32.13/0.8937}                        & \multicolumn{1}{c|}{28.6/0.7806}                         & \multicolumn{1}{c|}{27.58/0.7349}                        & \multicolumn{1}{c|}{26.07/0.7837}                        & 30.47/0.9084                        \\
		& EDSR-baseline~\cite{lim2017enhanced}        & 11            & 0.23              & 1.52                & 114                & \multicolumn{1}{c|}{32.09/0.8938}                        & \multicolumn{1}{c|}{28.58/0.7813}                        & \multicolumn{1}{c|}{27.57/0.7357}                        & \multicolumn{1}{c|}{26.04/0.7849}                        & 30.35/0.9067                        \\
		& LAPAR-A~\cite{li2020lapar}              & 11          &   0.35           & 0.66                & 94                 & \multicolumn{1}{c|}{32.15/0.8944}                        & \multicolumn{1}{c|}{28.61/0.7818}                        & \multicolumn{1}{c|}{27.61/0.7366}                        & \multicolumn{1}{c|}{26.14/0.7871}                        & 30.42/0.9074                        \\
		& MAFFSRN~\cite{muqeet2020multi}              & 13           &   0.29          & 0.44                & 19               & \multicolumn{1}{c|}{32.18/0.8948}                        & \multicolumn{1}{c|}{28.58/0.7812}                        & \multicolumn{1}{c|}{27.57/0.7361}                        & \multicolumn{1}{c|}{26.04/0.7848}                        & - / -                               \\
		& RFDN~\cite{liu2020residual}                 & 9           &   0.25           & 0.55                & 32                  & \multicolumn{1}{c|}{32.24/0.8952}                        & \multicolumn{1}{c|}{28.61/0.7819}                        & \multicolumn{1}{c|}{27.57/0.7360}                        & \multicolumn{1}{c|}{26.11/0.7858}                        & 30.58/0.9089                        \\ 
        & SAFMN~\cite{sun2023safmn}                & 10           &   0.24          & 0.24                & 14                 & 32.18/0.8948                        & 28.60/0.7813                        & 27.58/0.7359                        & 25.97/0.7809                        & 30.43/0.9063                        \\ \cdashline{2-11}[1pt/1pt] 
		& ESRT~\cite{zhisheng2021efficient}                 & 286            &   3.78        & 0.75                & 64                  & \multicolumn{1}{c|}{32.19/0.8947}                        & \multicolumn{1}{c|}{28.69/0.7833}                        & \multicolumn{1}{c|}{27.69/0.7379}                        & \multicolumn{1}{c|}{26.39/0.7962}                        & 30.75/0.9100                        \\
		& N-Gram~\cite{choi2023n}               & 108          & 2.49           & 1.02                & 36               & \multicolumn{1}{c|}{32.33/0.8963}                        & \multicolumn{1}{c|}{28.78/0.7859}                        & \multicolumn{1}{c|}{27.66/0.7396}                        & \multicolumn{1}{c|}{26.45/0.7963}                        & 30.80/0.9128                        \\
		& SwinIR-Light~\cite{liang2021swinir}         & 179          &  1.87           & 0.90                 & 50               & \multicolumn{1}{c|}{32.44/0.8976}                        & \multicolumn{1}{c|}{28.77/0.7858}                        & \multicolumn{1}{c|}{27.69/0.7406}                        & \multicolumn{1}{c|}{26.47/0.7980}                        & 30.92/{\color[HTML]{00B0F0}0.9151}                        \\
		& ELAN-Light~\cite{zhang2022efficient}           & 45          &       3.09       & 0.60                 & 43               & \multicolumn{1}{c|}{32.43/0.8975}                        & \multicolumn{1}{c|}{28.78/0.7858}                        & \multicolumn{1}{c|}{27.69/0.7406}                        & \multicolumn{1}{c|}{26.54/0.7982}                        & 30.92/0.9150                        \\
		& SPIN~\cite{zhang2023lightweight}                 & 682           &     4.32          & 0.55                & 46               & {\color[HTML]{00B0F0} 32.48}/0.8983 & {\color[HTML]{00B0F0} 28.80/0.7862} & {\color[HTML]{00B0F0} 27.70/0.7415} & {\color[HTML]{00B0F0} 26.55}/0.7998 & {\color[HTML]{00B0F0} 30.98}/{\color[HTML]{FF0000}0.9156} \\
                      & DLGSANet-Tiny~\cite{li2023dlgsanet}        & 63            &   -         & 0.58                & 32                 & 32.46/{\color[HTML]{00B0F0}0.8984}                        & 28.79/0.7861                        & 27.70/0.7408                        & {\color[HTML]{00B0F0} 26.55/0.8002} & {\color[HTML]{00B0F0} 30.98}/0.9137 \\
                      & \textbf{LIPT-Tiny}  & {\color[HTML]{FF0000} 8}   & {\color[HTML]{FF0000}0.19}    & 0.36                & 18               & 32.19/0.8961                        & 28.60/0.7860                        & 27.59/0.7385                        & 26.16/0.7883                        & 30.44/0.9083                        \\
\multirow{-14}{*}{×4} & \textbf{LIPT-Small} & {\color[HTML]{00B0F0} 31}  & {\color[HTML]{00B0F0}0.58}  & 2.81                & 141              & {\color[HTML]{FF0000} 32.51/0.8990} & {\color[HTML]{FF0000} 28.81/0.7870} & {\color[HTML]{FF0000} 27.72/0.7427} & {\color[HTML]{FF0000} 26.57/0.8015} & 
{\color[HTML]{FF0000} 31.02}/0.9150 \\ \hline
	\end{tabular}}
 \vspace{-1em}
\end{table}

\textbf{Implementation details.}
We provide three LIPT versions with different GPU running times, including LIPT-Tiny, LIPT-Small and LIPT-Base. The detailed network setups of these three LIPT networks are presented in the Appendix. All networks are trained by Pytorch~\cite{paszke2017automatic} using 8 NVIDIA V100 GPUs. The window expansion size $s$ is default set to be 2. 
For SR, bicubic downsampling is used to generate training image pairs. During training, the batch size is set to 64, and each input image is randomly cropped to the patch size of $64 \times 64$. Both models are trained using the ADAM~\cite{kingma2014adam} optimizer with $\beta_1=0.9, \beta_2=0.999$, and $\epsilon=10^{-8}$  for 500 epochs. The initial learning rate is set to be $2 \times 10^{-4}$ with a multi-step scheduler every 500K iterations. Training hyper-parameters of other tasks are presented in the Appendix.
%
%

\begin{figure}[t]
        \centering
	\begin{minipage}{0.48\linewidth}
		\centerline{\includegraphics[width=\textwidth]{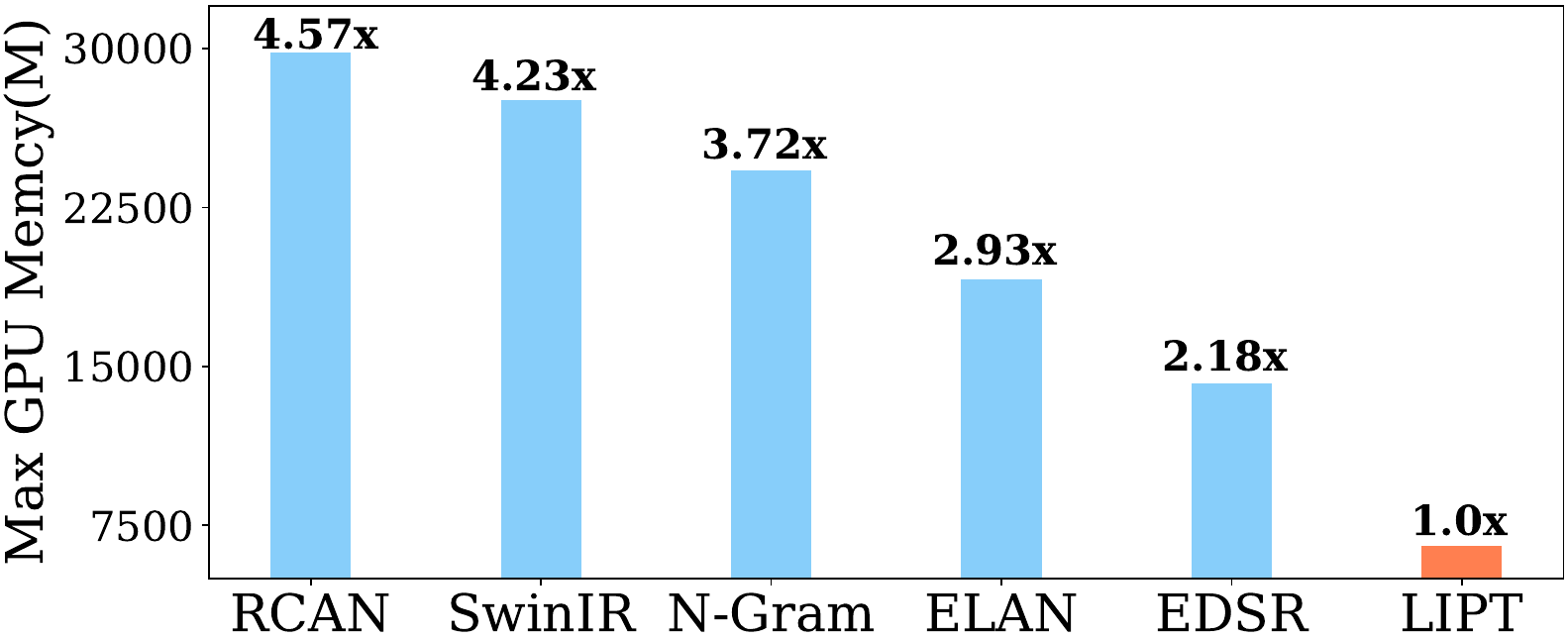}}
	\end{minipage}
        \begin{minipage}{0.48\linewidth}
		\centerline{\includegraphics[width=\textwidth]{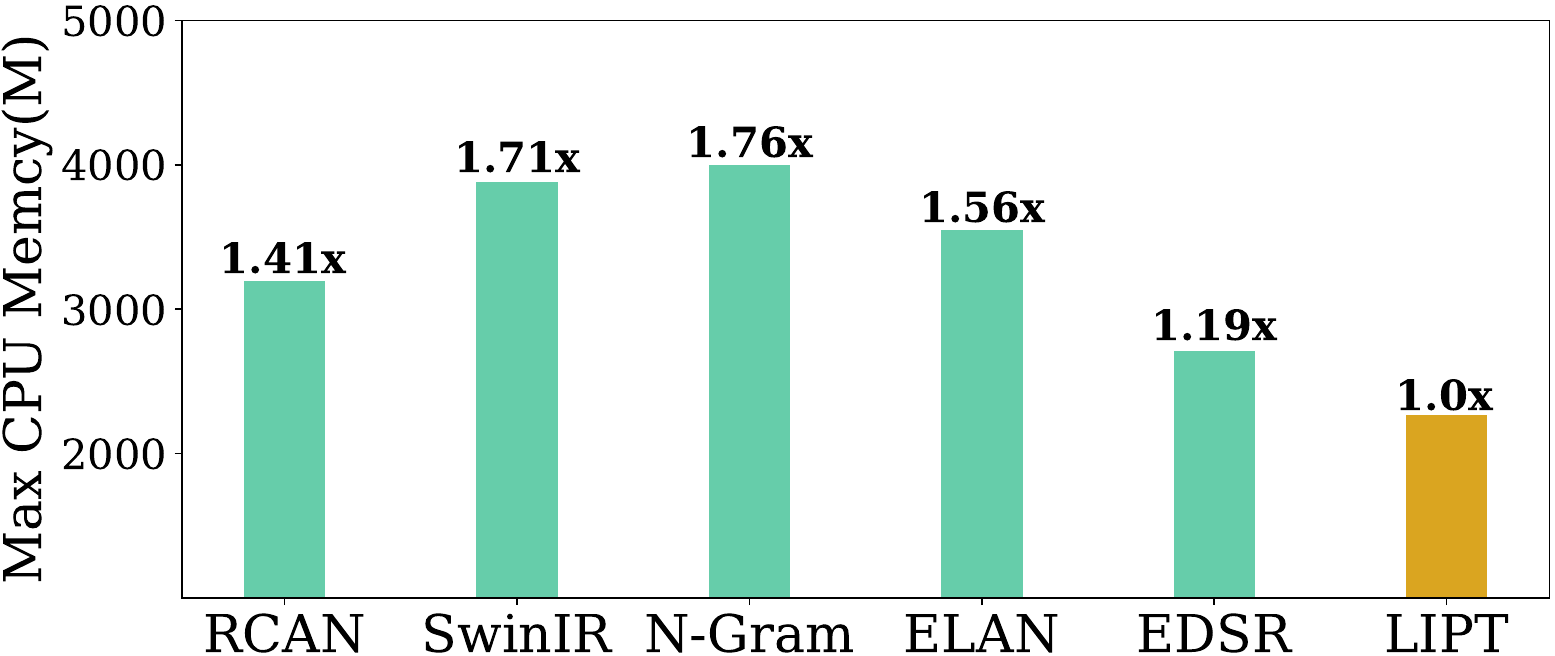}}
	\end{minipage}
	\caption{Maximum memory allocation during inference on DIV2K validation set. Statistics are collected following the implementation of \cite{zhang2019aim}.}
	\label{tab:memory}
  \vspace{-2em}
\end{figure}

\subsection{Image Super-Resolution}


\textbf{Comparison with lightweight SR models.}
We first compare our LIPT-Tiny and LIPT-Base with SOTA lightweight SR models. 
The quantitative results for different methods are reported in Tab.~\ref{tab:light}. By leveraging the long-range dependency modeling capability of self-attention, Transformer-based methods (such as SPIN~\cite{zhang2023lightweight} and DLGSANet-Tiny~\cite{li2023dlgsanet}) demonstrate superior performance over CNN-based methods in terms of PSNR/SSIM. 
However, even with smaller FLOPs using Transformer-based methods, their inference time is significantly increased, compared to CNN models. For example, SPIN only requires smaller 116 GFLOPs (\emph{vs.} 223 GFLOPs in CARN~\cite{ahn2018fast}), while its running GPU latency attains 3,454ms for $\times2$ SR, which is about 93$\times$ larger than CARN.
This significant difference in inference speed can be attributed to the memory-intensive operations and inefficient self-attention computations inherent in Transformer-based models. 
Benefiting from our effective LIPT block design, PTV3-Small not only runs at the lowest latency but achieves the best or second-best PSNR/SSIM at all SR scales compared to Transformer models. For example, LIPT-Small is 1.8$\times$ faster than ELAN-Light on the GPU platform (\emph{i.e.}, 99ms \emph{vs.} 177ms), while outperforming ELAN-Light by 0.11db PSNR on Urban100 for $\times$2 SR.
%
%
Moreover, LIPT-Tiny is the first time for Transformer architecture to achieve real-time inference on GPU at three SR scales while delivering comparable or even superior performance compared to CNN models.

\begin{figure*}[h]
	\vspace{-2em}
	\centering
	\begin{minipage}{0.28\linewidth}
		\vspace{4pt}
		\centerline{\includegraphics[width=\textwidth]{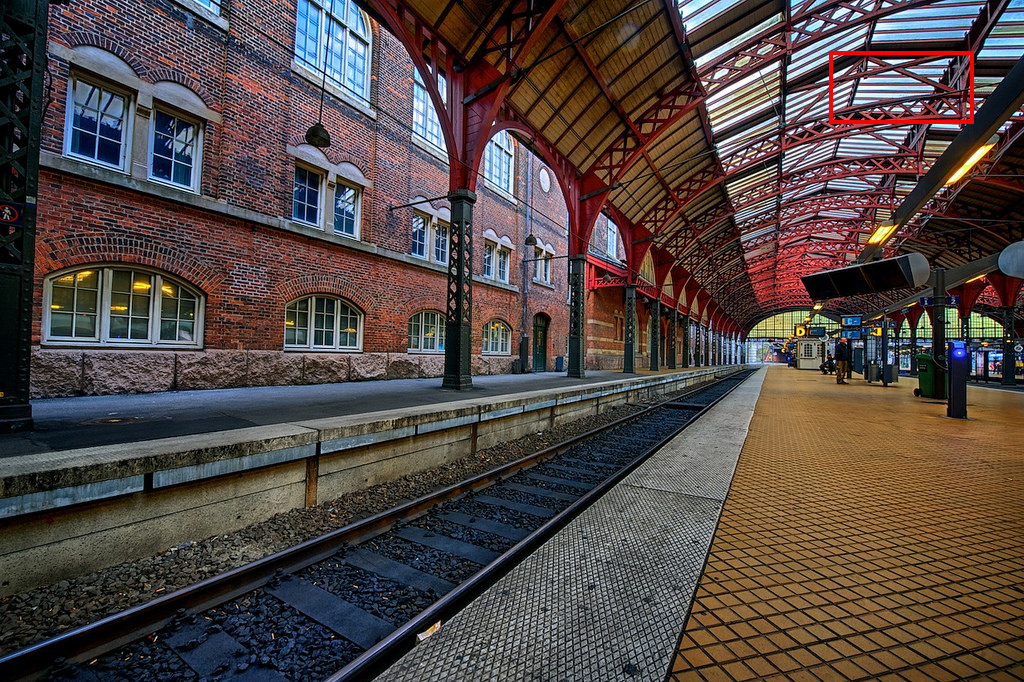}}
		\centerline{\scriptsize img098 from Urban100}
		\vspace{3pt}

	\end{minipage}
	\begin{minipage}{0.13\linewidth}
		\centerline{\includegraphics[width=\textwidth]{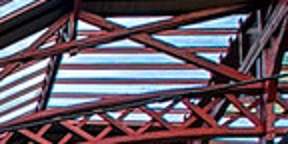}}
        \vspace{-0.12cm}
		\centerline{\tiny (a)HR}
            \vspace{-0.5em}
            \centerline{\tiny PSNR/SSIM}
		\vspace{0.02cm}
		\centerline{\includegraphics[width=\textwidth]{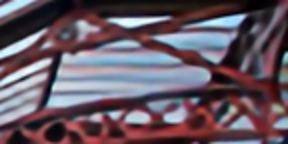}}
        \vspace{-0.12cm}
		\centerline{\tiny (f)SAFMN~\cite{sun2023safmn} }
            \vspace{-0.5em}
            \centerline{\tiny 17.97/0.4625}

	\end{minipage}
	\begin{minipage}{0.13\linewidth}
		\centerline{\includegraphics[width=\textwidth]{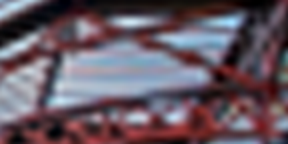}}
        \vspace{-0.12cm}
		\centerline{\tiny (b)Bicubic}
            \vspace{-0.5em}
            \centerline{\tiny 16.22/0.2792}
		\vspace{0.02cm}
		\centerline{\includegraphics[width=\textwidth]{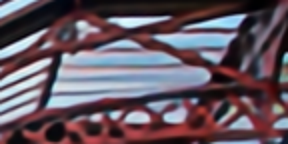}}
        \vspace{-0.12cm}
		\centerline{\tiny (g)ESRT~\cite{zhisheng2021efficient}}
            \vspace{-0.5em}
            \centerline{\tiny 18.22/0.897}

	\end{minipage}
	\begin{minipage}{0.13\linewidth}
		\centerline{\includegraphics[width=\textwidth]{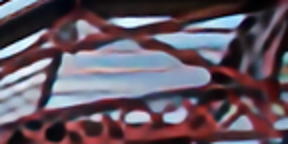}}
        \vspace{-0.12cm}
		\centerline{\tiny (c)CARN~\cite{ahn2018fast}}
            \vspace{-0.5em}
            \centerline{\tiny 17.03/0.3944}

		\vspace{0.02cm}
		\centerline{\includegraphics[width=\textwidth]{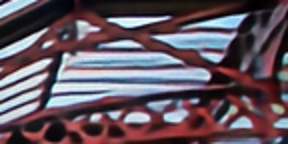}}
        \vspace{-0.12cm}
		\centerline{\tiny (h)SwinIR-L~\cite{liang2021swinir}}
            \vspace{-0.5em}
            \centerline{\tiny 18.88/0.5457}

	\end{minipage}
	\begin{minipage}{0.13\linewidth}
		\centerline{\includegraphics[width=\textwidth]{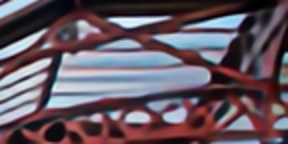}}
        \vspace{-0.12cm}
		\centerline{\tiny (d)EDSR~\cite{lim2017enhanced} }
            \vspace{-0.5em}
            \centerline{\tiny 18.26/0.4828}
		\vspace{0.02cm}
		\centerline{\includegraphics[width=\textwidth]{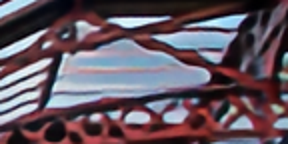}}
        \vspace{-0.12cm}
		\centerline{\tiny (i)N-Gram~\cite{choi2023n}}
            \vspace{-0.5em}
            \centerline{\tiny 17.30/0.4339}

	\end{minipage}
	\begin{minipage}{0.13\linewidth}
		\centerline{\includegraphics[width=\textwidth]{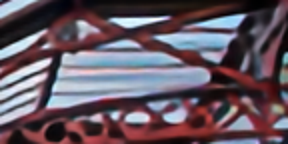}}
        \vspace{-0.12cm}
		\centerline{\tiny (e)LAPAR-A~\cite{li2020lapar}}
            \vspace{-0.5em}
            \centerline{\tiny 17.99/0.4717}
		\vspace{0.02cm}
		\centerline{\includegraphics[width=\textwidth]{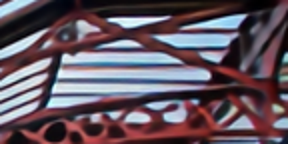}}
        \vspace{-0.12cm}
		\centerline{\tiny (j)LIPT-Small}
            \vspace{-0.5em}
            \centerline{\textbf{\tiny 19.64/0.5754}}

	\end{minipage}
	\vspace{-1em}
	\caption{Qualitative Comparison on the ``img098'' image of Urban100 for $\times$4 SR. }
	\vspace{-1em}
	\label{fig:light}
\end{figure*}

In real-world scenarios, GPU and CPU computing resources are frequently fully utilized. Thus, we also provide the maximum GPU and CPU memory allocation of different models. Despite LIPT requiring more parameters, it achieves the lowest maximum GPU and CPU memory consumption in Fig.~\ref{tab:memory}.

For qualitative comparison, we present our results for $\times4$ SR in Fig.~\ref{fig:light}. We observe various methods yield highly blurred and distorted edges, while LIPT can effectively reconstruct details and alleviate the blurring artifacts on the building structure. 
More visual comparisons for various examples and models are presented in the Appendix. 

\begin{table*}[t]
	\centering
	\scriptsize
	\caption{Comparison of different PSNR-oriented SR models on five benchmarks.}
        \vspace{-1em}
    \resizebox{.99\columnwidth}{!}{
	\label{tab:base}
	\begin{tabular}{|c|c|c|c|c|c|c|c|c|c|c|}
		\hline
		\multirow{2}*{\textbf{Scale}}        & \multirow{2}*{\textbf{Model}}       & \multicolumn{1}{c|}{\textbf{Latency}}    &\multicolumn{1}{c|}{\textbf{Latency}}    & \multicolumn{1}{c|}{\textbf{Params}} & \multicolumn{1}{c|}{\textbf{FLOPs}} & \multirow{2}*{\textbf{Set5}}                       & \multirow{2}*{\textbf{Set14}}                      & \multirow{2}*{\textbf{BSD100}}                       & \multirow{2}*{\textbf{Urban100}}                       & \multirow{2}*{\textbf{Manga109}} \\
  & & GPU (ms) & CPU (s) & (M)& (G) & & & & & \\ \hline
		& RNAN~\cite{zhang2019residual}            & 13,332           &      125               & 9.11                & -                  & 37.17/0.9611                                 & 33.87/0.9207                                 & 32.32/0.9014                                 & 32.73/0.9340                                 & 39.23/0.9785                                 \\
		& OISR~\cite{he2019ode}            & 912                 &    29.28            & 41.91               & 9,657               & 38.21/0.9612                                 & 33.94/0.9206                                 & 32.36/0.9019                                 & 33.03/0.9365                                 & - / -                                        \\
    & RCAN~\cite{zhang2018image}            & 469                    & 13.79            & 15.45                & 3,530                  & 38.27/0.9614                                 & 34.12/0.9216                                 & 32.41/0.9027                                 & 33.34/0.9384                                 & 39.44/0.9786                                 \\
		& SAN~\cite{dai2019second}             & -               &  -                & 15.87               & 3,050               & 38.31/0.9620                                 & 34.07/0.9213                                 & 32.42/0.9028                                 & 33.10/0.9370                                 & 39.32/0.9792                                 \\
		& HAN~\cite{niu2020single}             & 1,420              &  38.58                & 63.61               & 14,551              & 38.27/0.9614                                 & 34.16/0.9217                                 & 32.41/0.9027                                 & 33.35/0.9385                                 & 39.46/0.9785                                 \\
		& NLSN~\cite{mei2021image}            & 1,229               &   75.76              & 41.80                & 9,632               & 38.34/0.9618                                 & 34.08/0.9231                                 & 32.43/0.9027                                 & 33.42/0.9394                                 & 39.59/0.9789                                 \\ \cdashline{2-10}[1pt/1pt] 
		& SwinIR~\cite{liang2021swinir}          & 1,943            &   26.42                 & 11.75               & 3,213               & 38.35/{\color[HTML]{00B0F0}0.9620}                                 & 34.14/0.9227                                 & 32.44/{\color[HTML]{00B0F0}0.9030}                                 & 33.40/{\color[HTML]{00B0F0}0.9393}                                 & 39.60/0.9792                                 \\
		& ELAN~\cite{zhang2022efficient}            & {\color[HTML]{00B0F0}924}  &  91.01                              & 8.25                & 1,965               & {\color[HTML]{00B0F0}38.36/0.9620}                                 & 33.20/0.9228                                  & {\color[HTML]{00B0F0}32.45/0.9030}                                 & {\color[HTML]{00B0F0}33.44}/0.9391                                 & {\color[HTML]{00B0F0}39.62/0.9793}                                 \\
		& SRFormer~\cite{zhou2023srformer}            & 1,188          &    {\color[HTML]{00B0F0}23.51}                   & 0.85               & 236                & 38.23/0.9613                                  & 33.94/0.9209                                 & 32.36/0.9019                                 & 32.91/0.9353                                  & 39.28/0.9785                                 \\
        & DLGSANet~\cite{li2023dlgsanet}        & 1,277                 &    -           & 4.73                & 1,097               & 38.34/0.9617                                 & {\color[HTML]{00B0F0}34.25/0.9231}                                 & 32.38/0.9025                                 & 33.41/{\color[HTML]{00B0F0}0.9393}                                 & 39.57/0.9789                                 \\
		\multirow{-9}{*}{×2} & \textbf{LIPT-Base}      & {\color[HTML]{FF0000} 892}  & {\color[HTML]{FF0000}19.48} & 26.05               & 5,211               & {\color[HTML]{FF0000} 38.38/0.9621} & {\color[HTML]{FF0000} 34.33/0.9239} & {\color[HTML]{FF0000} 32.47/0.9035} & {\color[HTML]{FF0000} 33.55/0.9411} & {\color[HTML]{FF0000} 39.66/0.9795} \\ \hline
		& RNAN~\cite{zhang2019residual}            & 393                 &   53.54             & 9.26                & 480                & 32.49/08982                                  & 28.83/0.7878                                 & 27.72/0.7421                                 & 26.61/0.8023                                 & 31.09/0.9151                                 \\
		& OISR~\cite{he2019ode}            & 241                      &  8.01          & 44.27               & 2,963               & 32.53/0.8992                                 & 28.86/0.7878                                 & 27.75/0.7428                                 & 26.79/0.8068                                 & - / -                                        \\
  & RCAN~\cite{zhang2018image}            &  176                & 3.48               & 15.59                & 918                  & 32.63/0.9002                                 & 28.87/0.7889                                 & 27.77/0.7436                                 & 26.82/0.8087                                 & 31.22/0.9173                                 \\
		& SAN~\cite{dai2019second}             & 3,448               &     83.48            & 15.86               & 937                & 32.64/0.9003                                 & 28.92/0.7888                                 & 27.78/0.7436                                 & 26.79/0.8068                                 & 31.18/0.9169                                 \\
		& HAN~\cite{niu2020single}             & 379                  &  9.64             & 64.20                & 3,776               & 32.64/0.9002                                 & 28.90/0.7890                                  & 27.80/0.7442                                 & 26.85/0.8094                                 & 31.42/0.9177                                 \\
		& NLSN~\cite{mei2021image}            & 309                 &    19.32            & 44.16               & 2,956               & 32.59/0.9000                                 & 28.87/0.7891                                 & 27.78/0.7444                                 & 26.96/0.8109                                 & 31.27/0.9184                                 \\ \cdashline{2-10}[1pt/1pt] 
		& SwinIR~\cite{liang2021swinir}          & 375               &   {\color[HTML]{00B0F0}7.25}               & 11.90                & 584                & 32.72/0.9021                                 & 28.94/{\color[HTML]{00B0F0}0.7914}                                 & 27.83/0.7459                                 & 27.07/0.8164                                 & 31.67/{\color[HTML]{00B0F0}0.9226}                                 \\
		& ELAN~\cite{zhang2022efficient}            & {\color[HTML]{00B0F0}237}    &  20.33                            & 8.31                & 494                & 32.75/{\color[HTML]{00B0F0}0.9022}                                 & {\color[HTML]{00B0F0}28.96/0.7914}                                 & 27.83/0.7459                                 & 27.13/0.8167                                 & 31.68/{\color[HTML]{00B0F0}0.9226}                                 \\
		& DLGSANet~\cite{li2023dlgsanet}        & 293              &      -             & 4.76                & 274                & {\color[HTML]{00B0F0}32.80}/0.9021                                  & 28.95/0.7907                                 & {\color[HTML]{00B0F0}27.85/0.7464}                                 & {\color[HTML]{00B0F0}27.17/0.8175}                                 & {\color[HTML]{00B0F0}31.68}/0.9219                                 \\
		\multirow{-9}{*}{×4} & \textbf{LIPT-Base} & {\color[HTML]{FF0000} 223}       & {\color[HTML]{FF0000}6.47}   & 26.09               & 1,361               & {\color[HTML]{FF0000} 32.82/ 0.9025} & {\color[HTML]{FF0000} 28.99/.07921} & {\color[HTML]{FF0000} 27.86/0.7486} & {\color[HTML]{FF0000} 27.19/ 0.8181} & {\color[HTML]{FF0000} 31.70/0.9230}                                     \\ \hline
	\end{tabular}}
 \vspace{-2em}
\end{table*}

\textbf{Comparison with PSNR-oriented SR models.}
To validate the scalability of LIPT, we further compare our LIPT-Base with SOTA PSNR-oriented SR models, as shown in Tab.~\ref{tab:base}. LIPT-Base also surpasses all other models in PSNR/SSIM across all benchmarks and SR factors. For example, LIPT-Base outperforms the best CNN-based NLSN~\cite{mei2021image} and Transformer-based ELAN~\cite{zhang2022efficient} by 0.13db and 0.11db PSNR on Urban100 for $\times2$ SR, while maintaining the fastest inference speed. In addition, LIPT-Base also achieves the best visualization results compared to other methods, which are presented in the Appendix.
%

\begin{table*}[h]
	\centering
	\footnotesize
        \vspace{-2em}
	\caption{Comparison with SOTA methods for JPEG compression artifact reduction. C: $x$ \& G: $y$ means the method runs $x$ and $y$ latency on CPU and GPU, respectively.}
        \vspace{-.5em}
    \resizebox{.99\columnwidth}{!}{
	\label{tab:jpeg}
	\begin{tabular}{|c|c|cc|cc|cc|cc|cc|cc|}
		\hline
                          &                     & \multicolumn{2}{c|}{\textbf{RNAN~\cite{zhang2019residual}} with}      & \multicolumn{2}{c|}{\textbf{SwinIR~\cite{liang2021swinir}} with}                           & \multicolumn{2}{c|}{\textbf{GRL-S}~\cite{li2023efficient} with}                    & \multicolumn{2}{c|}{\textbf{CAT}~\cite{chen2022cross} with}                                                     & \multicolumn{2}{c|}{\textbf{ART}~\cite{zhang2022accurate} with}                                                     & \multicolumn{2}{c|}{\textbf{LIPT} with}                                                  \\
                          &                     & \multicolumn{2}{c|}{C: 708s \& G: 7.8s}    & \multicolumn{2}{c|}{C:{\color[HTML]{00B0F0}83.2s} \& G: 7.4s}                           & \multicolumn{2}{c|}{C:87.5s \& G:{\color[HTML]{00B0F0} 6.9s}}                   & \multicolumn{2}{c|}{C:331.3 \& G: 34.7s}                                                  & \multicolumn{2}{c|}{C:177.3s \& G: 10.1s}                                                  & \multicolumn{2}{c|}{C:{\color[HTML]{FF0000}{\textbf{71.4s}}} \& G:{\color[HTML]{FF0000}{\textbf{3.6s}}} }                                                 \\ \cline{3-14} 
		\multirow{-3}{*}{\textbf{Dataset}}  & \multirow{-3}{*}{\textbf{Q}} & \multicolumn{1}{c}{PSNR} & SSIM   & \multicolumn{1}{c}{PSNR} & SSIM  & \multicolumn{1}{c}{PSNR}    & SSIM   & \multicolumn{1}{c}{PSNR}    & SSIM                          & \multicolumn{1}{c}{PSNR}    & SSIM                          & \multicolumn{1}{c}{PSNR}         & SSIM          \\ \hline
		& 10                  & 29.63                     & 0.8239 &   29.86                   & 0.8287 &  29.80                       & 0.8279 & {\color[HTML]{FF0000} 29.89} & 0.8295                        & {\color[HTML]{FF0000} 29.89} & {\color[HTML]{FF0000} 0.8300}   & {\color[HTML]{00B0F0}29.88}                             & {\color[HTML]{00B0F0}0.8297}        \\
		& 30                  & 33.45                     & 0.9149 &  33.69                    &  0.9174 &  33.64                       & 0.9169  & {\color[HTML]{FF0000} 33.73} & 0.9177                        & {\color[HTML]{00B0F0}33.71}                        & {\color[HTML]{00B0F0} 0.9178} & {\color[HTML]{00B0F0}33.71 }                            & {\color[HTML]{FF0000}0.9181}        \\
		\multirow{-3}{*}{LIVE1}    & 40                  & 34.47                     & 0.9299 &   34.67                   & 0.9317   &  34.63                       & 0.9314  & {\color[HTML]{FF0000} 34.72} & 0.9320                         & {\color[HTML]{00B0F0}34.70}                         & {\color[HTML]{00B0F0} 0.9322} & {\color[HTML]{FF0000}34.72}                             & {\color[HTML]{FF0000}0.9323}        \\ \hline
		& 10                  & 29.96                     & 0.8178 & {\color[HTML]{00B0F0} 30.27}                      & 0.8249  & 30.19  & {\color[HTML]{FF0000}0.8287}  & 30.26                        & 0.8250                         & {\color[HTML]{00B0F0} 30.27} & {\color[HTML]{FF0000} 0.8258} & {\color[HTML]{FF0000}30.28}                             & 0.8245        \\
		& 30                  & 33.38                     & 0.8924 & 33.73                      & 0.8961  &  33.72                       & {\color[HTML]{FF0000}0.8986}  & {\color[HTML]{00B0F0}33.77}                         &  0.8964 & {\color[HTML]{FF0000} 33.78} & 0.8964 & {\color[HTML]{00B0F0}33.77}                             & {\color[HTML]{00B0F0}0.8967}        \\
		\multirow{-3}{*}{Classic5} & 40                  & 34.27                     & 0.9061 &  34.52                    & 0.9082  &     34.53                    & {\color[HTML]{FF0000}0.9108}  & {\color[HTML]{FF0000} 34.58} & 0.9087 & {\color[HTML]{00B0F0}34.55}                        & 0.9086                        & {\color[HTML]{FF0000}34.58}                             & {\color[HTML]{00B0F0} 0.9090}        \\ \hline
	\end{tabular}}
\vspace{-.5em}
\end{table*}

\begin{figure}[t]
	\centering
	\begin{minipage}{0.128\linewidth}
		\vspace{2pt}
		\centerline{\includegraphics[width=\textwidth]{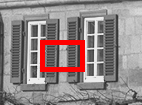}}
		\centerline{\tiny buildings}
		\vspace{3pt}
	\end{minipage}
	\begin{minipage}{0.13\linewidth}
		\centerline{\includegraphics[width=\textwidth]{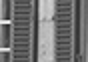}}
            \vspace{-0.5em}
		\centerline{\tiny (a) HR}
            \vspace{-0.5em}
            \centerline{\tiny PSNR/SSIM}
		\vspace{0.01cm}

	\end{minipage}
	\begin{minipage}{0.13\linewidth}
		
		\centerline{\includegraphics[width=\textwidth]{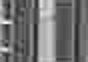}}
            \vspace{-0.5em}
		\centerline{\tiny (b) JPEG}
            \vspace{-0.5em}
            \centerline{\tiny 29.45/0.7617}
		\vspace{0.01cm}

	\end{minipage}
	\begin{minipage}{0.13\linewidth}
		\centerline{\includegraphics[width=\textwidth]{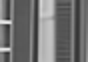}}
            \vspace{-0.5em}
		\centerline{\tiny (c) SwinIR~\cite{liang2021swinir}}
            \vspace{-0.5em}
            \centerline{\tiny 29.60/0.7747}
		\vspace{0.01cm}

	\end{minipage}
 	\begin{minipage}{0.13\linewidth}
		\centerline{\includegraphics[width=\textwidth]{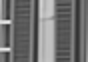}}
            \vspace{-0.5em}
		\centerline{\tiny (d) ART~\cite{zhang2022accurate}}
            \vspace{-0.5em}
        \centerline{\tiny 29.84/0.7825}

	\end{minipage}
  	\begin{minipage}{0.13\linewidth}
		\centerline{\includegraphics[width=\textwidth]{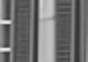}}
            \vspace{-0.5em}
		\centerline{\tiny (e) GRL-S~\cite{li2023efficient}}
            \vspace{-0.5em}
        \centerline{\tiny 29.69/0.7798}
		
	\end{minipage}
        \begin{minipage}{0.13\linewidth}
		\centerline{\includegraphics[width=\textwidth]{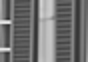}}
            \vspace{-0.5em}
		\centerline{\tiny (f) LIPT}
            \vspace{-0.5em}
            \centerline{\textbf{\tiny 29.98/0.7964}}
		
	\end{minipage}
	\vspace{-0.8em}
	\caption{Visual Comparison on LIVE1``buildings'' for CAR q=10.}
	\label{fig:jpeg}
	\vspace{-1em}
\end{figure}

\subsection{JPEG Compression Artifact Reduction}
As presented in Tab.~\ref{tab:jpeg}, we further compare our LIPT-Base with SOTA JPEG compression artifact removal methods, 
across three JPEG quality settings of 10, 30 and 40. 
Our LIPT achieves the fastest GPU and CPU inference speeds of 3.6s and 71.4s with the highest SSIM for both LIVE1 and Classic5 datasets on the quality of 30 and 40, compared to other SOTA methods. At the quality of 10, our method also achieves comparable results.
In Fig.~\ref{fig:jpeg}, we further provide visual comparisons at the quality of q=10. We observe that LIPT can remove the artifact and restore the sharpest textures.


\subsection{Image Denoising}
Tab.~\ref{tab:denoising} shows quantitative comparisons on CBSD68, Kodak24, McMaster and Urban100 with noise levels of 15, 25 and 50.
our LIPT achieves the fastest CPU and the second-fastest GPU inference speed of 76s and 4,095ms to handle one image with a resolution of 1280$\times$720. In fact, on the Urban100 dataset, we outperform Restormer with only 106ms GPU latency reduction by 0.16db, 0.19dB and 0.14dB with noise levels of 15, 25 and 50, respectively. Meanwhile, LIPT achieves a higher PSNR on CBSD68 and a comparable PSNR with ART on Ubran100 and McMaster, while achieving $3\times$ GPU running faster than ART. It is important to note that our training iteration number is only half of that using ART (\emph{i.e.}, 800K \emph{vs.} 1.6M iterations).

\begin{table}[t]
	\centering 
	\footnotesize
	\caption{Comparison with state-of-the-art methods for image denoising using PSNR. }
	\vspace{-0.8em}
	\label{tab:denoising}
        \resizebox{.99\columnwidth}{!}{
	\begin{tabular}{|c|c|c|ccc|ccc|ccc|ccc|}
		\hline
		\multirow{2}{*}{\textbf{Method}} & \multicolumn{1}{c|}{\textbf{Latency}} & \multicolumn{1}{c|}{\textbf{Latency}} & \multicolumn{3}{c|}{\textbf{CBSD68}}                                              & \multicolumn{3}{c|}{\textbf{Kodak24}}                                             & \multicolumn{3}{c|}{\textbf{McMaster}}                                            & \multicolumn{3}{c|}{\textbf{Urban100}}                                            \\ \cline{4-15} 
		&      GPU (ms)           & CPU (s)                & \multicolumn{1}{c|}{\textbf{15}} & \multicolumn{1}{c|}{\textbf{25}} & \textbf{50} & \multicolumn{1}{c|}{\textbf{15}} & \multicolumn{1}{c|}{\textbf{25}} & \textbf{50} & \multicolumn{1}{c|}{\textbf{15}} & \multicolumn{1}{c|}{\textbf{25}} & \textbf{50} & \multicolumn{1}{c|}{\textbf{15}} & \multicolumn{1}{c|}{\textbf{25}} & \textbf{50} \\ \hline
		RNAN~\cite{zhang2019residual}                             & 9,763    &   739                      & -                                & -                                & 28.27       & -                                & -                                & 29.58       & -                                & -                                & 29.72       & -                                & -                                & 29.08       \\ \cline{1-1}
		IPT~\cite{chen2021pre}                              & 8,414       &   454                   & -                                & -                                & 28.39       & -                                & -                                & 29.64       & -                                & -                                & 29.98       & -                                & -                                & 29.71       \\ \cline{1-1}
		EDT-B~\cite{li2021efficient}                            & 19,698    &    -                  & 34.39                            & 31.76                            & 28.56       & 35.37                            & 32.94                            & 29.87       & 35.61                            & 33.34                            & 30.25       & 35.22                            & 33.07                            & 30.16       \\ \cline{1-1}
		SwinIR~\cite{liang2021swinir}                           & 9,191        &    94               & {\color[HTML]{00B0F0}34.42}                            & 31.78                            & 28.56       & 35.34                            & 32.89                            & 29.79       & 35.61                            & 33.2                             & 30.22       & 35.13                            & 32.9                             & 29.82       \\ \cline{1-1}
		Restormer~\cite{zamir2022restormer}                        & {\color[HTML]{FF0000}3,989}       &     205               & 34.4                             & 31.79                            & 28.60        & {\color[HTML]{FF0000}35.47}                            & {\color[HTML]{FF0000}33.04}                            & {\color[HTML]{FF0000}30.01}       & 35.61                            & 33.34                            & {\color[HTML]{00B0F0}30.30}        & 35.13                            & 32.96                            & 30.02       \\ \cline{1-1}
		GRL-S~\cite{li2023efficient}                            & 8,612       &   {\color[HTML]{00B0F0}92}                 & 34.36                            & 31.72                            & 28.51       & 35.32                            & 32.88                            & 29.77       & 35.59                            & 33.29                            & 30.18       & {\color[HTML]{00B0F0}35.24}                            & 33.07                            & 30.09       \\ \cline{1-1}
  		ELAN~\cite{zhang2022efficient}                            &  4,325        &  352                  & -                            & {\color[HTML]{00B0F0}31.82}                            & -       & -                           & 32.89                            & -       & -                            & 33.21                            & -       & -                            & 32.94                            & -       \\ \cline{1-1}
		ART~\cite{zhang2022accurate}                              & 12,024       &  170                 & {\color[HTML]{FF0000}34.46}                            & {\color[HTML]{00B0F0}31.84}                            & {\color[HTML]{00B0F0}28.63}       & {\color[HTML]{00B0F0}35.39}                            & {\color[HTML]{00B0F0}32.95}                            & {\color[HTML]{00B0F0}29.87}       & {\color[HTML]{00B0F0}35.68}                            & {\color[HTML]{FF0000}33.41}                            & {\color[HTML]{FF0000}30.31}       & {\color[HTML]{FF0000}35.29}                            & {\color[HTML]{00B0F0}33.14}                            & {\color[HTML]{FF0000}30.19}       \\ \cline{1-1}
		LIPT                             & {\color[HTML]{00B0F0}4,095}                 &  {\color[HTML]{FF0000}76 }        & {\color[HTML]{FF0000}34.46}                          & {\color[HTML]{FF0000}31.85}                         & {\color[HTML]{FF0000}28.64}     & 35.36                          & {\color[HTML]{00B0F0}32.95}                          & 29.85     & {\color[HTML]{FF0000}35.69}                           & {\color[HTML]{00B0F0}33.39}                          & {\color[HTML]{00B0F0}30.30}     & {\color[HTML]{FF0000}35.29}                          & {\color[HTML]{FF0000}33.15}                          & {\color[HTML]{00B0F0}30.16}     \\ \hline
	\end{tabular}}
 \vspace{-.8em}
\end{table}

\subsection{Ablation Study}
\label{sec:aba}

\begin{table}[t]
\centering 
\footnotesize
\caption{Effect of MSA-Conv (Eq.~\ref{stru}), NVSM-SA (Eq.~\ref{eq_nvsm}) and HRM (Eq.~\ref{eq_hrm_o}) in LIPT. Baseline comprises standard MSA and MLP stacks with a depth equal to LIPT-Small.} 
\vspace{-.8em}
\label{tab:components}
\begin{tabular}{l|c|ccccc}
\hline
\multicolumn{1}{c|}{Architecture}       & GPU (ms) & \multicolumn{1}{c|}{Set5} & \multicolumn{1}{c|}{Set14} & \multicolumn{1}{c|}{BSD100} & \multicolumn{1}{c|}{Urban100} & Manga109  \\ \hline
Baseline                   & 179      & 38.17                     & 33.94                      & 32.33                     & 32.74                     & 39.17 \\ \hline
+ MSA-Conv (Eq. \ref{stru})                       & 99       & 38.19                     & 33.95                      & 32.32                     & 32.72                     & 39.09 \\ \hline
+ (Eqs. \ref{stru} \& \ref{eq_nvsm})                                  & 99       & 38.19                     & 33.96                      & 32.34                     & 32.82                     & 39.15 \\ \hline
+ (Eqs. \ref{stru} \& \ref{eq_nvsm} \& \ref{eq_hrm_o})                                  & 99       & 38.20                      & 33.96                      & 32.36                     & 32.87                     & 39.21 \\ \hline
\end{tabular}
\vspace{-1em}
\end{table}

\textbf{Effect of MSA-Conv, NVSM-SA and HRM.} To evaluate the effect of various components within LIPT, we construct a simple baseline with a stack of MSA-MLP models and the same depth as LIPT-Small.  As presented in Tab.~\ref{tab:components}, the baseline achieves 179ms GPU latency.
By utilizing our low computation-density MSA-Conv, the GPU inference time decreases from 179ms to 99ms, while PSNR decreases slightly on BSD100 and Urban100. We also explore the ratio of MSA and convolution block in Eq.~\ref{stru} to obtain the best combination, which is presented in the Appendix.
Replacing MSA with the proposed NVSM-SA (see + (Eqs. \ref{stru} \& \ref{eq_nvsm}) \emph{vs.} + MSA-Conv (Eqs. \ref{stru})) significantly improves PSNR of 0.1dB on Urban100. 
Finally, by adding the proposed HRM, we further improve the performance of 0.05db PSNR on Urban100 without incurring any additional inference cost (\emph{i.e.}, the last row \emph{vs.} penultimate row). The effectiveness of isotropic Sobel in HRM is evaluated in the Appendix.

\begin{table}[t]
\centering 
\footnotesize
\caption{Ablation study on the effect of SLWA and DLWA in Eq.~\ref{eq_mask}.}
\vspace{-1em}
\label{tab:SLWA}
\begin{tabular}{cc|ccccc}
\hline
\multicolumn{1}{c|}{DLWA} & \multicolumn{1}{|c|}{SLWA} &\multicolumn{1}{c|}{Set5} & \multicolumn{1}{c|}{Set14} & \multicolumn{1}{c|}{BSD100} & \multicolumn{1}{c|}{Urban100} & Manga109                  \\ \hline
\Checkmark           &  \XSolidBrush                 & 38.19                     & 33.95                      & 32.32                     & 32.72                     & 39.09                 \\ \hline  \XSolidBrush
&  \Checkmark       & 38.18     & 33.93       & 32.31     & 32.80      &39.15 \\ \hline
\Checkmark            & \Checkmark            & 38.20                     & 33.96                      & 32.36                     & 32.87                     & 39.21                 \\ \hline
\end{tabular}
\vspace{-2em}
\end{table}

\begin{figure}[h]
	\centering
 \vspace{-2em}
	\includegraphics[width=0.6\linewidth]{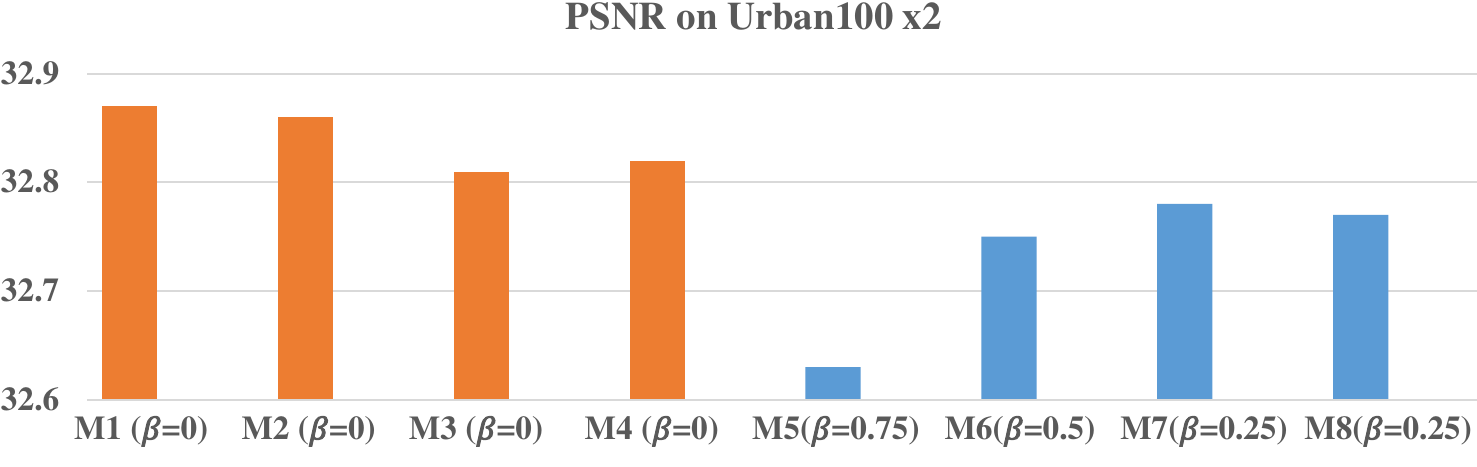}
 \vspace{-.5em}
	\caption{Effect of non-volatile (M1-M4) and volatile (M5-M8) sampling masks.}
	\label{fig:non-volatile_result}
 \vspace{-2em}
\end{figure}

\textbf{Effect of DLWA and SLWA in NVSM-SA.} 
To further evaluate the importance of SLWA and DLWA in NVSM-SA (\emph{i.e.}, Eq.~\ref{eq_mask}), we conducted ablation by using only DLWA, only SLWA, and both DLWA and SLWA in Tab.~\ref{tab:SLWA}. 
The results 
demonstrate the effectiveness of both DLWA and SLWA, as both local dense information and global attention with a larger receptive field are crucial to enhancing the strong representation ability of Transformer-based models.


\textbf{Effect of Non-volatile and volatile sampling masks.} To further explore the relationship between non-volatile sparse masks and the sampling drop rate $\beta$, we searched four non-volatile masks and four volatile masks individually (in Fig.~\ref{fig:non-volatile}). 
In Fig.~\ref{fig:non-volatile_result}, The left non-volatile mask (M1-M4) demonstrates significantly superior performance compared to the right volatile mask (M5-M8). Moreover, the larger $\beta$ of the volatile mask achieves a relatively lower PSNR.



\section{Conclusion}
In this paper, we propose a latency-aware image-processing transformer (LIPT), whose block is adopted the low-proportion MSA-Conv design for substantial acceleration. We further propose a novel NVSM-SA to capture the contextual information by utilizing a pre-computing sparse mask acting on a larger window attention with no extra computation overhead. Benefiting from the revival of convolutional structures, we propose an HRM to enhance the model's detail reconstruction capability. Extensive experiments on various image processing tasks demonstrate that our LIPT outperforms previous SOTA Transformers.

%
%
\bibliographystyle{splncs04}
\bibliography{main}

\begin{thebibliography}{10}
\providecommand{\url}[1]{\texttt{#1}}
\providecommand{\urlprefix}{URL }
\providecommand{\doi}[1]{https://doi.org/#1}

\bibitem{ahn2018fast}
Ahn, N., Kang, B., Sohn, K.A.: Fast, accurate, and lightweight super-resolution with cascading residual network. In: ECCV. pp. 252--268 (2018)

\bibitem{arbelaez2010contour}
Arbelaez, P., Maire, M., Fowlkes, C., Malik, J.: Contour detection and hierarchical image segmentation. IEEE TPAMI  \textbf{33}(5),  898--916 (2010)

\bibitem{arora2018optimization}
Arora, S., Cohen, N., Hazan, E.: On the optimization of deep networks: Implicit acceleration by overparameterization. In: ICML. pp. 244--253 (2018)

\bibitem{bevilacqua2012low}
Bevilacqua, M., Roumy, A., Guillemot, C., Alberi-Morel, M.L.: Low-complexity single-image super-resolution based on nonnegative neighbor embedding. In: British Machine Vision Conference, {BMVC} 2012, Surrey, UK, September 3-7, 2012. pp. 1--10. {BMVA} Press (2012)

\bibitem{carion2020end}
Carion, N., Massa, F., Synnaeve, G., Usunier, N., Kirillov, A., Zagoruyko, S.: End-to-end object detection with transformers. In: ECCV. pp. 213--229 (2020)

\bibitem{charbonnier1994two}
Charbonnier, P., Blanc-Feraud, L., Aubert, G., Barlaud, M.: Two deterministic half-quadratic regularization algorithms for computed imaging. In: ICIP. vol.~2, pp. 168--172 (1994)

\bibitem{chen2021pre}
Chen, H., Wang, Y., Guo, T., Xu, C., Deng, Y., Liu, Z., Ma, S., Xu, C., Xu, C., Gao, W.: Pre-trained image processing transformer. In: CVPR. pp. 12299--12310 (2021)

\bibitem{chen2023recursive}
Chen, Z., Zhang, Y., Gu, J., Kong, L., Yang, X.: Recursive generalization transformer for image super-resolution. In: ICLR (2024)

\bibitem{chen2023dual}
Chen, Z., Zhang, Y., Gu, J., Kong, L., Yang, X., Yu, F.: Dual aggregation transformer for image super-resolution. In: ICCV. pp. 12312--12321 (2023)

\bibitem{chen2022cross}
Chen, Z., Zhang, Y., Gu, J., Kong, L., Yuan, X., et~al.: Cross aggregation transformer for image restoration. NeurIPS  \textbf{35},  25478--25490 (2022)

\bibitem{chen2024hierarchical}
Chen, Z., Zhang, Y., Liu, D., Gu, J., Kong, L., Yuan, X., et~al.: Hierarchical integration diffusion model for realistic image deblurring. In: NeurIPS (2023)

\bibitem{choi2023n}
Choi, H., Lee, J., Yang, J.: N-gram in swin transformers for efficient lightweight image super-resolution. In: CVPR. pp. 2071--2081 (2023)

\bibitem{dai2019second}
Dai, T., Cai, J., Zhang, Y., Xia, S.T., Zhang, L.: Second-order attention network for single image super-resolution. In: CVPR (2019)

\bibitem{ding2019acnet}
Ding, X., Guo, Y., Ding, G., Han, J.: Acnet: Strengthening the kernel skeletons for powerful cnn via asymmetric convolution blocks. In: ICCV. pp. 1911--1920 (2019)

\bibitem{ding2021diverse}
Ding, X., Zhang, X., Han, J., Ding, G.: Diverse branch block: Building a convolution as an inception-like unit. In: CVPR. pp. 10886--10895 (2021)

\bibitem{ding2021repvgg}
Ding, X., Zhang, X., Ma, N., Han, J., Ding, G., Sun, J.: Repvgg: Making vgg-style convnets great again. In: CVPR (2021)

\bibitem{dong2015compression}
Dong, C., Deng, Y., Loy, C.C., Tang, X.: Compression artifacts reduction by a deep convolutional network. In: ICCV. pp. 576--584 (2015)

\bibitem{dong2014learning}
Dong, C., Loy, C.C., He, K., Tang, X.: Learning a deep convolutional network for image super-resolution. In: ECCV. pp. 184--199 (2014)

\bibitem{dosovitskiy2020image}
Dosovitskiy, A., Beyer, L., Kolesnikov, A., Weissenborn, D., Zhai, X., Unterthiner, T., Dehghani, M., Minderer, M., Heigold, G., Gelly, S., et~al.: An image is worth 16x16 words: Transformers for image recognition at scale. ICLR  (2021)

\bibitem{foi2007pointwise}
Foi, A., Katkovnik, V., Egiazarian, K.: Pointwise shape-adaptive dct for high-quality denoising and deblocking of grayscale and color images. IEEE TIP  \textbf{16}(5),  1395--1411 (2007)

\bibitem{franzen1999kodak}
Franzen, R.: Kodak lossless true color image suite. source: http://r0k. us/graphics/kodak  \textbf{4}(2), ~9 (1999)

\bibitem{he2019ode}
He, X., Mo, Z., Wang, P., Liu, Y., Yang, M., Cheng, J.: Ode-inspired network design for single image super-resolution. In: CVPR (2019)

\bibitem{huang2015single}
Huang, J.B., Singh, A., Ahuja, N.: Single image super-resolution from transformed self-exemplars. In: CVPR. pp. 5197--5206 (2015)

\bibitem{Hui-IMDN-2019}
Hui, Z., Gao, X., Yang, Y., Wang, X.: Lightweight image super-resolution with information multi-distillation network. In: Proceedings of the 27th ACM International Conference on Multimedia (ACM MM). pp. 2024--2032 (2019)

\bibitem{kim2016accurate}
Kim, J., Lee, J.K., Lee, K.M.: Accurate image super-resolution using very deep convolutional networks. In: CVPR. pp. 1646--1654 (2016)

\bibitem{kingma2014adam}
Kingma, D.P., Ba, J.: Adam: A method for stochastic optimization. arXiv preprint arXiv:1412.6980  (2014)

\bibitem{li2023feature}
Li, A., Zhang, L., Liu, Y., Zhu, C.: Feature modulation transformer: Cross-refinement of global representation via high-frequency prior for image super-resolution. In: CVPR. pp. 12514--12524 (2023)

\bibitem{li2021efficient}
Li, W., Lu, X., Qian, S., Lu, J., Zhang, X., Jia, J.: On efficient transformer-based image pre-training for low-level vision. arXiv preprint arXiv:2112.10175  (2021)

\bibitem{li2020lapar}
Li, W., Zhou, K., Qi, L., Jiang, N., Lu, J., Jia, J.: Lapar: Linearly-assembled pixel-adaptive regression network for single image super-resolution and beyond. Advances in Neural Information Processing Systems  \textbf{33},  20343--20355 (2020)

\bibitem{li2023dlgsanet}
Li, X., Pan, J., Tang, J., Dong, J.: Dlgsanet: Lightweight dynamic local and global self-attention networks for image super-resolution. In: ICCV (2023)

\bibitem{li2023efficient}
Li, Y., Fan, Y., Xiang, X., Demandolx, D., Ranjan, R., Timofte, R., Van~Gool, L.: Efficient and explicit modelling of image hierarchies for image restoration. In: CVPR. pp. 18278--18289 (2023)

\bibitem{li2022blueprint}
Li, Z., Liu, Y., Chen, X., Cai, H., Gu, J., Qiao, Y., Dong, C.: Blueprint separable residual network for efficient image super-resolution. In: CVPR. pp. 833--843 (2022)

\bibitem{liang2021swinir}
Liang, J., Cao, J., Sun, G., Zhang, K., Van~Gool, L., Timofte, R.: Swinir: Image restoration using swin transformer. In: CVPR. pp. 1833--1844 (2021)

\bibitem{lim2017enhanced}
Lim, B., Son, S., Kim, H., Nah, S., Mu~Lee, K.: Enhanced deep residual networks for single image super-resolution. In: CVPRW. pp. 136--144 (2017)

\bibitem{lin2017espace}
Lin, S., Ji, R., Chen, C., Huang, F.: Espace: Accelerating convolutional neural networks via eliminating spatial and channel redundancy. In: AAAI (2017)

\bibitem{liu2020residual}
Liu, J., Tang, J., Wu, G.: Residual feature distillation network for lightweight image super-resolution. In: ECCVW. pp. 41--55 (2020)

\bibitem{liu2021swin}
Liu, Z., Lin, Y., Cao, Y., Hu, H., Wei, Y., Zhang, Z., Lin, S., Guo, B.: Swin transformer: Hierarchical vision transformer using shifted windows. In: ICCV. pp. 10012--10022 (2021)

\bibitem{ma2016waterloo}
Ma, K., Duanmu, Z., Wu, Q., Wang, Z., Yong, H., Li, H., Zhang, L.: Waterloo exploration database: New challenges for image quality assessment models. IEEE TIP  \textbf{26}(2),  1004--1016 (2016)

\bibitem{martin2001database}
Martin, D., Fowlkes, C., Tal, D., Malik, J.: A database of human segmented natural images and its application to evaluating segmentation algorithms and measuring ecological statistics. In: ICCV. vol.~2 (2001)

\bibitem{matsui2017sketch}
Matsui, Y., Ito, K., Aramaki, Y., Fujimoto, A., Ogawa, T., Yamasaki, T., Aizawa, K.: Sketch-based manga retrieval using manga109 dataset. Multimedia Tools and Applications  \textbf{76},  21811--21838 (2017)

\bibitem{mei2021image}
Mei, Y., Fan, Y., Zhou, Y.: Image super-resolution with non-local sparse attention. In: CVPR. pp. 3517--3526 (2021)

\bibitem{muqeet2020multi}
Muqeet, A., Hwang, J., Yang, S., Kang, J., Kim, Y., Bae, S.H.: Multi-attention based ultra lightweight image super-resolution. In: ECCVW. pp. 103--118 (2020)

\bibitem{niu2020single}
Niu, B., Wen, W., Ren, W., Zhang, X., Yang, L., Wang, S., Zhang, K., Cao, X., Shen, H.: Single image super-resolution via a holistic attention network. In: ECCV. pp. 191--207 (2020)

\bibitem{paszke2017automatic}
Paszke, A., Gross, S., Chintala, S., Chanan, G., Yang, E., DeVito, Z., Lin, Z., Desmaison, A., Antiga, L., Lerer, A.: Automatic differentiation in pytorch  (2017)

\bibitem{ramachandran2019stand}
Ramachandran, P., Parmar, N., Vaswani, A., Bello, I., Levskaya, A., Shlens, J.: Stand-alone self-attention in vision models. NeurIPS  \textbf{32} (2019)

\bibitem{sheikh2006statistical}
Sheikh, H.R., Sabir, M.F., Bovik, A.C.: A statistical evaluation of recent full reference image quality assessment algorithms. IEEE TIP  \textbf{15}(11),  3440--3451 (2006)

\bibitem{sun2023safmn}
Sun, L., Dong, J., Tang, J., Pan, J.: Spatially-adaptive feature modulation for efficient image super-resolution. In: ICCV (2023)

\bibitem{timofte2017ntire}
Timofte, R., Agustsson, E., Van~Gool, L., Yang, M.H., Zhang, L.: Ntire 2017 challenge on single image super-resolution: Methods and results. In: CVPRW. pp. 114--125 (2017)

\bibitem{touvron2021training}
Touvron, H., Cord, M., Douze, M., Massa, F., Sablayrolles, A., J{\'e}gou, H.: Training data-efficient image transformers \& distillation through attention. In: ICML. pp. 10347--10357 (2021)

\bibitem{wang2020model}
Wang, H., Xie, Q., Zhao, Q., Meng, D.: A model-driven deep neural network for single image rain removal. In: CVPR. pp. 3103--3112 (2020)

\bibitem{wang2022uformer}
Wang, Z., Cun, X., Bao, J., Zhou, W., Liu, J., Li, H.: Uformer: A general u-shaped transformer for image restoration. In: CVPR. pp. 17683--17693 (2022)

\bibitem{wang2004image}
Wang, Z., Bovik, A.C., Sheikh, H.R., Simoncelli, E.P.: Image quality assessment: from error visibility to structural similarity. IEEE TIP  \textbf{13}(4),  600--612 (2004)

\bibitem{zagoruyko2017diracnets}
Zagoruyko, S., Komodakis, N.: Diracnets: Training very deep neural networks without skip-connections. arXiv preprint arXiv:1706.00388  (2017)

\bibitem{zamir2022restormer}
Zamir, S.W., Arora, A., Khan, S., Hayat, M., Khan, F.S., Yang, M.H.: Restormer: Efficient transformer for high-resolution image restoration. In: CVPR. pp. 5728--5739 (2022)

\bibitem{zeyde2012single}
Zeyde, R., Elad, M., Protter, M.: On single image scale-up using sparse-representations. In: Curves and Surfaces. pp. 711--730 (2012)

\bibitem{zhang2023lightweight}
Zhang, A., Ren, W., Liu, Y., Cao, X.: Lightweight image super-resolution with superpixel token interaction. In: ICCV. pp. 12728--12737 (2023)

\bibitem{zhang2022accurate}
Zhang, J., Zhang, Y., Gu, J., Zhang, Y., Kong, L., Yuan, X.: Accurate image restoration with attention retractable transformer (2023)

\bibitem{zhang2019aim}
Zhang, K., Gu, S., Timofte, R., Hui, Z., Wang, X., Gao, X., Xiong, D., Liu, S., Gang, R., Nan, N., et~al.: Aim 2019 challenge on constrained super-resolution: Methods and results. In: ICCVW. pp. 3565--3574 (2019)

\bibitem{zhang2017beyond}
Zhang, K., Zuo, W., Chen, Y., Meng, D., Zhang, L.: Beyond a gaussian denoiser: Residual learning of deep cnn for image denoising. TIP  \textbf{26}(7),  3142--3155 (2017)

\bibitem{zhang2018ffdnet}
Zhang, K., Zuo, W., Zhang, L.: Ffdnet: Toward a fast and flexible solution for cnn-based image denoising. IEEE TIP  \textbf{27}(9),  4608--4622 (2018)

\bibitem{zhang2011color}
Zhang, L., Wu, X., Buades, A., Li, X.: Color demosaicking by local directional interpolation and nonlocal adaptive thresholding. Journal of Electronic imaging  \textbf{20}(2),  023016--023016 (2011)

\bibitem{zhang2022efficient}
Zhang, X., Zeng, H., Guo, S., Zhang, L.: Efficient long-range attention network for image super-resolution. In: ECCV. pp. 649--667 (2022)

\bibitem{zhang2018image}
Zhang, Y., Li, K., Li, K., Wang, L., Zhong, B., Fu, Y.: Image super-resolution using very deep residual channel attention networks. In: ECCV. pp. 286--301 (2018)

\bibitem{zhang2019residual}
Zhang, Y., Li, K., Li, K., Zhong, B., Fu, Y.: Residual non-local attention networks for image restoration. arXiv preprint arXiv:1903.10082  (2019)

\bibitem{zheng2021rethinking}
Zheng, S., Lu, J., Zhao, H., Zhu, X., Luo, Z., Wang, Y., Fu, Y., Feng, J., Xiang, T., Torr, P.H., et~al.: Rethinking semantic segmentation from a sequence-to-sequence perspective with transformers. In: CVPR. pp. 6881--6890 (2021)

\bibitem{zhisheng2021efficient}
Zhisheng, L., Hong, L., Juncheng, L., Linlin, Z.: Efficient transformer for single image super-resolution. arXiv preprint arXiv:2108.11084  (2021)

\bibitem{zhou2023srformer}
Zhou, Y., Li, Z., Guo, C.L., Bai, S., Cheng, M.M., Hou, Q.: Srformer: Permuted self-attention for single image super-resolution. In: ICCV (2023)

\end{thebibliography}

\clearpage

\section{Appendix of ``LIPT: Latency-aware Image Processing Transformer''}

\subsection{More Details for LIPT Block}

\subsubsection{Window Expansion and Padding}
\begin{figure}[h]
	\centering
	\includegraphics[width=0.92\linewidth]{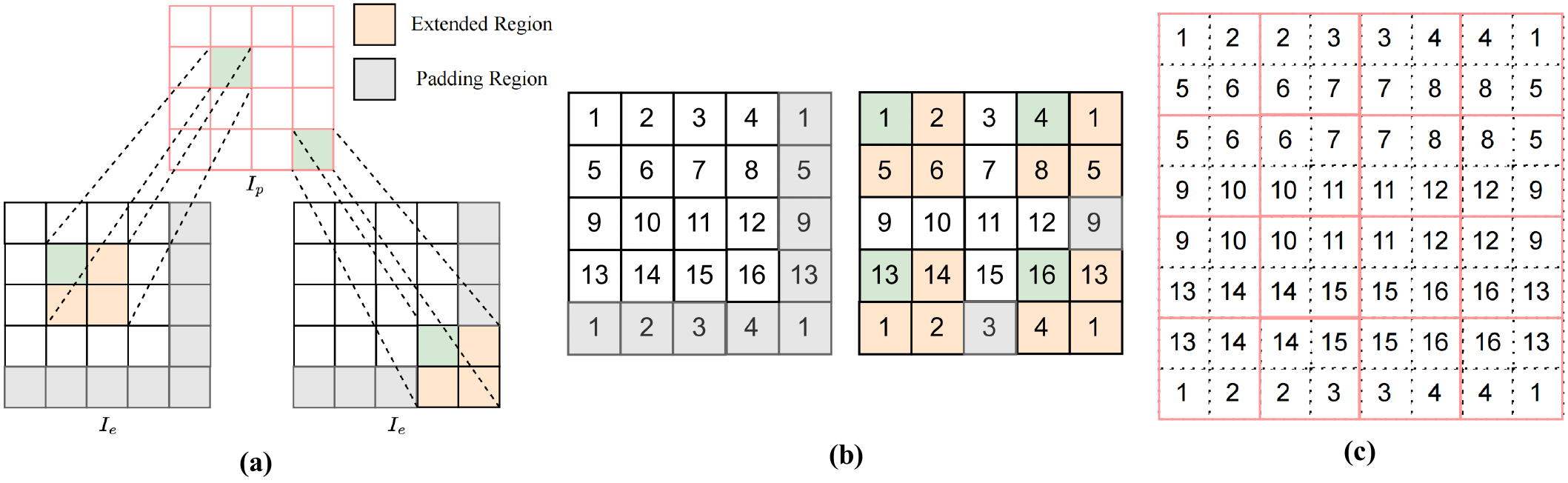}

	\caption{(a) Overview of window extension and padding $G_{we}$ by setting $s$ to 2. (b) Window padding. (c) The feature with a larger receptive field after $G_{we}$, where each pink box represents a new window.}
	\label{supfig:we_wp}
\end{figure}

Window expansion $G_{we}$ described in Eq. (8) of the main paper aims to expand each window to obtain a larger receptive field with an expansion size $s$ and padding. We take $s=2$ for example. 
As illustrated in Fig.~\ref{supfig:we_wp} (a), we expand each original green window $I_p^{(i,j)}$ to its larger window $I_e^{(i:i+s,j:j+s)}$ by selecting its adjacent $s^2$ windows at the bottom and right positions. Thus, 
%
%
the receptive field of the new large window is four times as that of the original window.

Note that we use the top and leftmost windows as the padding value for the outermost bottom and right windows to ensure the non-volatility of the upper-left corner window. Fig.~\ref{supfig:we_wp} (b) shows the window padding operation with $s=2$.
Fig.~\ref{supfig:we_wp} (c) shows the new feature map after window expansion and padding. The pink box represents the new large window, which obtains a wider receptive field. Obviously, all the small windows in $I_p^{(i,j)}$ will be sampled four times across the whole feature map, including the edge windows (\emph{e.g.}, $I_p^{(1,1)}$).


\subsubsection{Details of Isotropic Sobel Operation in HRM}

In Eq. (11) of the main paper, we design a high-frequency information extraction branch $G_{is}\circ conv1$ in HRM, which plays a vital role in low-level visual tasks. In $G_{is}$, we introduce an isotropic Sobel operation to extract more accurate local features. Given an input $I \in \mathbb{R}^{C\times H\times W}$, the feature $F_{is}$ processed by isotropic Sobel operation $G_{is}$ can be formulated as:
\begin{equation}
\begin{aligned}
& F_x=\left(s_x \cdot d_x\right) \otimes\left(k_x * I+b_x\right)+b_{d x}, \\
& F_y=\left(s_y \cdot d_y\right) \otimes\left(k_y * I+b_y\right)+b_{d y}, \\
& F_{is} = F_x + F_y,
\end{aligned}
\end{equation}
where $k_x$ and $b_x$ are the convolutional weight and bias with the kernels $1 \times 1$ in the horizontal direction. Correspondingly, $k_y$ and $b_y$ denote the weight and bias in the vertical direction.
$s_x/s_y$ and $b_{dx}/b_{dy}$ are the scaling parameters and bias with the shape of $C \times 1 \times 1 \times 1$. $\cdot$, $\otimes$ and $*$ denote the channel-wise broadcasting multiplication, depth-wise convolution and normal convolution, respectively. The $d_x$ and $d_y$ are the isotropic Sobel filters, which can be formulated as:
\begin{footnotesize}
\begin{equation}
d_x=\left[\begin{array}{ccc}
1 & 0 & -1 \\
\sqrt{2} & 0 & -\sqrt{2} \\
1 & 0 & -1
\end{array}\right] \text { and } d_y=\left[\begin{array}{ccc}
-1 & -\sqrt{2} & -1 \\
0 & 0 & 0 \\
1 & \sqrt{2} & 1
\end{array}\right].
\end{equation}
\end{footnotesize}
During the inference stage, the high-frequency branch can also be simplified to a $3 \times 3$ convolution.

\begin{table}[t]
\centering
\caption{The architectures of three-version LIPT.}
\begin{tabular}{c|c|c|c}
\hline
Options     & LIPT-Tiny & LIPT-Small & LIPT-Base \\ \hline
LIPT block  & 8         & 10         & 22        \\ \hline
Channel     & 24        & 64         & 144       \\ \hline
Window size & $8 \times 8$       & $8 \times 8$        & $16 \times 16$     \\ \hline
GPU (ms)    & 29        & 99         & 892       \\ \hline
CPU (s)  & 0.70        & 2.81         & 19.48       \\ \hline
Params(M)   & 0.36      & 2.33       & 26.05     \\ \hline
FLOPs(G)    & 69        & 455        & 5,211     \\ \hline
\end{tabular}

\label{tab:model}
\vspace{-12pt}
\end{table}

\subsection{Additional Experiments}
\subsubsection{Model Architecture and Training Hyper-parameter}
We provide three LIPT models, called LIPT-Tiny, LIPT-Small and LIPT-Base. 
Tab.~\ref{tab:model} presents the architecture details of these three LIPTs . 
LIPT-Tiny and LIPT-Small both adopt $8 \times 8$ window size, and consist of 8 LIPT blocks with the channel number of 24 and 10 LIPT blocks with the channel number of 64, respectively. 
For LIPT-Base, it has 22 LIPT blocks with the channel number of 144 and the window size of $16 \times 16$. 
%
We also report the running time, parameters and FLOPs of these three LIPT models for better comparison.

For CAR and denoising, their input image sizes are set to the same $128 \times 128$.
The batch sizes for CAR and denoising are set to 48 and 24, respectively. 
The number of total training iterations is set to 800K. The learning rate is initialized to $2 \times 10^{-4}$, which is halved at $[400K,600K,750K,775K]$. We employ randomly rotating $90^{\circ}$, $180^{\circ}$, $270^{\circ}$, and horizontal flip for training data augmentation for all tasks. We choose ADAM to optimize LIPT weights with $\beta_1 = 0.9$, $\beta_2 = 0.999$, and zero weight decay.

\subsubsection{More Visualization Results}
\begin{figure*}[t]
	\centering
	\begin{minipage}{0.265\linewidth}
		\vspace{-1pt}
		\centerline{\includegraphics[width=\textwidth]{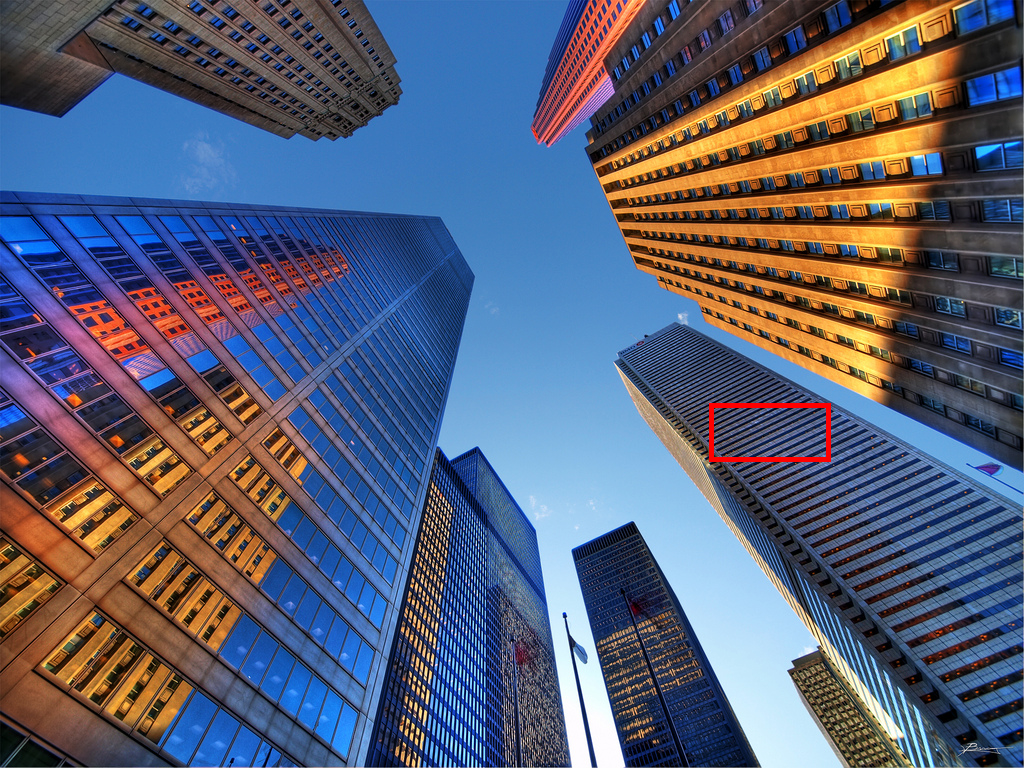}}
            \vspace{-0.cm}
		\centerline{\tiny img012 from Urban100}
            \vspace{3pt}
		
		\vspace{5pt}
		\centerline{\includegraphics[width=\textwidth]{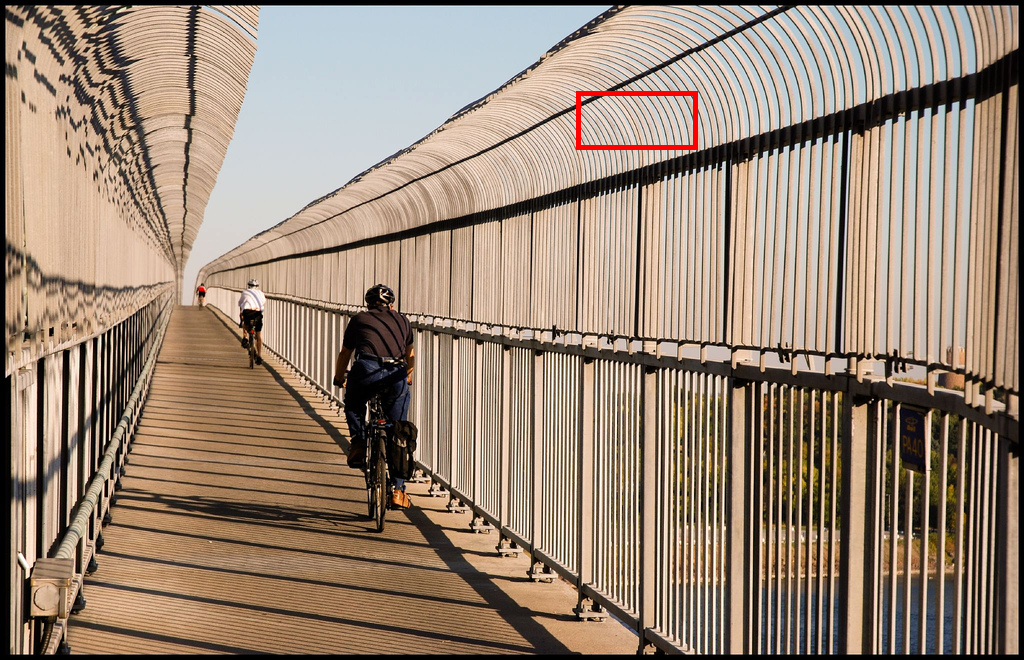}}
		\centerline{\tiny img024 from Urban100}
		\vspace{12pt}
		
		\centerline{\includegraphics[width=\textwidth]{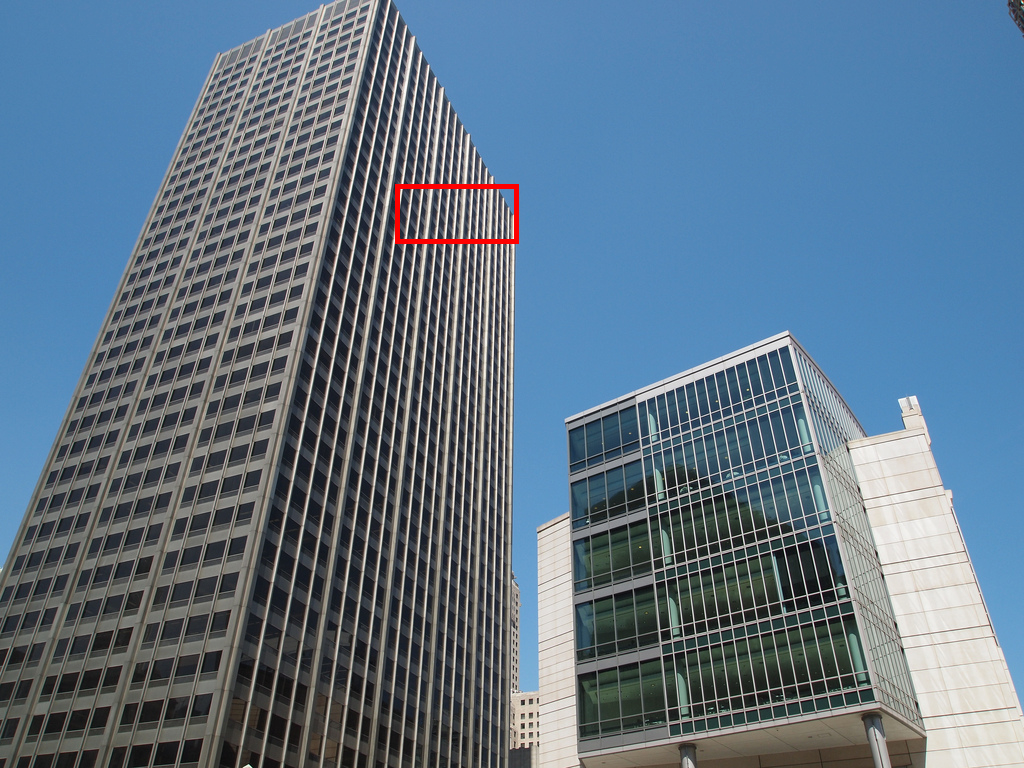}}
		\vspace{0pt}
		\centerline{\tiny img096 from Urban100}
		\vspace{8pt}
		
		\centerline{\includegraphics[width=\textwidth]{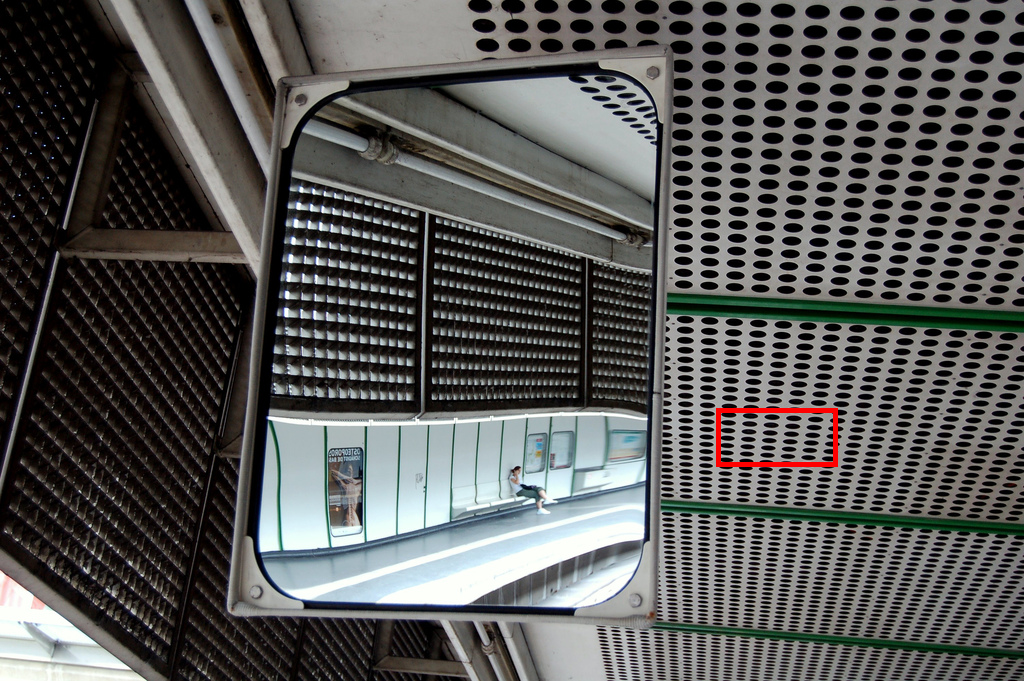}}
		\vspace{0pt}
		\centerline{\tiny img004 from Urban100}
		\vspace{8pt}
		
		\centerline{\includegraphics[width=\textwidth]{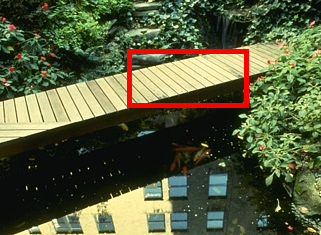}}
		\vspace{0pt}
		\centerline{\tiny img148024 from BSD100}
		\vspace{7pt}
	\end{minipage}
	\begin{minipage}{0.155\linewidth}
		\centerline{\includegraphics[width=\textwidth]{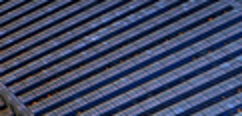}}
            \vspace{-0.15cm}
		\centerline{\tiny (a) HR}
            \vspace{-0.15cm}
            \centerline{\tiny PSNR/SSIM}
		\vspace{0.cm}
		\centerline{\includegraphics[width=\textwidth]{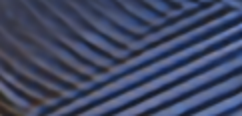}}
            \vspace{-0.15cm}
		\centerline{\tiny (e) LAPARA-A~\cite{li2020lapar}}
             \vspace{-0.15cm}
            \centerline{\tiny 20.02/0.4026}
		\vspace{0.15cm}
		
		\centerline{\includegraphics[width=\textwidth]{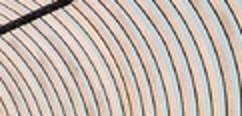}}
            \vspace{-0.18cm}
		\centerline{\tiny (a) HR}
            \vspace{-0.15cm}
            \centerline{\tiny PSNR/SSIM}
		\vspace{0.cm}
		\centerline{\includegraphics[width=\textwidth]{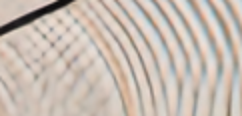}}
            \vspace{-0.15cm}
		\centerline{\tiny (e) LAPARA-A~\cite{li2020lapar}}
            \vspace{-0.15cm}
            \centerline{\tiny 20.11/0.4342}
		\vspace{0.2cm}
		
		\centerline{\includegraphics[width=\textwidth]{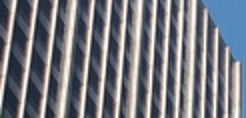}}
            \vspace{-0.15cm}
		\centerline{\tiny (a) HR}
            \vspace{-0.15cm}
            \centerline{\tiny PSNR/SSIM}
		\vspace{0.cm}
		\centerline{\includegraphics[width=\textwidth]{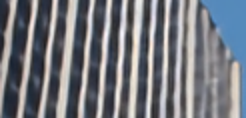}}
            \vspace{-0.15cm}
		\centerline{\tiny (e) LAPARA-A~\cite{li2020lapar}}
            \vspace{-0.15cm}
            \centerline{\tiny 20.68/0.7569}
		\vspace{0.1cm}
		
		\centerline{\includegraphics[width=\textwidth]{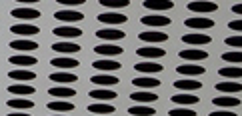}}
            \vspace{-0.15cm}
		\centerline{\tiny (a) HR}
            \vspace{-0.15cm}
            \centerline{\tiny PSNR/SSIM}
		\vspace{0.cm}
		\centerline{\includegraphics[width=\textwidth]{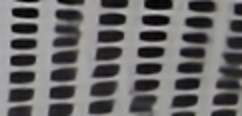}}
            \vspace{-0.15cm}
		\centerline{\tiny (e) LAPARA-A~\cite{li2020lapar}}
            \vspace{-0.15cm}
            \centerline{\tiny 15.7369/0.5260}
		\vspace{0.1cm}
		
		\centerline{\includegraphics[width=\textwidth]{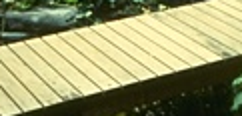}}
            \vspace{-0.15cm}
		\centerline{\tiny (a) HR}
            \vspace{-0.15cm}
            \centerline{\tiny PSNR/SSIM}
		\vspace{0.cm}
		\centerline{\includegraphics[width=\textwidth]{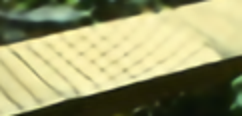}}
            \vspace{-0.15cm}
		\centerline{\tiny (e) LAPARA-A~\cite{li2020lapar}}
            \vspace{-0.15cm}
            \centerline{\tiny 24.34/0.6343}
		\vspace{0.27cm}

	\end{minipage}
	\begin{minipage}{0.155\linewidth}
		
		\centerline{\includegraphics[width=\textwidth]{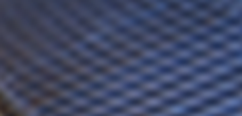}}
            \vspace{-0.15cm}
		\centerline{\tiny (b) Bicubic}
            \vspace{-0.15cm}
            \centerline{\tiny 19.16/0.2725}
		\vspace{0.cm}
		\centerline{\includegraphics[width=\textwidth]{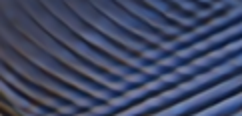}}
            \vspace{-0.15cm}
		\centerline{\tiny (f) IDMN~\cite{Hui-IMDN-2019}}
            \vspace{-0.15cm}
            \centerline{\tiny 18.94/0.3120}
		\vspace{0.15cm}
		
		\centerline{\includegraphics[width=\textwidth]{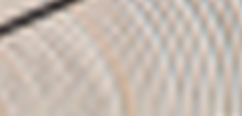}}
            \vspace{-0.18cm}
		\centerline{\tiny (b) Bicubic}
            \vspace{-0.15cm}
            \centerline{\tiny 18.52/0.2707}
		\vspace{0.cm}
		\centerline{\includegraphics[width=\textwidth]{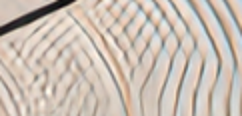}}
            \vspace{-0.15cm}
		\centerline{\tiny (f) IDMN~\cite{Hui-IMDN-2019}}
            \vspace{-0.15cm}
            \centerline{\tiny 19.13/0.4028}
		\vspace{0.2cm}
		
		\centerline{\includegraphics[width=\textwidth]{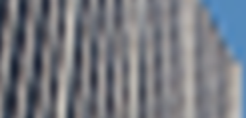}}
            \vspace{-0.15cm}
		\centerline{\tiny (b) Bicubic}
            \vspace{-0.15cm}
            \centerline{\tiny 15.21/0.3499}
		\vspace{0.cm}
		\centerline{\includegraphics[width=\textwidth]{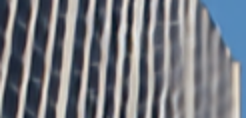}}
            \vspace{-0.15cm}
		\centerline{\tiny (f) IDMN~\cite{Hui-IMDN-2019}}
            \vspace{-0.15cm}
            \centerline{\tiny 19.42/0.7004}
		\vspace{0.1cm}
		
		\centerline{\includegraphics[width=\textwidth]{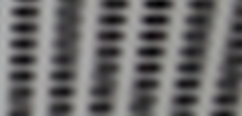}}
            \vspace{-0.15cm}
		\centerline{\tiny (b) Bicubic}
            \vspace{-0.15cm}
            \centerline{\tiny 15.6084/0.3995}
		\vspace{0.cm}
		\centerline{\includegraphics[width=\textwidth]{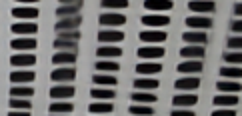}}
            \vspace{-0.15cm}
		\centerline{\tiny (f) IDMN~\cite{Hui-IMDN-2019}}
            \vspace{-0.15cm}
            \centerline{\tiny 17.53/0.6219}
		\vspace{0.1cm}
		
		\centerline{\includegraphics[width=\textwidth]{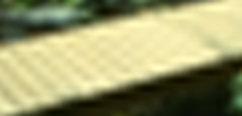}}
            \vspace{-0.15cm}
		\centerline{\tiny (b) Bicubic}
            \vspace{-0.15cm}
            \centerline{\tiny 21.46/0.4945}
		\vspace{0.cm}
		\centerline{\includegraphics[width=\textwidth]{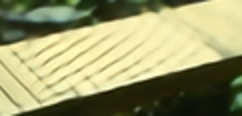}}
            \vspace{-0.15cm}
		\centerline{\tiny (f) IDMN~\cite{Hui-IMDN-2019}}
            \vspace{-0.15cm}
            \centerline{\tiny 23.02/0.5591}
		\vspace{0.27cm}
		
	\end{minipage}
	\begin{minipage}{0.155\linewidth}
		\centerline{\includegraphics[width=\textwidth]{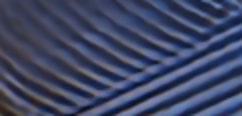}}
            \vspace{-0.15cm}
		\centerline{\tiny (c) CARN~\cite{ahn2018fast}}
            \vspace{-0.15cm}
            \centerline{\tiny 18.49/0.2774}
		\vspace{0.cm}
		\centerline{\includegraphics[width=\textwidth]{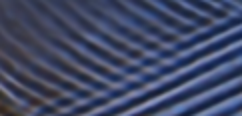}}
            \vspace{-0.15cm}
		\centerline{\tiny (g) SAFMN~\cite{sun2023safmn}}
            \vspace{-0.15cm}
            \centerline{\tiny 19.07/0.3000}
		\vspace{0.15cm}
		
		\centerline{\includegraphics[width=\textwidth]{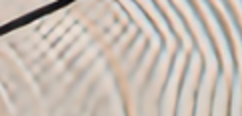}}
            \vspace{-0.18cm}
		\centerline{\tiny (c) CARN~\cite{ahn2018fast}}
            \vspace{-0.15cm}
            \centerline{\tiny 18.82/0.2935}
		\vspace{0.cm}
		\centerline{\includegraphics[width=\textwidth]{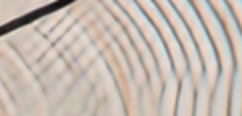}}
            \vspace{-0.15cm}
		\centerline{\tiny (g) SAFMN~\cite{sun2023safmn}}
            \vspace{-0.15cm}
            \centerline{\tiny 19.92/0.4227}
		\vspace{0.2cm}
		
		\centerline{\includegraphics[width=\textwidth]{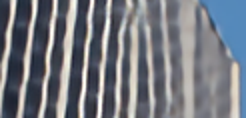}}
            \vspace{-0.15cm}
		\centerline{\tiny (c) CARN~\cite{ahn2018fast}}
            \vspace{-0.15cm}
            \centerline{\tiny 16.69/0.5531}
		\vspace{0.cm}
		\centerline{\includegraphics[width=\textwidth]{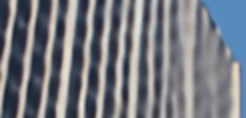}}
            \vspace{-0.15cm}
		\centerline{\tiny (g) SAFMN~\cite{sun2023safmn}}
            \vspace{-0.15cm}
            \centerline{\tiny 18.63/0.6537}
		\vspace{0.1cm}
		
		\centerline{\includegraphics[width=\textwidth]{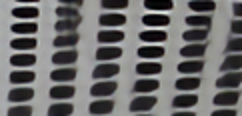}}
            \vspace{-0.15cm}
		\centerline{\tiny (c) CARN~\cite{ahn2018fast}}
            \vspace{-0.15cm}
            \centerline{\tiny 14.47/0.4196}
		\vspace{0.cm}
		\centerline{\includegraphics[width=\textwidth]{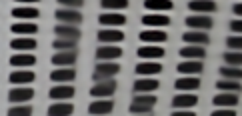}}
            \vspace{-0.15cm}
		\centerline{\tiny (g) SAFMN~\cite{sun2023safmn}}
            \vspace{-0.15cm}
            \centerline{\tiny 16.41/0.5473}
		\vspace{0.1cm}
		
		\centerline{\includegraphics[width=\textwidth]{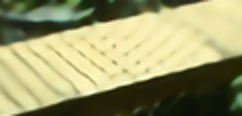}}
            \vspace{-0.15cm}
		\centerline{\tiny (c) CARN~\cite{ahn2018fast}}
            \vspace{-0.15cm}
            \centerline{\tiny 24.09/0.6370}
		\vspace{0.cm}
		\centerline{\includegraphics[width=\textwidth]{picture/supp/tiny_5/148026x4_test_LAPARA_SRx4_demo_patch.png}}
            \vspace{-0.15cm}
		\centerline{\tiny (g) SAFMN~\cite{sun2023safmn}}
            \vspace{-0.15cm}
            \centerline{\tiny 24.50/0.6399}
		\vspace{0.27cm}

	\end{minipage}
	\begin{minipage}{0.155\linewidth}
		\centerline{\includegraphics[width=\textwidth]{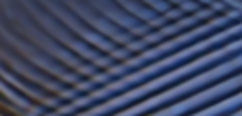}}
            \vspace{-0.15cm}
		\centerline{\tiny (d) EDSR-baseline~\cite{lim2017enhanced} }
            \vspace{-0.15cm}
            \centerline{\tiny 20.11/0.4018}
		\vspace{0.cm}
		\centerline{\includegraphics[width=\textwidth]{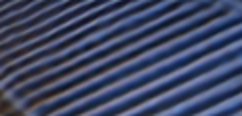}}
            \vspace{-0.15cm}
		\centerline{\tiny (i) LIPT-Tiny}
            \vspace{-0.15cm}
            \centerline{\textbf{\tiny 22.74/0.5774}}
		\vspace{0.15cm}
		
		\centerline{\includegraphics[width=\textwidth]{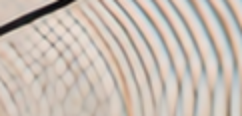}}
            \vspace{-0.18cm}
		\centerline{\tiny (d) EDSR-baseline~\cite{lim2017enhanced} }
            \vspace{-0.15cm}
            \centerline{\tiny 20.79/0.5087}
		\vspace{0.cm}
		\centerline{\includegraphics[width=\textwidth]{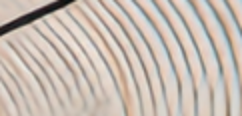}}
            \vspace{-0.15cm}
		\centerline{\tiny (i) LIPT-Tiny}
            \vspace{-0.15cm}
            \centerline{\textbf{\tiny 22.21/0.6367}}
		\vspace{0.2cm}
		
		\centerline{\includegraphics[width=\textwidth]{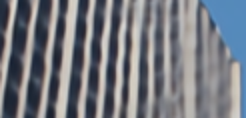}}
            \vspace{-0.15cm}
		\centerline{\tiny (d) EDSR-baseline~\cite{lim2017enhanced} }
            \vspace{-0.15cm}
            \centerline{\tiny 18.21/0.6347}
		\vspace{0.cm}
		\centerline{\includegraphics[width=\textwidth]{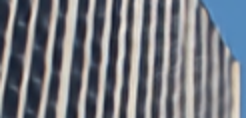}}
            \vspace{-0.15cm}
		\centerline{\tiny (i) LIPT-Tiny}
            \vspace{-0.15cm}
            \centerline{\textbf{\tiny 21.10/0.7762}}
		\vspace{0.1cm}
		
		\centerline{\includegraphics[width=\textwidth]{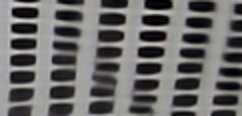}}
            \vspace{-0.15cm}
		\centerline{\tiny (d) EDSR-baseline~\cite{lim2017enhanced} }
            \vspace{-0.15cm}
            \centerline{\tiny 17.29/0.6171}
		\vspace{0.cm}
		\centerline{\includegraphics[width=\textwidth]{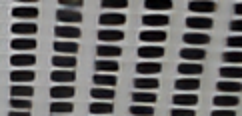}}
            \vspace{-0.15cm}
		\centerline{\tiny (i) LIPT-Tiny}
            \vspace{-0.15cm}
            \centerline{\textbf{\tiny 19.84/0.7074}}
		\vspace{0.1cm}
		
		\centerline{\includegraphics[width=\textwidth]{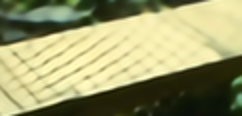}}
            \vspace{-0.15cm}
		\centerline{\tiny (d) EDSR-baseline~\cite{lim2017enhanced} }
            \vspace{-0.15cm}
            \centerline{\tiny 23.31/0.5747}
		\vspace{0.cm}
		\centerline{\includegraphics[width=\textwidth]{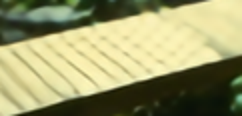}}
            \vspace{-0.15cm}
		\centerline{\tiny (i) LIPT-Tiny}
            \vspace{-0.15cm}
            \centerline{\textbf{\tiny 24.71/0.6737}} 
		\vspace{0.27cm}
	\end{minipage}
	
	\caption{Qualitative Comparison between LIPT-Tiny and other methods on Urban100 and BSD100 for $\times$4 SR. Zoom in for best views.}
	\vspace{-0.8em}
	\label{fig:Tiny}
	\vspace{-0.6em}
\end{figure*}

\begin{figure*}[t]
	\centering
	\begin{minipage}{0.27\linewidth}
		\vspace{2pt}
		\centerline{\includegraphics[width=\textwidth]{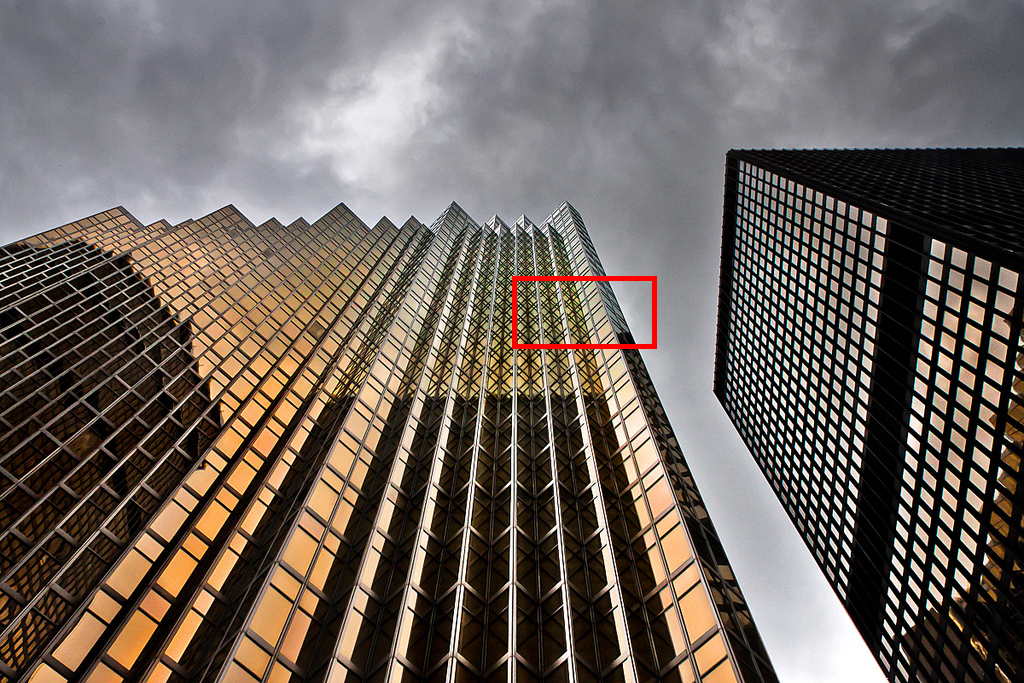}}
		\centerline{\tiny img019 from Urban100}
		\vspace{6pt}
		\centerline{\includegraphics[width=\textwidth]{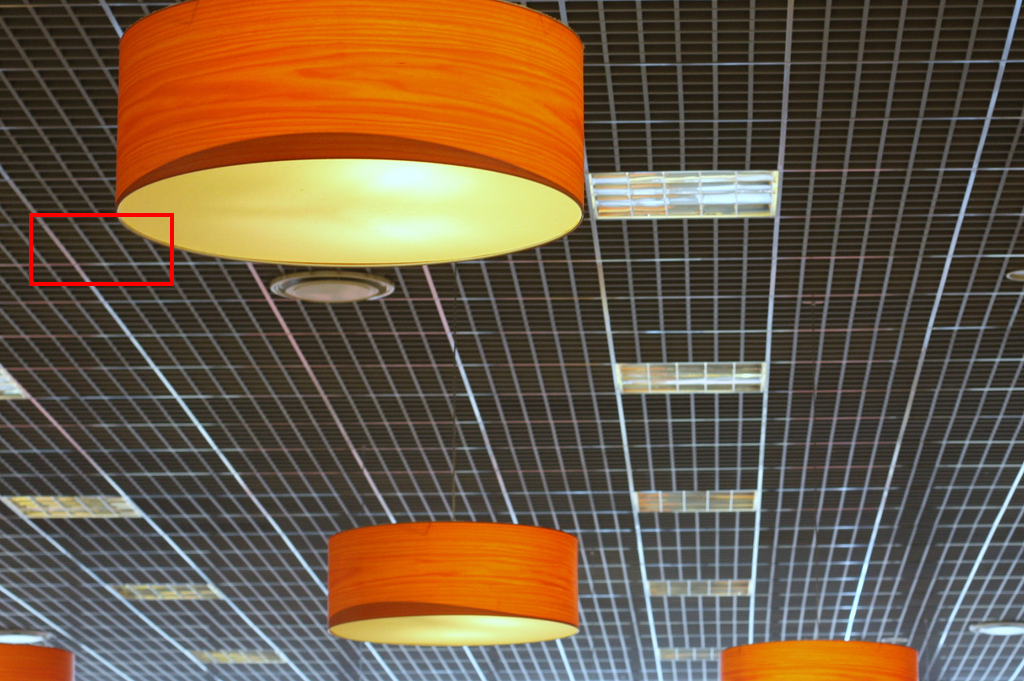}}
		\centerline{\tiny img044 from Urban100}
		\vspace{2pt}
		\centerline{\includegraphics[width=\textwidth]{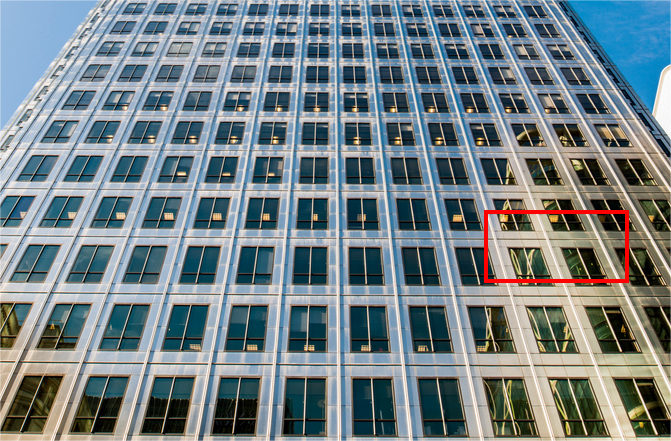}}
		\centerline{\tiny img030 from Urban100}
		\vspace{6pt}
		\centerline{\includegraphics[width=\textwidth]{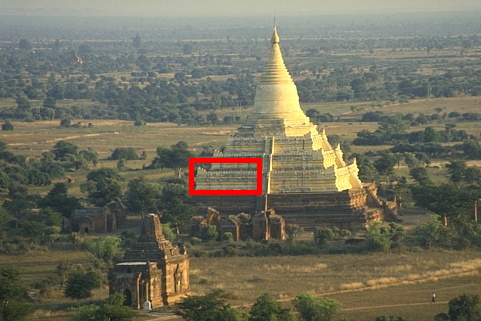}}
		\centerline{\tiny img76053 from BSD100}
		\vspace{6pt}
		\centerline{\includegraphics[width=\textwidth]{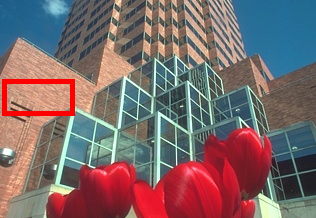}}
		\centerline{\tiny img86000 from BSD100}
		\vspace{6pt}
	\end{minipage}
	\begin{minipage}{0.13\linewidth}
		\centerline{\includegraphics[width=\textwidth]{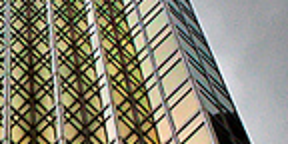}}
            \vspace{-0.15cm}
		\centerline{\tiny (a) HR}
            \vspace{-0.15cm}
            \centerline{\tiny PSNR/SSIM}
		\vspace{0.07cm}
		\centerline{\includegraphics[width=\textwidth]{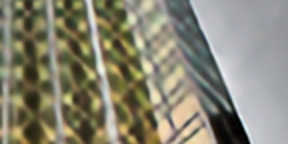}}
            \vspace{-0.15cm}
		\centerline{\tiny (f) SAFMN~\cite{sun2023safmn} }
            \vspace{-0.15cm}
            \centerline{\tiny 17.32/0.4398}
		\vspace{0.1cm}
		\centerline{\includegraphics[width=\textwidth]{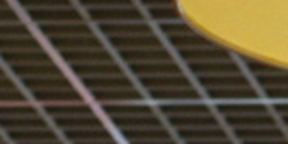}}
            \vspace{-0.15cm}
		\centerline{\tiny (a) HR}
            \vspace{-0.15cm}
            \centerline{\tiny PSNR/SSIM}
		\centerline{\includegraphics[width=\textwidth]{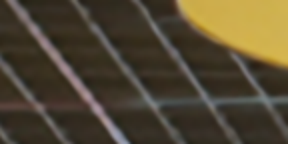}}
            \vspace{-0.15cm}
		\centerline{\tiny (f) SAFMN~\cite{sun2023safmn} }
            \vspace{-0.15cm}
            \centerline{\tiny 29.30/0.7862}
		
		\centerline{\includegraphics[width=\textwidth]{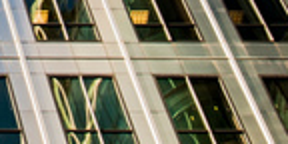}}
            \vspace{-0.15cm}
		\centerline{\tiny (a) HR}
            \vspace{-0.15cm}
            \centerline{\tiny PSNR/SSIM}
		\vspace{0.07cm}
		\centerline{\includegraphics[width=\textwidth]{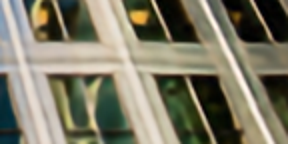}}
            \vspace{-0.15cm}
		\centerline{\tiny (f) SAFMN~\cite{sun2023safmn} }
            \vspace{-0.15cm}
            \centerline{\tiny 23.40/0.6757}
		\vspace{0.1cm}
		
		\centerline{\includegraphics[width=\textwidth]{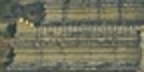}}
            \vspace{-0.15cm}
		\centerline{\tiny (a) HR}
            \vspace{-0.15cm}
            \centerline{\tiny PSNR/SSIM}
		\vspace{0.07cm}
		\centerline{\includegraphics[width=\textwidth]{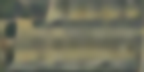}}
            \vspace{-0.15cm}
		\centerline{\tiny (f) SAFMN~\cite{sun2023safmn} }
            \vspace{-0.15cm}
            \centerline{\tiny 26.72/0.5351}
		\vspace{0.1cm}
		
		\centerline{\includegraphics[width=\textwidth]{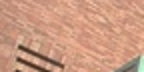}}
            \vspace{-0.15cm}
		\centerline{\tiny (a) HR}
            \vspace{-0.15cm}
            \centerline{\tiny PSNR/SSIM}
		\vspace{0.07cm}
		\centerline{\includegraphics[width=\textwidth]{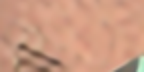}}
            \vspace{-0.15cm}
		\centerline{\tiny (f) SAFMN~\cite{sun2023safmn} }
            \vspace{-0.15cm}
            \centerline{\tiny 26.54/0.6155}
		\vspace{0.1cm}

	\end{minipage}
	\begin{minipage}{0.13\linewidth}
		\centerline{\includegraphics[width=\textwidth]{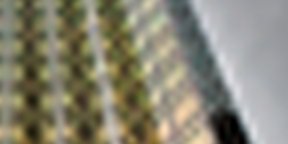}}
            \vspace{-0.15cm}
		\centerline{\tiny (b) Bicubic}
            \vspace{-0.15cm}
            \centerline{\tiny 16.56/0.3218}
		\vspace{0.07cm}
		\centerline{\includegraphics[width=\textwidth]{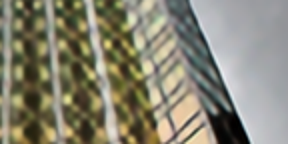}}
            \vspace{-0.15cm}
		\centerline{\tiny (g) ESRT~\cite{zhisheng2021efficient}}
            \vspace{-0.15cm}
            \centerline{\tiny 17.48/0.4624}
		\vspace{0.1cm}
		\centerline{\includegraphics[width=\textwidth]{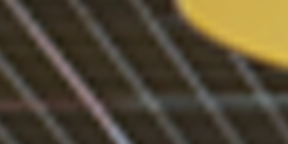}}
            \vspace{-0.15cm}
		\centerline{\tiny (b) Bicubic}
            \vspace{-0.15cm}
            \centerline{\tiny 27.59/0.7248}
		\centerline{\includegraphics[width=\textwidth]{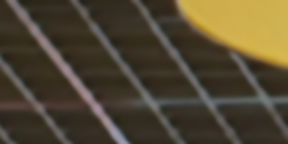}}
            \vspace{-0.15cm}
		\centerline{\tiny (g) ESRT~\cite{zhisheng2021efficient}}
            \vspace{-0.15cm}
            \centerline{\tiny 29.85/0.8025}
		
		\centerline{\includegraphics[width=\textwidth]{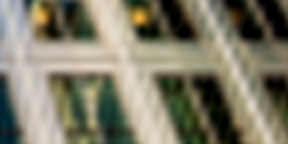}}
            \vspace{-0.15cm}
		\centerline{\tiny (b) Bicubic}
            \vspace{-0.15cm}
            \centerline{\tiny 20.14/0.4740}
		\vspace{0.07cm}
		\centerline{\includegraphics[width=\textwidth]{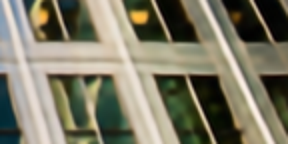}}
            \vspace{-0.15cm}
		\centerline{\tiny (g) ESRT~\cite{zhisheng2021efficient}}
            \vspace{-0.15cm}
            \centerline{\tiny 24.07/0.7092}
		\vspace{0.1cm}
		
		\centerline{\includegraphics[width=\textwidth]{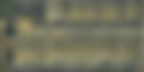}}
            \vspace{-0.15cm}
		\centerline{\tiny (b) Bicubic}
            \vspace{-0.15cm}
            \centerline{\tiny 24.43/0.4482}
		\vspace{0.07cm}
		\centerline{\includegraphics[width=\textwidth]{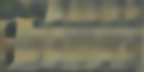}}
            \vspace{-0.15cm}
		\centerline{\tiny (g) ESRT~\cite{zhisheng2021efficient}}
            \vspace{-0.15cm}
            \centerline{\tiny 26.96/0.5530}
		\vspace{0.1cm}
		
		\centerline{\includegraphics[width=\textwidth]{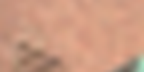}}
            \vspace{-0.15cm}
		\centerline{\tiny (b) Bicubic}
            \vspace{-0.15cm}
            \centerline{\tiny 25.01/0.5652}
		\vspace{0.07cm}
		\centerline{\includegraphics[width=\textwidth]{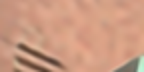}}
            \vspace{-0.15cm}
		\centerline{\tiny (g) ESRT~\cite{zhisheng2021efficient}}
            \vspace{-0.15cm}
            \centerline{\tiny 28.71/0.6407}
		\vspace{0.1cm}

	\end{minipage}
	\begin{minipage}{0.13\linewidth}
		\centerline{\includegraphics[width=\textwidth]{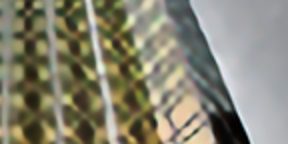}}
            \vspace{-0.15cm}
		\centerline{\tiny (c) CARN~\cite{ahn2018fast}}
            \vspace{-0.15cm}
            \centerline{\tiny 17.17/0.4197}
		\vspace{0.07cm}
		\centerline{\includegraphics[width=\textwidth]{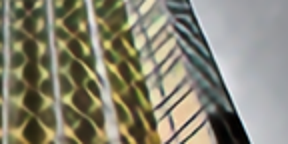}}
            \vspace{-0.15cm}
		\centerline{\tiny (h) SwinIR-L~\cite{liang2021swinir}}
            \vspace{-0.15cm}
            \centerline{\tiny 17.49/0.4897}
		\vspace{0.1cm}
		\centerline{\includegraphics[width=\textwidth]{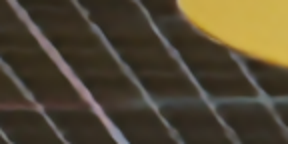}}
            \vspace{-0.15cm}
		\centerline{\tiny (c) CARN~\cite{ahn2018fast}}
            \vspace{-0.15cm}
            \centerline{\tiny 28.57/0.7605}
		\centerline{\includegraphics[width=\textwidth]{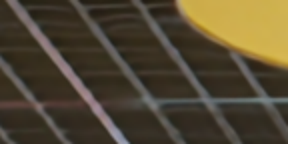}}
            \vspace{-0.15cm}
		\centerline{\tiny (h) SwinIR-L~\cite{liang2021swinir}}
            \vspace{-0.15cm}
            \centerline{\tiny 30.24/0.8132}
		
		\centerline{\includegraphics[width=\textwidth]{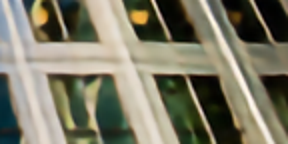}}
            \vspace{-0.15cm}
		\centerline{\tiny (c) CARN~\cite{ahn2018fast}}
            \vspace{-0.15cm}
            \centerline{\tiny 23.54/0.6862}
		\vspace{0.07cm}
		\centerline{\includegraphics[width=\textwidth]{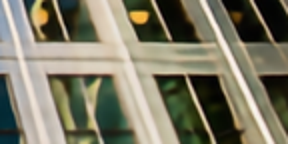}}
            \vspace{-0.15cm}
		\centerline{\tiny (h) SwinIR-L~\cite{liang2021swinir}}
            \vspace{-0.15cm}
            \centerline{\tiny 24.32/0.7235}
		\vspace{0.1cm}
		
		\centerline{\includegraphics[width=\textwidth]{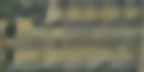}}
            \vspace{-0.15cm}
		\centerline{\tiny (c) CARN~\cite{ahn2018fast}}
            \vspace{-0.15cm}
            \centerline{\tiny 26.75/0.5359}
		\vspace{0.07cm}
		\centerline{\includegraphics[width=\textwidth]{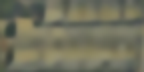}}
            \vspace{-0.15cm}
		\centerline{\tiny (h) SwinIR-L~\cite{liang2021swinir}}
            \vspace{-0.15cm}
            \centerline{\tiny 26.27/0.4970}
		\vspace{0.1cm}
		
		\centerline{\includegraphics[width=\textwidth]{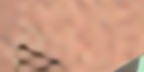}}
            \vspace{-0.15cm}
		\centerline{\tiny (c) CARN~\cite{ahn2018fast}}
            \vspace{-0.15cm}
            \centerline{\tiny 26.22/0.6089}
		\vspace{0.07cm}
		\centerline{\includegraphics[width=\textwidth]{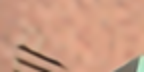}}
            \vspace{-0.15cm}
		\centerline{\tiny (h) SwinIR-L~\cite{liang2021swinir}}
            \vspace{-0.15cm}
            \centerline{\tiny 29.12/0.6460}
		\vspace{0.1cm}

	\end{minipage}
	\begin{minipage}{0.13\linewidth}
		\centerline{\includegraphics[width=\textwidth]{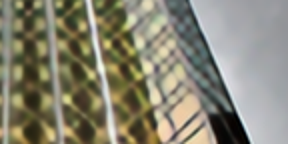}}
            \vspace{-0.15cm}
		\centerline{\tiny (d) EDSR-b~\cite{lim2017enhanced} }
            \vspace{-0.15cm}
            \centerline{\tiny 17.33/0.4471}
		\vspace{0.07cm}
		\centerline{\includegraphics[width=\textwidth]{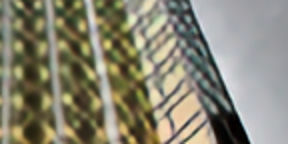}}
            \vspace{-0.15cm}
		\centerline{\tiny (i) N-Gram~\cite{choi2023n}}
            \vspace{-0.15cm}
            \centerline{\tiny 17.43/0.4615}
		\vspace{0.1cm}
		\centerline{\includegraphics[width=\textwidth]{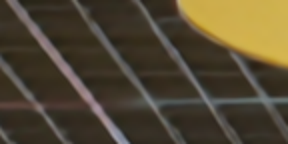}}
            \vspace{-0.15cm}
		\centerline{\tiny (d) EDSR-b~\cite{lim2017enhanced} }
            \vspace{-0.15cm}
            \centerline{\tiny 29.80/0.8025}
		\centerline{\includegraphics[width=\textwidth]{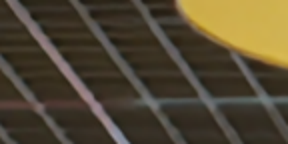}}
            \vspace{-0.15cm}
		\centerline{\tiny (i) N-Gram~\cite{choi2023n}}
            \vspace{-0.15cm}
            \centerline{\tiny 30.50/0.8367}
		
		\centerline{\includegraphics[width=\textwidth]{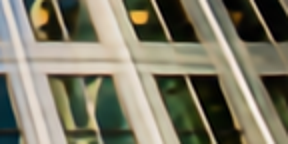}}
            \vspace{-0.15cm}
		\centerline{\tiny (d) EDSR-b~\cite{lim2017enhanced} }
            \vspace{-0.15cm}
            \centerline{\tiny 23.89/0.6961}
		\vspace{0.07cm}
		\centerline{\includegraphics[width=\textwidth]{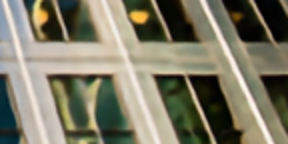}}
            \vspace{-0.15cm}
		\centerline{\tiny (i) N-Gram~\cite{choi2023n}}
            \vspace{-0.15cm}
            \centerline{\tiny 23.93/0.7231}
		\vspace{0.1cm}
		
		\centerline{\includegraphics[width=\textwidth]{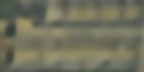}}
            \vspace{-0.15cm}
		\centerline{\tiny (d) EDSR-b~\cite{lim2017enhanced} }
            \vspace{-0.15cm}
            \centerline{\tiny 26.88/0.5445}
		\vspace{0.07cm}
		\centerline{\includegraphics[width=\textwidth]{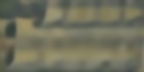}}
            \vspace{-0.15cm}
		\centerline{\tiny (i) N-Gram~\cite{choi2023n}}
            \vspace{-0.15cm}
            \centerline{\tiny 26.75/0.5425}
		\vspace{0.1cm}
		
		\centerline{\includegraphics[width=\textwidth]{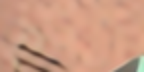}}
            \vspace{-0.15cm}
		\centerline{\tiny (d) EDSR-b~\cite{lim2017enhanced} }
            \vspace{-0.15cm}
            \centerline{\tiny 28.09/0.6333}
		\vspace{0.07cm}
		\centerline{\includegraphics[width=\textwidth]{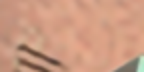}}
            \vspace{-0.15cm}
		\centerline{\tiny (i) N-Gram~\cite{choi2023n}}
            \vspace{-0.15cm}
            \centerline{\tiny 28.25/0.6366}
		\vspace{0.1cm}

	\end{minipage}
	\begin{minipage}{0.13\linewidth}
		\centerline{\includegraphics[width=\textwidth]{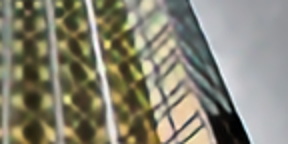}}
            \vspace{-0.15cm}
		\centerline{\tiny (e) LAPAR-A~\cite{li2020lapar}}
            \vspace{-0.15cm}
            \centerline{\tiny 17.32/0.4577}
		\vspace{0.07cm}
		\centerline{\includegraphics[width=\textwidth]{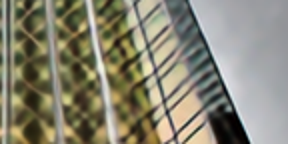}}
            \vspace{-0.15cm}
		\centerline{\tiny (j) LIPT-Small}
            \vspace{-0.15cm}
            \centerline{\textbf{\tiny 17.86/0.4965}}
		\vspace{0.1cm}
		\centerline{\includegraphics[width=\textwidth]{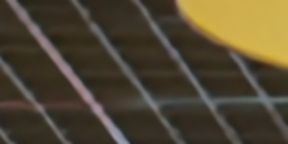}}
            \vspace{-0.15cm}
		\centerline{\tiny (e) LAPAR-A~\cite{li2020lapar}}
            \vspace{-0.15cm}
            \centerline{\tiny 29.35/0.7880}
		\centerline{\includegraphics[width=\textwidth]{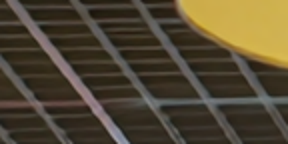}}
            \vspace{-0.15cm}
		\centerline{\tiny (j) LIPT-Small}
            \vspace{-0.15cm}
            \centerline{\textbf{\tiny 31.62/0.8581}}
		
		\centerline{\includegraphics[width=\textwidth]{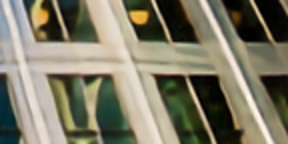}}
            \vspace{-0.15cm}
		\centerline{\tiny (e) LAPAR-A~\cite{li2020lapar}}
            \vspace{-0.15cm}
            \centerline{\tiny 23.62/0.6884}
		\vspace{0.07cm}
		\centerline{\includegraphics[width=\textwidth]{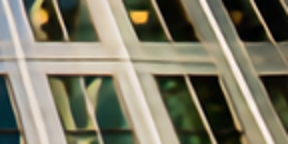}}
            \vspace{-0.15cm}
		\centerline{\tiny (j) LIPT-Small}
            \vspace{-0.15cm}
            \centerline{\textbf{\tiny 24.27/0.7295}}
		\vspace{0.1cm}
		
		\centerline{\includegraphics[width=\textwidth]{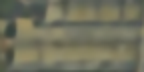}}
            \vspace{-0.15cm}
		\centerline{\tiny (e) LAPAR-A~\cite{li2020lapar}}
            \vspace{-0.15cm}
            \centerline{\tiny 27.02/0.5648}
		\vspace{0.07cm}
		\centerline{\includegraphics[width=\textwidth]{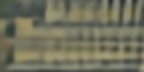}}
            \vspace{-0.15cm}
		\centerline{\tiny (j) LIPT-Small}
            \vspace{-0.15cm}
            \centerline{\textbf{\tiny 27.23/0.5685}}
		\vspace{0.1cm}
		
		\centerline{\includegraphics[width=\textwidth]{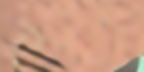}}
            \vspace{-0.15cm}
		\centerline{\tiny (e) LAPAR-A~\cite{li2020lapar}}
            \vspace{-0.15cm}
            \centerline{\tiny 28.26/0.6351}
		\vspace{0.07cm}
		\centerline{\includegraphics[width=\textwidth]{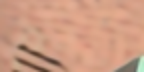}}
            \vspace{-0.15cm}
		\centerline{\tiny (j) LIPT-Small}
            \vspace{-0.15cm}
            \centerline{\textbf{\tiny 29.26/0.6733}}
		\vspace{0.1cm}

	\end{minipage}
	\vspace{-0.5em}
	\caption{Qualitative Comparison between LIPT-Small and other methods on Urban100 and BSD100 for $\times$4 SR. Zoom in for best views. SwinIR-L and EDSR-b are denote SwinIR-Light and EDSR-baseline, respectively.}
	\label{supfig:light}
\end{figure*}

\begin{figure*}[t]
	\centering
	\begin{minipage}{0.3\linewidth}
		\vspace{2pt}
		\centerline{\includegraphics[width=\textwidth]{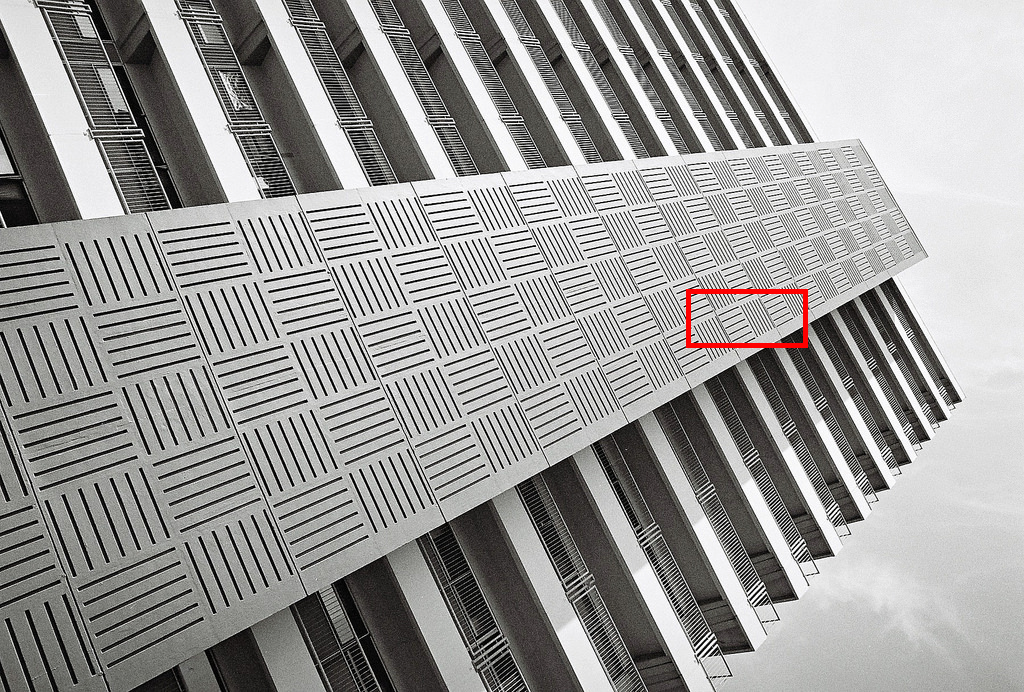}}
		\centerline{\tiny img092 from Urban100}
		\vspace{6pt}
		
		\vspace{2pt}
		\centerline{\includegraphics[width=\textwidth]{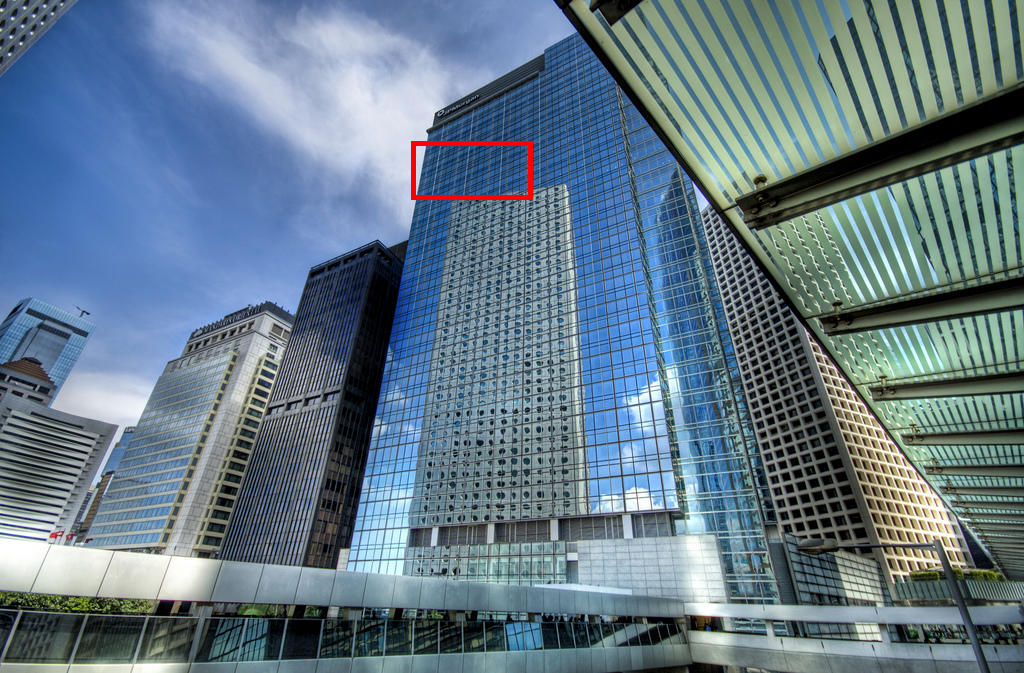}}
		\centerline{\tiny img061 from Urban100}
		\vspace{6pt}
		
		\vspace{2pt}
		\centerline{\includegraphics[width=\textwidth]{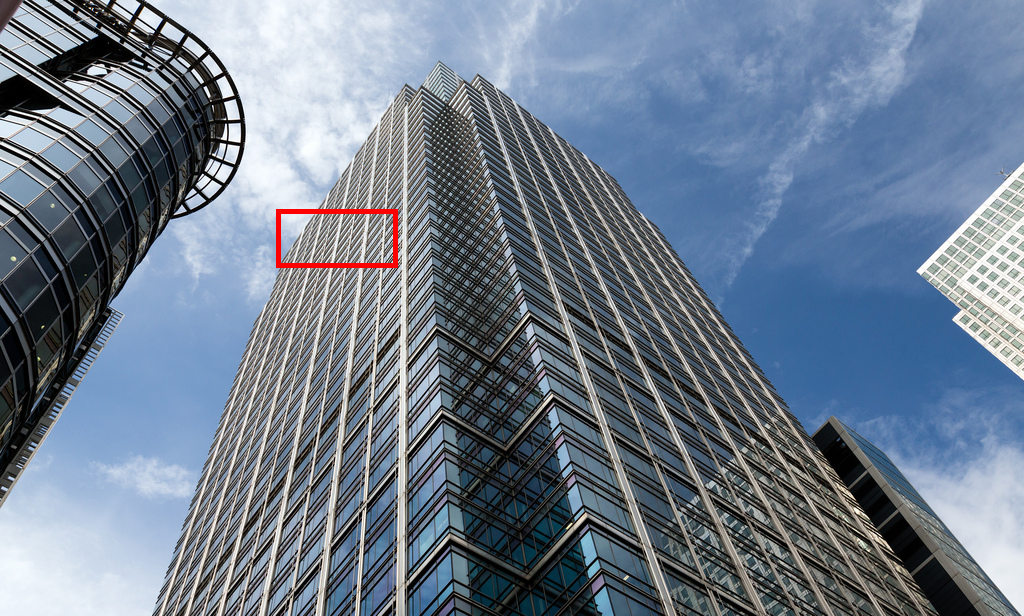}}
		\centerline{\tiny img047 from Urban100}
		\vspace{8pt}
		
		\vspace{6pt}
		\centerline{\includegraphics[width=\textwidth]{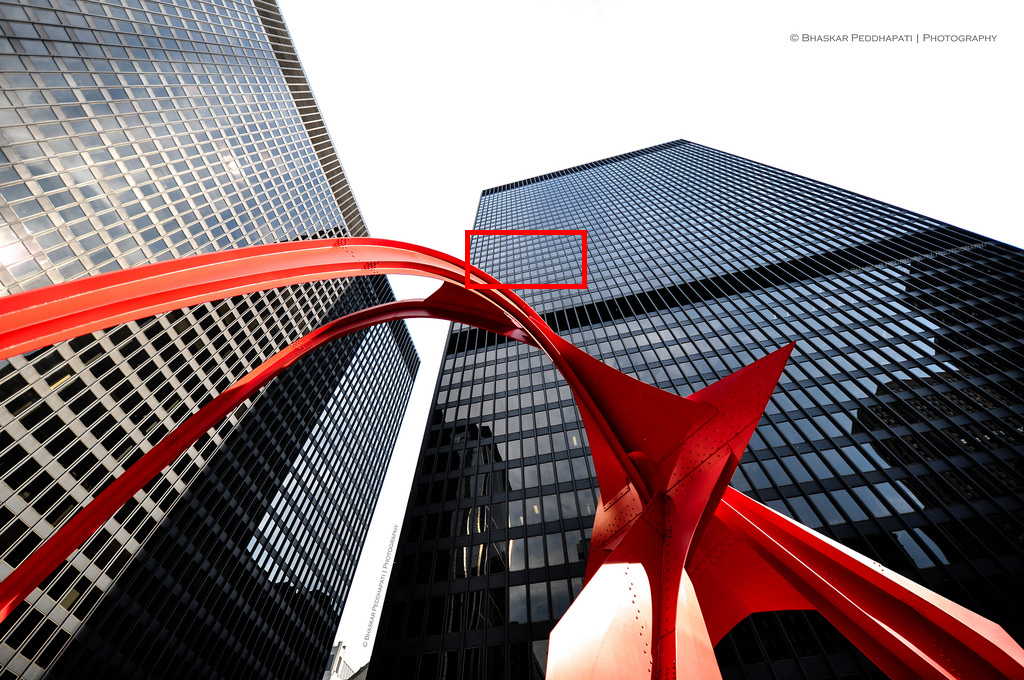}}
		\centerline{\tiny img062 from Urban100}
		\vspace{2pt}
		
		\vspace{2pt}
		\centerline{\includegraphics[width=\textwidth]{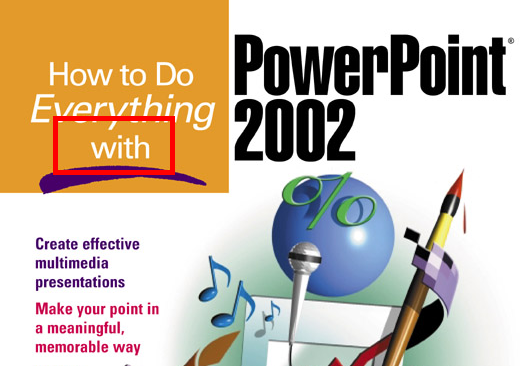}}
		\centerline{\tiny imgppt3 from Set14}
		\vspace{6pt}
	\end{minipage}
	\begin{minipage}{0.16\linewidth}
		\centerline{\includegraphics[width=\textwidth]{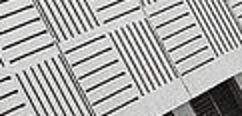}}
            \vspace{-0.15cm}
		\centerline{\tiny (a) HR}
            \vspace{-0.15cm}
            \centerline{\tiny PSNR/SSIM}
		\vspace{0.07cm}
		\centerline{\includegraphics[width=\textwidth]{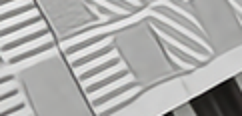}}
            \vspace{-0.15cm}
		\centerline{\tiny (e) SwinIR~\cite{liang2021swinir}}
            \vspace{-0.15cm}
            \centerline{\tiny 15.25/0.2179}
		\vspace{0.1cm}
		
		\centerline{\includegraphics[width=\textwidth]{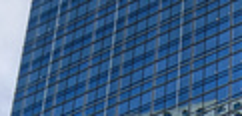}}
            \vspace{-0.15cm}
		\centerline{\tiny (a) HR}
            \vspace{-0.15cm}
            \centerline{\tiny PSNR/SSIM}
		\vspace{0.07cm}
		\centerline{\includegraphics[width=\textwidth]{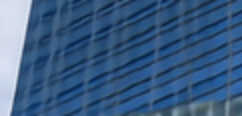}}
            \vspace{-0.15cm}
		\centerline{\tiny (e) SwinIR~\cite{liang2021swinir}}
            \vspace{-0.15cm}
            \centerline{\tiny 27.85/0.6312}
		\vspace{0.1cm}
		
		\centerline{\includegraphics[width=\textwidth]{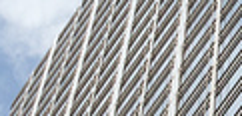}}
            \vspace{-0.15cm}
		\centerline{\tiny (a) HR}
            \vspace{-0.15cm}
            \centerline{\tiny PSNR/SSIM}
		\vspace{0.07cm}
		\centerline{\includegraphics[width=\textwidth]{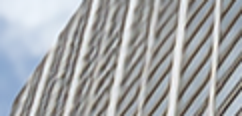}}
            \vspace{-0.15cm}
		\centerline{\tiny (e) SwinIR~\cite{liang2021swinir}}
            \vspace{-0.15cm}
            \centerline{\tiny 18.99/0.5393}
		\vspace{0.1cm}
		
		\centerline{\includegraphics[width=\textwidth]{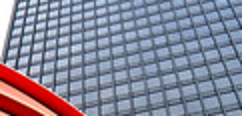}}
            \vspace{-0.15cm}
		\centerline{\tiny (a) HR}
            \vspace{-0.15cm}
            \centerline{\tiny PSNR/SSIM}
		\vspace{0.07cm}
		\centerline{\includegraphics[width=\textwidth]{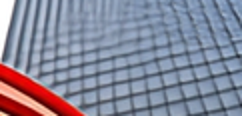}}
            \vspace{-0.15cm}
		\centerline{\tiny (e) SwinIR~\cite{liang2021swinir}}
            \vspace{-0.15cm}
            \centerline{\tiny 21.5690/0.5749}
		\vspace{0.1cm}
		
		\centerline{\includegraphics[width=\textwidth]{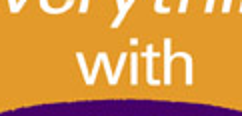}}
            \vspace{-0.15cm}
		\centerline{\tiny (a) HR}
            \vspace{-0.15cm}
            \centerline{\tiny PSNR/SSIM}
		\vspace{0.07cm}
		\centerline{\includegraphics[width=\textwidth]{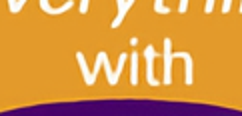}}
            \vspace{-0.15cm}
		\centerline{\tiny (e) SwinIR~\cite{liang2021swinir}}
            \vspace{-0.15cm}
            \centerline{\tiny 35.69/0.9719}
		\vspace{0.1cm}

	\end{minipage}
	\begin{minipage}{0.16\linewidth}
		
		\centerline{\includegraphics[width=\textwidth]{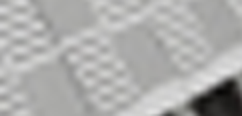}}
            \vspace{-0.15cm}
		\centerline{\tiny (b) Bicubic}
            \vspace{-0.15cm}
            \centerline{\tiny 14.88/0.1527}
		\vspace{0.07cm}
		\centerline{\includegraphics[width=\textwidth]{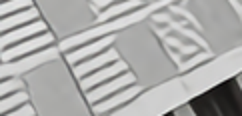}}
            \vspace{-0.15cm}
		\centerline{\tiny (f) SRFormer~\cite{zhou2023srformer}}
            \vspace{-0.15cm}
            \centerline{\tiny 15.95/0.3003}
		\vspace{0.1cm}
		
		\centerline{\includegraphics[width=\textwidth]{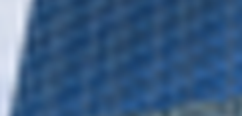}}
            \vspace{-0.15cm}
		\centerline{\tiny (b) Bicubic}
            \vspace{-0.15cm}
            \centerline{\tiny 25.96/0.4879}
		\vspace{0.07cm}
		\centerline{\includegraphics[width=\textwidth]{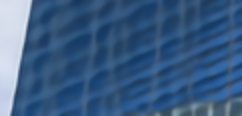}}
            \vspace{-0.15cm}
		\centerline{\tiny (f) SRFormer~\cite{zhou2023srformer}}
            \vspace{-0.15cm}
            \centerline{\tiny 28.05/0.6052}
		\vspace{0.1cm}
		
		\centerline{\includegraphics[width=\textwidth]{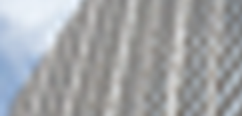}}
            \vspace{-0.15cm}
		\centerline{\tiny (b) Bicubic}
            \vspace{-0.15cm}
            \centerline{\tiny 17.36/0.2462}
		\vspace{0.07cm}
		\centerline{\includegraphics[width=\textwidth]{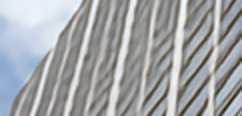}}
            \vspace{-0.15cm}
		\centerline{\tiny (f) SRFormer~\cite{zhou2023srformer}}
            \vspace{-0.15cm}
            \centerline{\tiny 18.28/0.4347}
		\vspace{0.1cm}
		
		\centerline{\includegraphics[width=\textwidth]{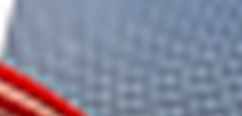}}
            \vspace{-0.15cm}
		\centerline{\tiny (b) Bicubic}
            \vspace{-0.15cm}
            \centerline{\tiny 19.18/0.2369}
		\vspace{0.07cm}
		\centerline{\includegraphics[width=\textwidth]{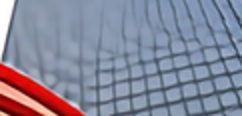}}
            \vspace{-0.15cm}
		\centerline{\tiny (f) SRFormer~\cite{zhou2023srformer}}
            \vspace{-0.15cm}
            \centerline{\tiny 20.31/0.4372}
		\vspace{0.1cm}
		
		\centerline{\includegraphics[width=\textwidth]{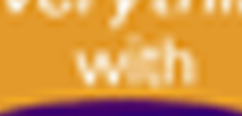}}
            \vspace{-0.15cm}
		\centerline{\tiny (b) Bicubic}
            \vspace{-0.15cm}
            \centerline{\tiny 24.53/0.7589}
		\vspace{0.07cm}
		\centerline{\includegraphics[width=\textwidth]{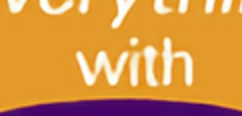}}
            \vspace{-0.15cm}
		\centerline{\tiny (f) SRFormer~\cite{zhou2023srformer}}
            \vspace{-0.15cm}
            \centerline{\tiny 32.30/0.9510}
		\vspace{0.1cm}

	\end{minipage}
	\begin{minipage}{0.16\linewidth}
		\centerline{\includegraphics[width=\textwidth]{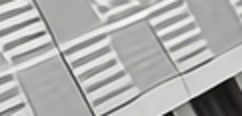}}
            \vspace{-0.15cm}
		\centerline{\tiny (c) RCAN~\cite{zhang2018image}}
            \vspace{-0.15cm}
            \centerline{\tiny 15.56/0.2426}
		\vspace{0.07cm}
		\centerline{\includegraphics[width=\textwidth]{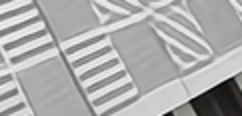}}
            \vspace{-0.15cm}
		\centerline{\tiny (g) DLGSANet~\cite{li2023dlgsanet}}
            \vspace{-0.15cm}
            \centerline{\tiny 15.19/0.1970}
		\vspace{0.1cm}
		
		\centerline{\includegraphics[width=\textwidth]{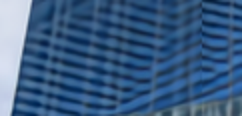}}
            \vspace{-0.15cm}
		\centerline{\tiny (c) RCAN~\cite{zhang2018image}}
            \vspace{-0.15cm}
            \centerline{\tiny 24.01/0.4411}
		\vspace{0.07cm}
		\centerline{\includegraphics[width=\textwidth]{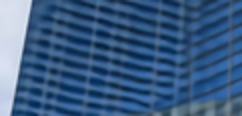}}
            \vspace{-0.15cm}
		\centerline{\tiny (g) DLGSANet~\cite{li2023dlgsanet}}
            \vspace{-0.15cm}
            \centerline{\tiny 23.33/0.4267}
		\vspace{0.1cm}
		
		\centerline{\includegraphics[width=\textwidth]{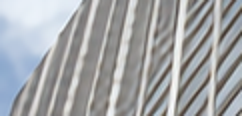}}
            \vspace{-0.15cm}
		\centerline{\tiny (c) RCAN~\cite{zhang2018image}}
            \vspace{-0.15cm}
            \centerline{\tiny 18.27/0.4319}
		\vspace{0.07cm}
		\centerline{\includegraphics[width=\textwidth]{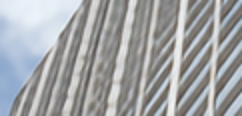}}
            \vspace{-0.15cm}
		\centerline{\tiny (g) DLGSANet~\cite{li2023dlgsanet}}
            \vspace{-0.15cm}
            \centerline{\tiny 18.80/0.4916}
		\vspace{0.1cm}
		
		\centerline{\includegraphics[width=\textwidth]{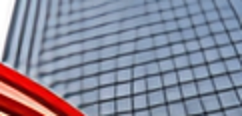}}
            \vspace{-0.15cm}
		\centerline{\tiny (c) RCAN~\cite{zhang2018image}}
            \vspace{-0.15cm}
            \centerline{\tiny 21.50/0.6078}
		\vspace{0.07cm}
		\centerline{\includegraphics[width=\textwidth]{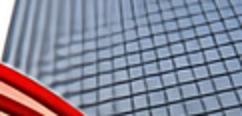}}
            \vspace{-0.15cm}
		\centerline{\tiny (g) DLGSANet~\cite{li2023dlgsanet}}
            \vspace{-0.15cm}
            \centerline{\tiny 21.03/0.5980}
		\vspace{0.1cm}
		
		\centerline{\includegraphics[width=\textwidth]{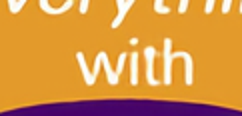}}
            \vspace{-0.15cm}
		\centerline{\tiny (c) RCAN~\cite{zhang2018image}}
            \vspace{-0.15cm}
            \centerline{\tiny 34.82/0.9675}
		\vspace{0.07cm}
		\centerline{\includegraphics[width=\textwidth]{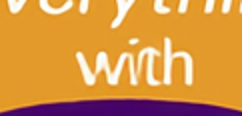}}
            \vspace{-0.15cm}
		\centerline{\tiny (g) DLGSANet~\cite{li2023dlgsanet}}
            \vspace{-0.15cm}
            \centerline{\tiny 30.33/0.9420}
		\vspace{0.1cm}

	\end{minipage}
	\begin{minipage}{0.16\linewidth}
		\centerline{\includegraphics[width=\textwidth]{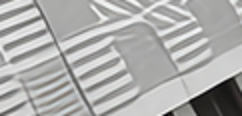}}
            \vspace{-0.15cm}
		\centerline{\tiny (d) SAN~\cite{dai2019second} }
            \vspace{-0.15cm}
            \centerline{\tiny 15.33/0.2290}
		\vspace{0.07cm}
		\centerline{\includegraphics[width=\textwidth]{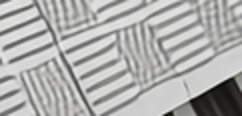}}
            \vspace{-0.15cm}
		\centerline{\tiny (i) LIPT-Base}
            \vspace{-0.15cm}
            \centerline{\textbf{\tiny 15.99/0.3482}}
		\vspace{0.1cm}
		
		\centerline{\includegraphics[width=\textwidth]{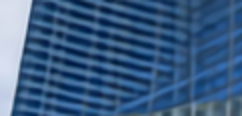}}
            \vspace{-0.15cm}
		\centerline{\tiny (d) SAN~\cite{dai2019second} }
            \vspace{-0.15cm}
            \centerline{\tiny 22.27/0.3741}
		\vspace{0.07cm}
		\centerline{\includegraphics[width=\textwidth]{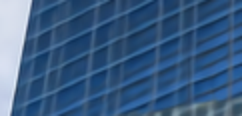}}
            \vspace{-0.15cm}
		\centerline{\tiny (i) LIPT-Base}
            \vspace{-0.15cm}
            \centerline{\textbf{\tiny 28.28/0.6286}}
		\vspace{0.1cm}
		
		\centerline{\includegraphics[width=\textwidth]{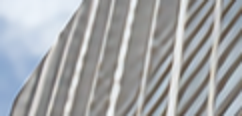}}
            \vspace{-0.15cm}
		\centerline{\tiny (d) SAN~\cite{dai2019second} }
            \vspace{-0.15cm}
            \centerline{\tiny 18.72/0.4497}
		\vspace{0.07cm}
		\centerline{\includegraphics[width=\textwidth]{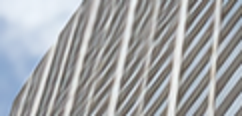}}
            \vspace{-0.15cm}
		\centerline{\tiny (i) LIPT-Base}
            \vspace{-0.15cm}
            \centerline{\textbf{\tiny 20.67/0.6595}}
		\vspace{0.1cm}
		
		\centerline{\includegraphics[width=\textwidth]{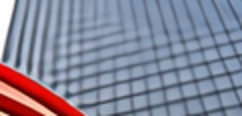}}
            \vspace{-0.15cm}
		\centerline{\tiny (d) SAN~\cite{dai2019second} }
            \vspace{-0.15cm}
            \centerline{\tiny 21.63/0.5799}
		\vspace{0.07cm}
		\centerline{\includegraphics[width=\textwidth]{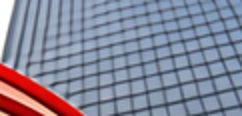}}
            \vspace{-0.15cm}
		\centerline{\tiny (i) LIPT-Base}
            \vspace{-0.15cm}
            \centerline{\textbf{\tiny 22.68/0.6774}}
		\vspace{0.1cm}
		
		\centerline{\includegraphics[width=\textwidth]{picture/supp/base_5/ppt3x4_test_BICUBIC_SRx4_demo_patch.png}}
            \vspace{-0.15cm}
		\centerline{\tiny (d) SAN~\cite{dai2019second} }
            \vspace{-0.15cm}
            \centerline{\tiny 30.68/0.9411}
		\vspace{0.07cm}
		\centerline{\includegraphics[width=\textwidth]{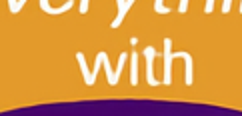}}
            \vspace{-0.15cm}
		\centerline{\tiny (i) LIPT-Base}
            \vspace{-0.15cm}
            \centerline{\textbf{\tiny 35.76/0.9726}}
		\vspace{0.1cm}

	\end{minipage}
	
	\caption{Qualitative Comparison between LIPT-Base and other methods on Urban100 and Set14 for $\times$4 SR. Zoom in for best views.}
	\vspace{-0.8em}
	\label{fig:classic}
	\vspace{-0.6em}
\end{figure*}

\begin{figure*}[t]
	\centering
	\begin{minipage}{0.205\linewidth}
		\vspace{2pt}
		\centerline{\includegraphics[width=\textwidth]{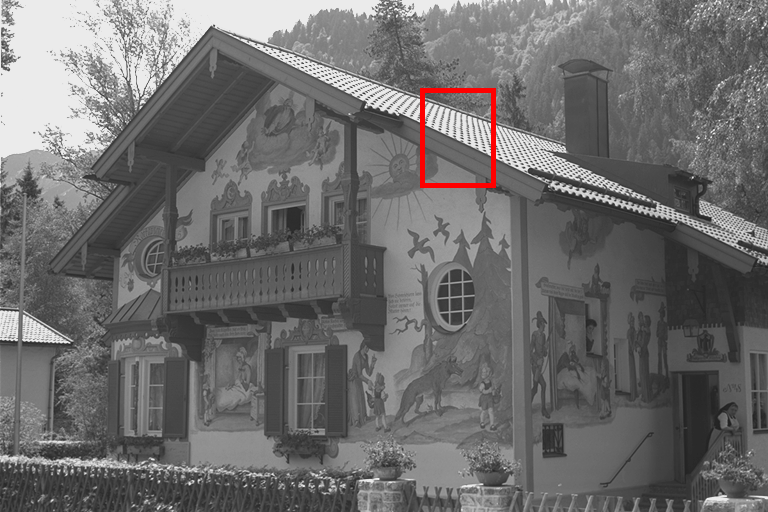}}
		\centerline{\tiny paintedhouse}
		\vspace{6pt}
		
		\vspace{2pt}
		\centerline{\includegraphics[width=\textwidth]{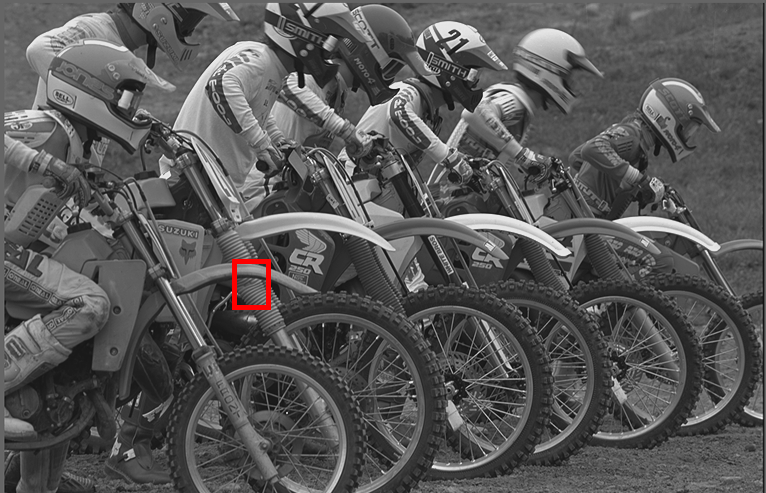}}
		\centerline{\tiny bikes}
		\vspace{6pt}
		
		\vspace{2pt}
		\centerline{\includegraphics[width=\textwidth]{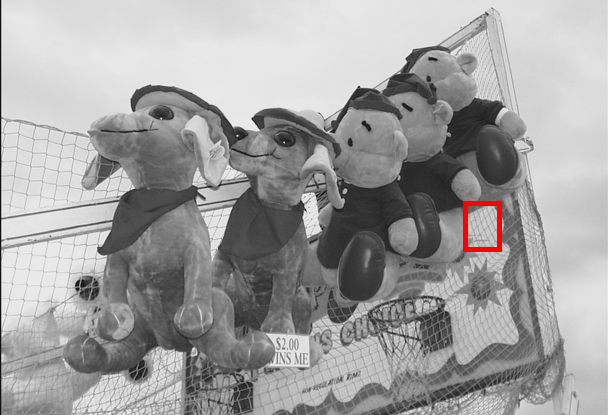}}
		\centerline{\tiny carnivaldolls}
		\vspace{6pt}
		
		\vspace{2pt}
		\centerline{\includegraphics[width=\textwidth]{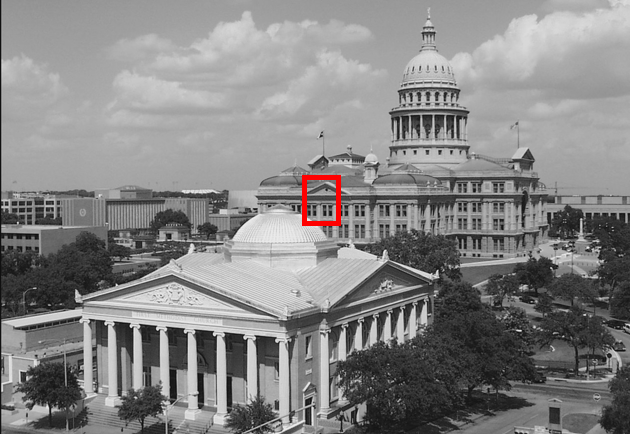}}
		\centerline{\tiny  churchandcapitol}
		\vspace{6pt}
		
		\vspace{2pt}
		\centerline{\includegraphics[width=\textwidth]{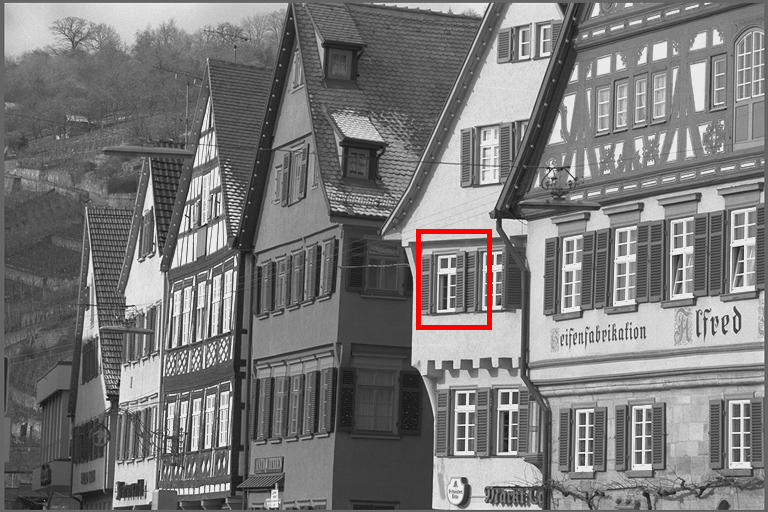}}
		\centerline{\tiny buildings}
		\vspace{6pt}
	\end{minipage}
	\begin{minipage}{0.1\linewidth}
		\vspace{3pt}
		\centerline{\includegraphics[width=\textwidth]{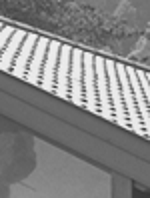}}
            \vspace{-0.15cm}
		\centerline{\tiny (a)HR}
            \vspace{-0.15cm}
            \centerline{\tiny PSNR/SSIM}
		\vspace{6pt}
		\centerline{\includegraphics[width=\textwidth]{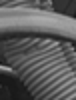}}
            \vspace{-0.15cm}
		\centerline{\tiny (a)HR}
            \vspace{-0.15cm}
            \centerline{\tiny PSNR/SSIM}
		\vspace{6pt}
		\centerline{\includegraphics[width=\textwidth]{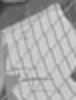}}
            \vspace{-0.15cm}
		\centerline{\tiny (a)HR}
            \vspace{-0.15cm}
            \centerline{\tiny PSNR/SSIM}
		\vspace{8pt}
		\centerline{\includegraphics[width=\textwidth]{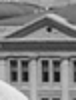}}
            \vspace{-0.15cm}
		\centerline{\tiny (a)HR}
            \vspace{-0.15cm}
            \centerline{\tiny PSNR/SSIM}
		\vspace{8pt}
		\centerline{\includegraphics[width=\textwidth]{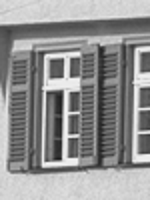}}
            \vspace{-0.15cm}
		\centerline{\tiny (a)HR}
            \vspace{-0.15cm}
            \centerline{\tiny PSNR/SSIM}
		\vspace{3pt}

	\end{minipage}
	\begin{minipage}{0.1\linewidth}
		\vspace{3pt}
		\centerline{\includegraphics[width=\textwidth]{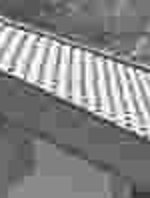}}
            \vspace{-0.15cm}
		\centerline{\tiny (b)JPEG}
            \vspace{-0.15cm}
            \centerline{\tiny 26.58/0.789}
		\vspace{6pt}
		\centerline{\includegraphics[width=\textwidth]{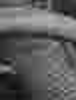}}
            \vspace{-0.15cm}
		\centerline{\tiny (b)JPEG}
            \vspace{-0.15cm}
            \centerline{\tiny 26.86/0.629}
		\vspace{6pt}
		\centerline{\includegraphics[width=\textwidth]{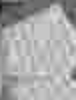}}
            \vspace{-0.15cm}
		\centerline{\tiny (b)JPEG}
            \vspace{-0.15cm}
            \centerline{\tiny 28.38/0.664}
		\vspace{8pt}
		\centerline{\includegraphics[width=\textwidth]{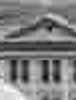}}
            \vspace{-0.15cm}
		\centerline{\tiny (b)JPEG}
            \vspace{-0.15cm}
            \centerline{\tiny 26.43/0.799}
		\vspace{8pt}
		\centerline{\includegraphics[width=\textwidth]{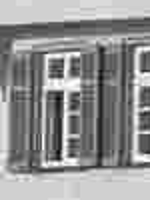}}
            \vspace{-0.15cm}
		\centerline{\tiny (b)JPEG}
            \vspace{-0.15cm}
            \centerline{\tiny 26.87/0.786}
		\vspace{3pt}
		
	\end{minipage}
	\begin{minipage}{0.1\linewidth}
		\vspace{3pt}
		\centerline{\includegraphics[width=\textwidth]{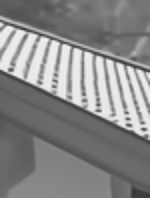}}
            \vspace{-0.15cm}
		\centerline{\tiny (c)SwinIR~\cite{liang2021swinir}}
            \vspace{-0.15cm}
            \centerline{\tiny 30.02/0.892}
		\vspace{6pt}
		\centerline{\includegraphics[width=\textwidth]{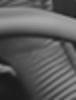}}
            \vspace{-0.15cm}
		\centerline{\tiny (c)SwinIR~\cite{liang2021swinir}}
            \vspace{-0.15cm}
            \centerline{\tiny 30.09/0.795}
		\vspace{6pt}
		\centerline{\includegraphics[width=\textwidth]{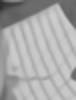}}
            \vspace{-0.15cm}
		\centerline{\tiny (c)SwinIR~\cite{liang2021swinir}}
            \vspace{-0.15cm}
            \centerline{\tiny 31.29/0.828}
		\vspace{8pt}
		\centerline{\includegraphics[width=\textwidth]{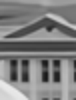}}
            \vspace{-0.15cm}
		\centerline{\tiny (c)SwinIR~\cite{liang2021swinir}}
            \vspace{-0.15cm}
            \centerline{\tiny 29.33/0.894}
		\vspace{8pt}
		\centerline{\includegraphics[width=\textwidth]{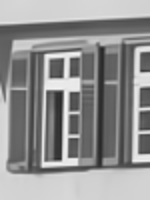}}
            \vspace{-0.15cm}
		\centerline{\tiny (c)SwinIR~\cite{liang2021swinir}}
            \vspace{-0.15cm}
            \centerline{\tiny 30.29/0.869}
		\vspace{3pt}
		
	\end{minipage}
		\begin{minipage}{0.1\linewidth}
		\vspace{3pt}
		\centerline{\includegraphics[width=\textwidth]{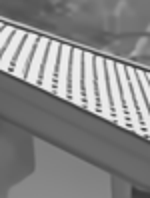}}
            \vspace{-0.15cm}
		\centerline{\tiny (d)ART~\cite{zhang2022accurate}}
            \vspace{-0.15cm}
            \centerline{\tiny 30.17/0.894}
		\vspace{6pt}
		\centerline{\includegraphics[width=\textwidth]{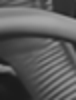}}
            \vspace{-0.15cm}
		\centerline{\tiny (d)ART~\cite{zhang2022accurate}}
            \vspace{-0.15cm}
            \centerline{\tiny 30.29/0.810}
		\vspace{6pt}
		\centerline{\includegraphics[width=\textwidth]{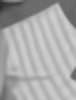}}
            \vspace{-0.15cm}
		\centerline{\tiny (d)ART~\cite{zhang2022accurate}}
            \vspace{-0.15cm}
            \centerline{\tiny 30.98/0.798}
		\vspace{8pt}
		\centerline{\includegraphics[width=\textwidth]{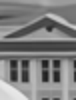}}
            \vspace{-0.15cm}
		\centerline{\tiny (d)ART~\cite{zhang2022accurate}}
            \vspace{-0.15cm}
            \centerline{\tiny 29.65/0.903}
		\vspace{8pt}
		\centerline{\includegraphics[width=\textwidth]{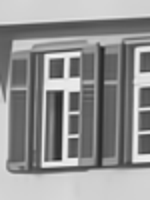}}
            \vspace{-0.15cm}
		\centerline{\tiny (d)ART~\cite{zhang2022accurate}}
            \vspace{-0.15cm}
            \centerline{\tiny 30.64/0.876}
		\vspace{3pt}

	\end{minipage}
		\begin{minipage}{0.1\linewidth}
		\vspace{3pt}
		\centerline{\includegraphics[width=\textwidth]{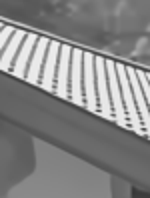}}
            \vspace{-0.15cm}
		\centerline{\tiny (e)CAT~\cite{chen2022cross}}
            \vspace{-0.15cm}
            \centerline{\tiny 30.24/0.891}
		\vspace{6pt}
		\centerline{\includegraphics[width=\textwidth]{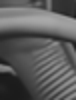}}
            \vspace{-0.15cm}
		\centerline{\tiny (e)CAT~\cite{chen2022cross}}
            \vspace{-0.15cm}
            \centerline{\tiny 29.78/0.795}
		\vspace{6pt}
		\centerline{\includegraphics[width=\textwidth]{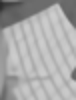}}
            \vspace{-0.15cm}
		\centerline{\tiny (e)CAT~\cite{chen2022cross}}
            \vspace{-0.15cm}
            \centerline{\tiny 31.41/0.834}
		\vspace{8pt}
		\centerline{\includegraphics[width=\textwidth]{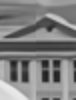}}
            \vspace{-0.15cm}
		\centerline{\tiny (e)CAT~\cite{chen2022cross}}
            \vspace{-0.15cm}
            \centerline{\tiny 29.14/0.882}
		\vspace{8pt}
		\centerline{\includegraphics[width=\textwidth]{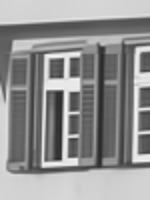}}
            \vspace{-0.15cm}
		\centerline{\tiny (e)CAT~\cite{chen2022cross}}
            \vspace{-0.15cm}
            \centerline{\tiny 30.88/0.880}
		\vspace{3pt}

	\end{minipage}
	\begin{minipage}{0.1\linewidth}
		\vspace{3pt}
		\centerline{\includegraphics[width=\textwidth]{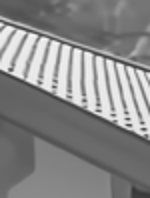}}
            \vspace{-0.15cm}
		\centerline{\tiny (e)GRL-S~\cite{li2023efficient}}
            \vspace{-0.15cm}
            \centerline{\tiny 30.20/0.892}
		\vspace{6pt}
		\centerline{\includegraphics[width=\textwidth]{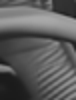}}
            \vspace{-0.15cm}
		\centerline{\tiny (e)GRL-S~\cite{li2023efficient}}
            \vspace{-0.15cm}
            \centerline{\tiny 29.22/0.754}
		\vspace{6pt}
		\centerline{\includegraphics[width=\textwidth]{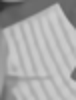}}
            \vspace{-0.15cm}
		\centerline{\tiny (e)GRL-S~\cite{li2023efficient}}
            \vspace{-0.15cm}
            \centerline{\tiny 31.00/0.807}
		\vspace{8pt}
		\centerline{\includegraphics[width=\textwidth]{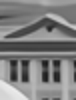}}
            \vspace{-0.15cm}
		\centerline{\tiny (e)GRL-S~\cite{li2023efficient}}
            \vspace{-0.15cm}
            \centerline{\tiny 29.04/0.885}
		\vspace{8pt}
		\centerline{\includegraphics[width=\textwidth]{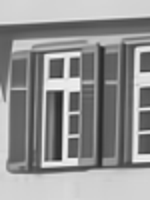}}
            \vspace{-0.15cm}
		\centerline{\tiny (e)GRL-S~\cite{li2023efficient}}
            \vspace{-0.15cm}
            \centerline{\tiny 30.39/0.871}
		\vspace{3pt}

	\end{minipage}
	\begin{minipage}{0.1\linewidth}
		\vspace{3pt}
		\centerline{\includegraphics[width=\textwidth]{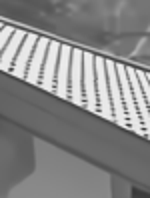}}
            \vspace{-0.15cm}
		\centerline{\tiny (f) LIPT}
            \vspace{-0.15cm}
            \centerline{\textbf{\tiny 30.85/0.899}}
		\vspace{6pt}
		\centerline{\includegraphics[width=\textwidth]{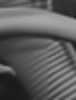}}
            \vspace{-0.15cm}
		\centerline{\tiny (f) LIPT}
            \vspace{-0.15cm}
            \centerline{\textbf{\tiny 30.40/0.818}}
		\vspace{6pt}
		\centerline{\includegraphics[width=\textwidth]{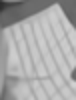}}
            \vspace{-0.15cm}
		\centerline{\tiny (f) LIPT}
            \vspace{-0.15cm}
            \centerline{\textbf{\tiny 31.47/0.835}}
		\vspace{8pt}
		\centerline{\includegraphics[width=\textwidth]{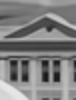}}
            \vspace{-0.15cm}
		\centerline{\tiny (f) LIPT}
            \vspace{-0.15cm}
            \centerline{\textbf{\tiny 31.06/0.932}}
		\vspace{8pt}
		\centerline{\includegraphics[width=\textwidth]{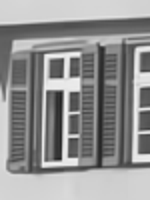}}
            \vspace{-0.15cm}
		\centerline{\tiny (f) LIPT}
            \vspace{-0.15cm}
            \centerline{\textbf{\tiny 31.75/0.921}}
		\vspace{3pt}
		
	\end{minipage}
	\vspace{-0.8em}
	\caption{Visual Comparison on LIVE1 for CAR q=10.}
	\label{supfig:jpeg}
	\vspace{-1em}
\end{figure*}

\begin{figure*}[t]
	\centering
	\begin{minipage}{0.2\linewidth}
		\vspace{2pt}
		\centerline{\includegraphics[width=\textwidth]{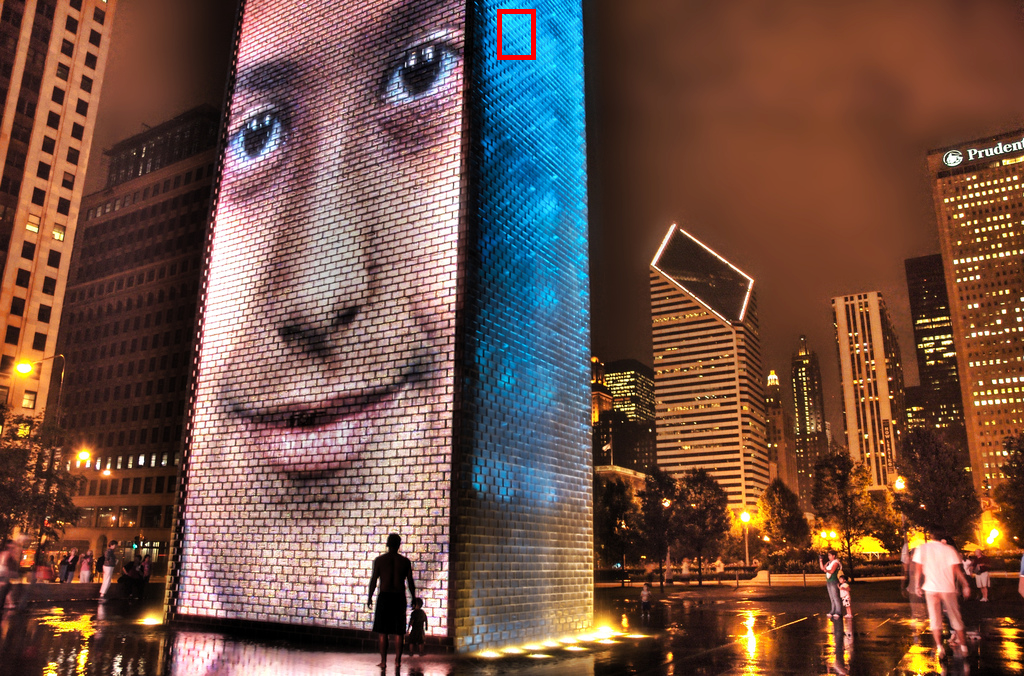}}
		\centerline{\tiny img076 from Urban100}
		\vspace{6pt}
		
		\vspace{2pt}
		\centerline{\includegraphics[width=\textwidth]{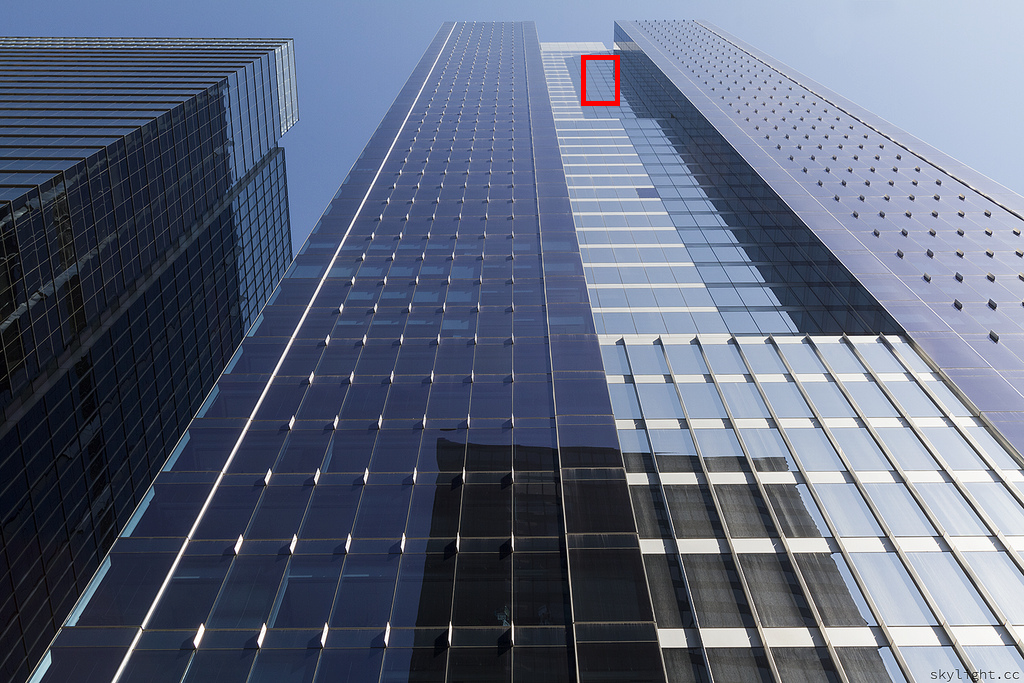}}
		\centerline{\tiny img033 from Urban100}
		\vspace{6pt}
		
		\vspace{2pt}
		\centerline{\includegraphics[width=\textwidth]{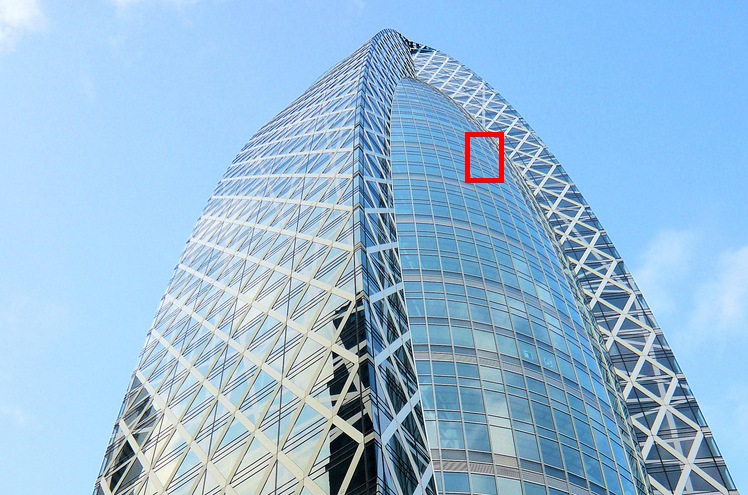}}
		\centerline{\tiny img039 from Urban100}
		\vspace{6pt}
		
		\vspace{2pt}
		\centerline{\includegraphics[width=\textwidth]{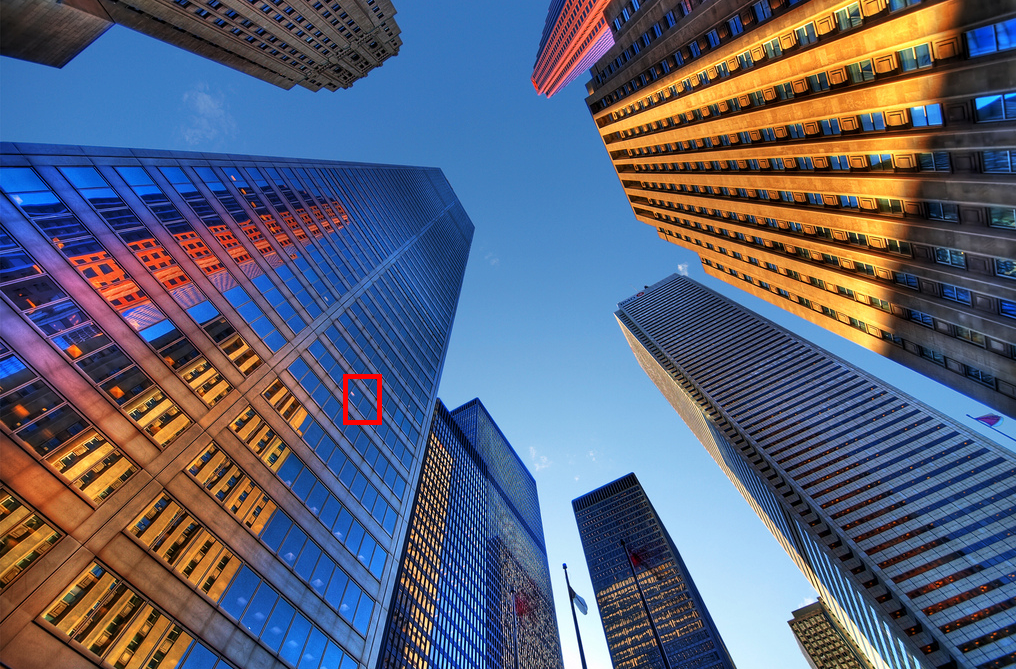}}
		\centerline{\tiny img012 from Urban100}
		\vspace{6pt}
		
		\vspace{2pt}
		\centerline{\includegraphics[width=\textwidth]{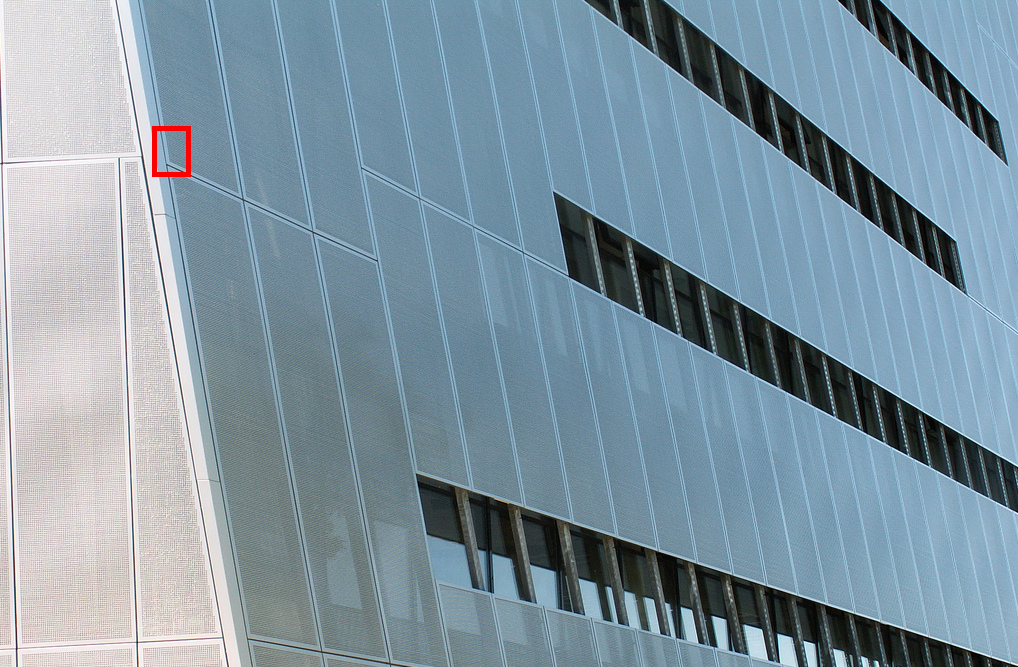}}
		\centerline{\tiny img026 from Urban100}
		\vspace{4pt}
	\end{minipage}
	\begin{minipage}{0.1\linewidth}
		\vspace{3pt}
		\centerline{\includegraphics[width=\textwidth]{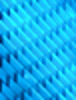}}
            \vspace{-0.15cm}
		\centerline{\tiny (a)HR}
            \vspace{-0.15cm}
            \centerline{\tiny PSNR/SSIM}
		\vspace{6pt}
		\centerline{\includegraphics[width=\textwidth]{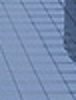}}
            \vspace{-0.15cm}
		\centerline{\tiny (a)HR}
            \vspace{-0.15cm}
            \centerline{\tiny PSNR/SSIM}
		\vspace{6pt}
		\centerline{\includegraphics[width=\textwidth]{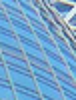}}
            \vspace{-0.15cm}
		\centerline{\tiny (a)HR}
            \vspace{-0.15cm}
            \centerline{\tiny PSNR/SSIM}
		\vspace{6pt}
		\centerline{\includegraphics[width=\textwidth]{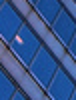}}
            \vspace{-0.15cm}
		\centerline{\tiny (a)HR}
            \vspace{-0.15cm}
            \centerline{\tiny PSNR/SSIM}
		\vspace{4pt}
		\centerline{\includegraphics[width=\textwidth]{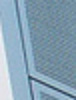}}
            \vspace{-0.15cm}
		\centerline{\tiny (a)HR}
            \vspace{-0.15cm}
            \centerline{\tiny PSNR/SSIM}
		\vspace{3pt}

	\end{minipage}
	\begin{minipage}{0.1\linewidth}
		\vspace{3pt}
		\centerline{\includegraphics[width=\textwidth]{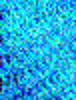}}
            \vspace{-0.15cm}
		\centerline{\tiny (b)Noisy}
            \vspace{-0.15cm}
            \centerline{\tiny 21.99/0.499}
		\vspace{6pt}
		\centerline{\includegraphics[width=\textwidth]{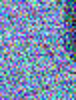}}
            \vspace{-0.15cm}
		\centerline{\tiny (b)Noisy}
            \vspace{-0.15cm}
            \centerline{\tiny 21.15/0.293}
		\vspace{6pt}
		\centerline{\includegraphics[width=\textwidth]{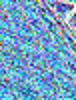}}
            \vspace{-0.15cm}
		\centerline{\tiny (b)Noisy}
            \vspace{-0.15cm}
            \centerline{\tiny 21.27/0.462}
		\vspace{6pt}
		\centerline{\includegraphics[width=\textwidth]{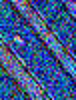}}
            \vspace{-0.15cm}
		\centerline{\tiny (b)Noisy}
            \vspace{-0.15cm}
            \centerline{\tiny 21.72/0.415}
		\vspace{4pt}
		\centerline{\includegraphics[width=\textwidth]{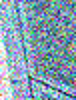}}
            \vspace{-0.15cm}
		\centerline{\tiny (b)Noisy}
            \vspace{-0.15cm}
            \centerline{\tiny 21.22/0.351}
		\vspace{3pt}
		
	\end{minipage}
	\begin{minipage}{0.1\linewidth}
		\vspace{3pt}
		\centerline{\includegraphics[width=\textwidth]{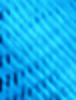}}
            \vspace{-0.15cm}
		\centerline{\tiny (c)SwinIR~\cite{liang2021swinir}}
            \vspace{-0.15cm}
            \centerline{\tiny 28.36/0.819}
		\vspace{6pt}
		\centerline{\includegraphics[width=\textwidth]{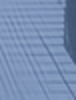}}
            \vspace{-0.15cm}
		\centerline{\tiny (c)SwinIR~\cite{liang2021swinir}}
            \vspace{-0.15cm}
            \centerline{\tiny 33.74/0.877}
		\vspace{6pt}
		\centerline{\includegraphics[width=\textwidth]{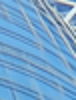}}
            \vspace{-0.15cm}
		\centerline{\tiny (c)SwinIR~\cite{liang2021swinir}}
            \vspace{-0.15cm}
            \centerline{\tiny 29.29/0.820}
		\vspace{6pt}
		\centerline{\includegraphics[width=\textwidth]{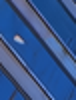}
            \vspace{-0.15cm}}
		\centerline{\tiny (c)SwinIR~\cite{liang2021swinir}}
            \vspace{-0.15cm}
            \centerline{\tiny 32.19/0.880}
		\vspace{4pt}
		\centerline{\includegraphics[width=\textwidth]{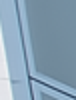}}
            \vspace{-0.15cm}
		\centerline{\tiny (c)SwinIR~\cite{liang2021swinir}}
            \vspace{-0.15cm}
            \centerline{\tiny 35.02/0.867}
		\vspace{3pt}
		
	\end{minipage}
	\begin{minipage}{0.1\linewidth}
		\vspace{3pt}
		\centerline{\includegraphics[width=\textwidth]{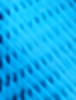}}
            \vspace{-0.15cm}
		\centerline{\tiny (d)RT~\cite{zamir2022restormer}}
            \vspace{-0.15cm}
            \centerline{\tiny 28.17/0.765}
		\vspace{6pt}
		\centerline{\includegraphics[width=\textwidth]{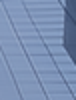}}
            \vspace{-0.15cm}
		\centerline{\tiny (d)RT~\cite{zamir2022restormer}}
            \vspace{-0.15cm}
            \centerline{\tiny 34.79/0.908}
		\vspace{6pt}
		\centerline{\includegraphics[width=\textwidth]{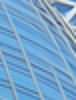}}
            \vspace{-0.15cm}
		\centerline{\tiny (d)RT~\cite{zamir2022restormer}}
            \vspace{-0.15cm}
            \centerline{\tiny 29.53/0.830}
		\vspace{6pt}
		\centerline{\includegraphics[width=\textwidth]{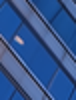}}
            \vspace{-0.15cm}
		\centerline{\tiny (d)RT~\cite{zamir2022restormer}}
            \vspace{-0.15cm}
            \centerline{\tiny 32.66/0.908}
		\vspace{4pt}
		\centerline{\includegraphics[width=\textwidth]{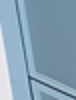}}
            \vspace{-0.15cm}
		\centerline{\tiny (d)RT~\cite{zamir2022restormer}}
            \vspace{-0.15cm}
            \centerline{\tiny 35.62/0.874}
		\vspace{3pt}

	\end{minipage}
	\begin{minipage}{0.1\linewidth}
		\vspace{3pt}
		\centerline{\includegraphics[width=\textwidth]{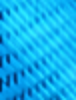}}
            \vspace{-0.15cm}
		\centerline{\tiny (d)ART~\cite{zhang2022accurate}}
            \vspace{-0.15cm}
            \centerline{\tiny 29.53/0.863}
		\vspace{6pt}
		\centerline{\includegraphics[width=\textwidth]{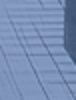}}
            \vspace{-0.15cm}
		\centerline{\tiny (d)ART~\cite{zhang2022accurate}}
            \vspace{-0.15cm}
            \centerline{\tiny 34.11/0.888}
		\vspace{6pt}
		\centerline{\includegraphics[width=\textwidth]{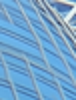}}
            \vspace{-0.15cm}
		\centerline{\tiny (d)ART~\cite{zhang2022accurate}}
            \vspace{-0.15cm}
            \centerline{\tiny 29.82/0.842}
		\vspace{6pt}
		\centerline{\includegraphics[width=\textwidth]{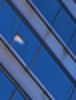}}
            \vspace{-0.15cm}
		\centerline{\tiny (d)ART~\cite{zhang2022accurate}}
            \vspace{-0.15cm}
            \centerline{\tiny 33.16/0.904}
		\vspace{4pt}
		\centerline{\includegraphics[width=\textwidth]{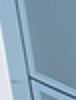}}
            \vspace{-0.15cm}
		\centerline{\tiny (d)ART~\cite{zhang2022accurate}}
            \vspace{-0.15cm}
            \centerline{\tiny 35.25/0.871}
		\vspace{3pt}

	\end{minipage}
	\begin{minipage}{0.1\linewidth}
		\vspace{3pt}
		\centerline{\includegraphics[width=\textwidth]{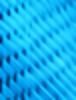}}
            \vspace{-0.15cm}
		\centerline{\tiny (e)GRL-S~\cite{li2023efficient}}
            \vspace{-0.15cm}
            \centerline{\tiny 29.32/0.862}
		\vspace{6pt}
		\centerline{\includegraphics[width=\textwidth]{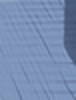}}
            \vspace{-0.15cm}
		\centerline{\tiny (e)GRL-S~\cite{li2023efficient}}
            \vspace{-0.15cm}
            \centerline{\tiny 33.71/0.857}
		\vspace{6pt}
		\centerline{\includegraphics[width=\textwidth]{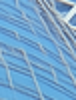}}
            \vspace{-0.15cm}
		\centerline{\tiny (e)GRL-S~\cite{li2023efficient}}
            \vspace{-0.15cm}
            \centerline{\tiny 29.24/0.841}
		\vspace{6pt}
		\centerline{\includegraphics[width=\textwidth]{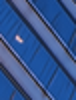}}
            \vspace{-0.15cm}
		\centerline{\tiny (e)GRL-S~\cite{li2023efficient}}
            \vspace{-0.15cm}
            \centerline{\tiny 32.99/0.911}
		\vspace{4pt}
		\centerline{\includegraphics[width=\textwidth]{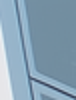}}
            \vspace{-0.15cm}
		\centerline{\tiny (e)GRL-S~\cite{li2023efficient}}
            \vspace{-0.15cm}
            \centerline{\tiny 35.16/0.876}
		\vspace{3pt}

	\end{minipage}
	\begin{minipage}{0.1\linewidth}
		\vspace{3pt}
		\centerline{\includegraphics[width=\textwidth]{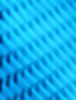}}
            \vspace{-0.15cm}
		\centerline{\tiny (f)LIPT}
            \vspace{-0.15cm}
            \centerline{\textbf{\tiny 30.20/0.883}}
		\vspace{6pt}
		\centerline{\includegraphics[width=\textwidth]{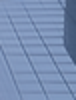}}
            \vspace{-0.15cm}
		\centerline{\tiny (f)LIPT}
            \vspace{-0.15cm}
            \centerline{\textbf{\tiny 35.00/0.922}}
		\vspace{6pt}
		\centerline{\includegraphics[width=\textwidth]{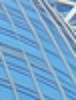}}
            \vspace{-0.15cm}
		\centerline{\tiny (f)LIPT}
            \vspace{-0.15cm}
            \centerline{\textbf{\tiny 29.85/0.848}}
		\vspace{6pt}
		\centerline{\includegraphics[width=\textwidth]{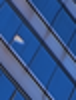}}
            \vspace{-0.15cm}
		\centerline{\tiny (f)LIPT}
            \vspace{-0.15cm}
            \centerline{\textbf{\tiny 33.50/0.920}}
		\vspace{4pt}
		\centerline{\includegraphics[width=\textwidth]{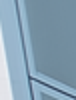}}
            \vspace{-0.15cm}
		\centerline{\tiny (f)LIPT}
            \vspace{-0.15cm}
            \centerline{\textbf{\tiny 35.93/0.877}}
		\vspace{3pt}
		
	\end{minipage}
	\vspace{-0.8em}
	\caption{Visual Comparison on Urban100 for image denoising $\sigma=50$. RT indicates Restormer.}
	\label{fig:denoising}
	\vspace{-1em}
\end{figure*}

We add more visualization results of our LIPT-Tiny, LIPT-Small, LIPT-Base, making comparisons with other models on different datasets for $ \times 4$ SR in Fig.~\ref{fig:Tiny}, Fig.~\ref{supfig:light}, and Fig.~\ref{fig:classic}, respectively. 
For example, in the first two rows of Fig.~\ref{fig:Tiny} (\emph{i.e.}, img012 and img024 SR) on Urban100, we can observe that our LIPT-Tiny recovers the flat and clean edge
 details. LIPT-Small reconstructs the clearest building outline in img76053 on BSD100. 
As shown in the first and second rows of Fig.~\ref{fig:classic}, all previous SOTA methods are difficult to recover high-frequency dense areas. In contrast, LIPT-Base is able to recover more details for high-quality SR. In addition, more visual comparisons of the JPEG CAR with $q=10$ and image denoising with $\sigma = 50$ are presented in Fig.~\ref{supfig:jpeg} and Fig.~\ref{fig:denoising}, respectively. For JPEG CAR in Fig.~\ref{supfig:jpeg}, LIPT is able to generate the clearest and sharpest roof details, while other methods are easy to generate the blur textures. 
For image denoising in Fig~\ref{fig:denoising}, our LIPT effectively recovers the lines for img039 denoising, while other compared methods fail to restore the line structures.

\subsubsection{Additional Ablation Study}

\begin{table}[t]
\centering 
\footnotesize
\caption{Ablation study at the proportion of MSA-Conv without using NVSM-SA and HRM for $\times2$ SR.}
\label{tab:MSA-Conv}
\begin{tabular}{|c|c|c|c|c|c|c|}
\hline
MSA:CB        & GPU (ms) & Set5  & Set14 & B100  & U100  & M109  \\ \hline
1:1 & 133      & 38.17 & 33.94 & 32.30  & 32.76 & 39.11     \\ \hline
1:2 & 112      & 38.14 & 33.87 & 32.30  & 32.59 & 38.99      \\ \hline
1:3 & 99      & 38.19  & 33.95 & 32.32 & 32.72 & 39.09     \\ \hline
1:4 & 95      & 38.17 & 33.87 & 32.31 & 32.63 & 38.99     \\ \hline
\end{tabular}
\vspace{-0.3cm}
\end{table}

\textbf{The proportion of MSA-Conv.} As shown in Table~\ref{tab:MSA-Conv},  we make the different proportions of MSA-Conv \emph{w.r.t.} latency and PSNR for comparison.  
We fix the total number of convolution blocks (CB) and MSA layers and change the internal ratio for experiments without using NVSM-SA and HRM.
We observe that increasing the proportion of CB achieves faster inference, which is relatively saturated at the ratio of 1:3. However, the excessively high proportion of CB decreases the reconstruction performance, due to the decreasing of the model's ability to capture long-range dependencies. Note that setting a rate of MSA:CB to $1:3$ achieves the best trade-off between inference time and PSNR.

\textbf{Effect of the high-frequency information extraction branch in HRM.} To demonstrate the superiority of our high-frequency information extraction branch on the low-level visual tasks, we conduct an ablation experiment on this branch in Tab.~\ref{tab:high-frequency}. This branch is able to improve PSNR of 0.05dB and 0.04dB on Urban100 and Manga109, respectively. This is due to the usage of high-frequency information is a benefit for the reconstruction of dense texture and edge details, especially on Urban100.
\begin{table}[]
\centering
\footnotesize
\vspace{-0.5cm}
\caption{Ablation study for the high-frequency information extraction branch $G_{is}\circ conv1$ in HRM using LIPT-Small.}
\label{tab:high-frequency}
\begin{tabular}{c|c|c|c|c|c}
\hline
  $G_{is}\circ conv1$      & Set5  & Set14 & BSD100  & Urban100  & Manga109  \\ \hline
\Checkmark     & 38.20 & 33.92 & 32.36 & 32.87 & 39.21 \\ \hline
\XSolidBrush & 38.19 & 33.92 & 32.35 & 32.82 & 39.17 \\ \hline
\end{tabular}
\vspace{-0.8cm}
\end{table}

\end{document}